\documentclass{article} % For LaTeX2e

\usepackage{main,times}

\usepackage{amsmath,amsfonts,bm}

\def\eqref#1{equation~\ref{#1}}
\def\1{\bm{1}}

\DeclareMathAlphabet{\mathsfit}{\encodingdefault}{\sfdefault}{m}{sl}
\SetMathAlphabet{\mathsfit}{bold}{\encodingdefault}{\sfdefault}{bx}{n}

\usepackage{hyperref}
\usepackage{url}

\hypersetup{
    colorlinks = true,
    allcolors = {purple},
    linkbordercolor = {white},
}

\usepackage{amsmath}
\usepackage{mathtools}
\usepackage{amssymb}
\usepackage{amsthm}
\usepackage[linesnumbered,ruled,vlined]{algorithm2e}
\usepackage{bbm} % \mathbbm{1} indicator function
\usepackage{tabularx}
\usepackage{longtable}
\usepackage{booktabs}
\usepackage{multirow}
\usepackage[dvipsnames]{xcolor}
\usepackage{enumitem}
\usepackage{float}
\usepackage{epigraph}

\usepackage{tikz}
\usetikzlibrary{arrows.meta, positioning, fit, backgrounds, calc}

\definecolor{nfgray}{HTML}{F1EFE8}   % gray fill
\definecolor{nsgray}{HTML}{5F5E5A}   % gray stroke / muted text
\definecolor{ntgray}{HTML}{2C2C2A}   % gray node text
\definecolor{nfpurp}{HTML}{EEEDFE}   % purple fill
\definecolor{nspurp}{HTML}{534AB7}   % purple stroke
\definecolor{ntpurp}{HTML}{26215C}   % purple node text
\definecolor{gradc}{HTML}{D85A30}    % backward-gradient arrow
\definecolor{gradtxt}{HTML}{993C1D}  % backward-gradient label
\definecolor{outred}{HTML}{A32D2D}   % collapse outcome
\definecolor{outgrn}{HTML}{27500A}   % recovery outcome
\definecolor{fogc}{HTML}{888780}     % fog overlay

\DeclareSymbolFont{bbold}{U}{bbold}{m}{n}
\DeclareSymbolFontAlphabet{\mathbbold}{bbold}

\newtheorem{theorem}{Theorem}[section]

\newtheorem{definition}[theorem]{Definition}
\usepackage{minitoc}

\usepackage{listings}
\usepackage[T1]{fontenc}
\usepackage[scaled=0.85]{beramono}

\definecolor{paperprimary}{RGB}{191, 0, 64}
\definecolor{papernavy}{RGB}{0, 111, 185}
\definecolor{paperolive}{RGB}{57, 127, 50}
\definecolor{warmcomment}{RGB}{100, 100, 80}
\definecolor{warmkeyword}{RGB}{180, 60, 0}
\definecolor{warmstring}{RGB}{160, 82, 45}
\definecolor{warmnumber}{RGB}{120, 110, 100}

\lstdefinestyle{mystyle}{
    language=Python,
    backgroundcolor=\color{purple!4},
    commentstyle=\color{warmcomment}\itshape,
    keywordstyle=\color{warmkeyword}\bfseries,
    stringstyle=\color{warmstring},
    basicstyle=\ttfamily\footnotesize,
    numberstyle=\tiny\color{warmnumber},
    breakatwhitespace=false,
    breaklines=true,
    captionpos=b,
    keepspaces=true,
    numbers=left,
    numbersep=5pt,
    showspaces=false,
    showstringspaces=false,
    showtabs=false,
    tabsize=4,
    frame=single,
    framerule=0.5pt,
    rulecolor=\color{warmnumber!40},
}
\definecolor{abstractbg}{RGB}{250, 240, 240}

\usepackage[skins]{tcolorbox}
\definecolor{abstractbg}{RGB}{250, 240, 240}

\renewenvironment{abstract}{%
    \begin{tcolorbox}[
        enhanced,
        colback=abstractbg,
        colframe=purple,
        coltitle=purple,
        colbacktitle=abstractbg,
        boxrule=0.2pt,
        arc=2mm,
        titlerule=0pt,
        left=10pt, right=10pt, top=5pt, bottom=10pt,
        fonttitle=\bfseries\large,
        title=Abstract,
        halign title=center,
    ]
}{%
    \end{tcolorbox}
}

\usepackage{etoolbox}
\AtBeginEnvironment{thebibliography}{\clearpage}

\title{\centering \color{purple} \fontsize{13pt}{16pt}\selectfont The Reciprocity Gradient}

\usepackage{authblk}
\author[1, 5]{Yue Lin}
\author[2, 4]{Pascal Poupart}
\author[2, 4]{Shuhui Zhu}
\author[1]{Dan Qiao}
\author[3]{Wenhao Li}
\author[5]{\\Yuan Liu}
\author[1]{Hongyuan Zha}
\author[1, 4]{Baoxiang Wang}

\affil[1]{The Chinese University of Hong Kong, Shenzhen}
\affil[2]{University of Waterloo}
\affil[3]{Tongji University}
\affil[4]{Vector Institute}
\affil[5]{LIGHTSPEED}

\iclrfinalcopy % Uncomment for camera-ready version, but NOT for submission.

\begin{document}

% toc
\doparttoc
% \faketableofcontents

\maketitle
\thispagestyle{empty}

\begin{abstract}
Communication is fundamental to sustaining reciprocity and cooperation in strategic interactions. We identify and formulate the \emph{influence attribution problem} as the central optimization difficulty inherent in such dynamics for a learning agent: any action or signal the agent emits reshapes the reputations of many third parties along combinatorially branching paths before feeding back into its own future rewards, forcing the agent to account for all of these indirect channels at once when choosing every action. To address this, we introduce the \emph{reciprocity gradient}, which explicitly backpropagates reward gradients through private estimators of opponents' policies trained from public observations. The gradient flows through the reputation chain itself analytically, rather than being estimated from sampled returns. It jointly optimizes actions and evaluative signals without intrinsic rewards or reward shaping. Empirically, the method recovers near-optimal context-sensitive policies, while sample-based baselines collapse into constant-output policies.
\end{abstract}

%\newpage

% \vspace{1em}
%\setcounter{tocdepth}{1}
%\tableofcontents

%\setcounter{page}{0}

%\newpage
%\setcounter{page}{1}

% ── Shared content (sections + bibliography + appendix) ─────────────────
\section{Introduction}

\setlength{\epigraphwidth}{0.9\textwidth}
\epigraph{Ye cannot live for yourselves; a thousand fibres connect you with your fellow-men, and along those fibres, as along sympathetic\footnotemark\ threads, run your actions as causes, and return to you as effects.}{--- \textup{\citet{melvill1855partaking}}}

\footnotetext{Here, ``sympathetic'' denotes physical resonance between coupled systems (e.g., sympathetic strings in acoustics).}
%devoid of any anthropomorphic or emotional connotation.

As artificial agents are increasingly deployed in socially interactive ecosystems, ranging from e-commerce platforms governed by complex seller-rating algorithms to digital gig economies (e.g., Uber, Airbnb, and Upwork) reliant on two-way peer evaluations, endowing them with the ability to navigate reputation mechanisms has become a critical challenge. In both natural societies and modern digital platforms, cooperation and trust are largely sustained through two primary channels: direct reciprocity, where repeated interactions allow partners to reward past behavior~\citep{trivers1971evolution}, and indirect reciprocity, where behavior is regulated by third parties observing a public reputation~\citep{nowak1998evolution,nowak2005evolution,alexander2017biology}. From an artificial intelligence perspective, a fundamental motivation of this work is understanding whether a single learning agent can seamlessly integrate into an established population operating under a complex, potentially opaque rating system. To maximize long-term utility, the agent must learn not only to act appropriately but also to manage its standing by decoding the community's evaluation criteria.

However, learning to optimize behavior in such reputation-mediated games exposes a fundamental structural difficulty, which we formalize as the \textbf{influence attribution problem}. In a reputation system, the causal chain linking an agent's current action to its future rewards is deeply indirect. When agent $i$ acts toward agent $j$, the immediate environment does not provide the full learning feedback. Instead, the feedback is routed through a third party's reputational assessment (gossip) of $i$, which subsequently shifts the behavioral policy of future partners interacting with $i$. Because this feedback is delayed, diffuse, and heavily dependent on the internal logic of other agents, sample-based estimators (score-function policy gradients and temporal-difference bootstrapping) struggle to attribute expected returns to specific past actions, since they do not exploit the analytical structure of the reputation chain. Consequently, agents trained with such estimators frequently collapse into trivial, constant-output policies (e.g., unconditional cooperation or mutual defection), failing to learn the context-dependent discrimination required for complex social norms.

To address this structural bottleneck, we introduce the \textbf{reciprocity gradient}, a gradient-based optimization framework that explicitly mathematically stitches this broken causal chain back together. Rather than treating the environment and other agents as a black box, the reciprocity gradient directly backpropagates the expected reward gradients through the multi-agent reputation dynamics. This formulation enables the agent to optimize not only its direct actions but also its evaluative signals (gossip), leveraging the latter as a deeply indirect channel to steer the future behavior of the population to its advantage. By tracing the influence of both the agent's actions and its evaluative signals through the opponents' assessment functions and back into future interactions, we provide the learning agent with an analytic, highly directed optimization feedback that explicitly maps how its behavior shapes its social standing and, ultimately, its utility. To execute this without the unrealistic assumption of oracle access to opponents' internal parameters~\citep{foerster2018learning,letcher2019stable,willi2022cola}, we integrate \textbf{differentiable opponent modeling}~\citep{albrecht2018autonomous,lin2023information,li2024backpropagation}. By fitting private, fully differentiable surrogate estimators of opponents' policies strictly from public observations, the agent reconstructs the causal computation graph internally and computes the reciprocity gradient without requiring any privileged access.

We frame the empirical investigation around the question of whether the reciprocity gradient operates as a competent \emph{best-response learner}: against a given, established population, can a single learning agent compute a near-optimal, context-dependent strategy purely through gradient ascent? Our empirical evaluations affirmatively answer this question. On the joint setting under strictly observational access (private differentiable surrogates fitted from public observations only), the reciprocity gradient simultaneously recovers a context-dependent action rule and a context-dependent signal scheme, reaching $99\%$ of the full-cooperation reference and matching the $99\%$ obtained under an oracle-access reference. Strong continuous-control baselines on the same environment, DPG~\citep{silver2014deterministic}, DDPG~\citep{lillicrap2015ddpg}, TD3~\citep{fujimoto2018addressing}, and the reputation-shaped LR2~\citep{ren2025bottom}, all collapse to constant-output policies on every setting that requires a context-dependent policy.

\paragraph{Contributions.}
\begin{enumerate}[label=\textbf{(\roman*)},leftmargin=2.2em,labelsep=0.5em,itemsep=0.45em,topsep=0.35em,parsep=0pt]
    \item \textbf{The Influence Attribution Problem} (Section~\ref{sec:credit_assignment}).\quad We identify and formalize the bottleneck that makes optimization in reputation-mediated games hard: every action or signal a learner emits alters the reputations of many third parties along combinatorially branching paths before affecting its own future rewards, and a rational learner has to simultaneously account for every such indirect feedback channel when selecting each current action.
    \item \textbf{The Reciprocity Gradient} (Sections~\ref{sec:gradient},~\ref{sec:experiments}).\quad We formulate an analytic gradient that solves influence attribution by backpropagating through multi-agent reputation dynamics, and remove the oracle assumption via private, differentiable opponent models fitted from public observations. Experiments show that the method recovers highly discriminative action and signal policies and substantially outperforms model-free baselines.
\end{enumerate}

\begin{figure}[!ht]
\centering
\includegraphics[width=\textwidth]{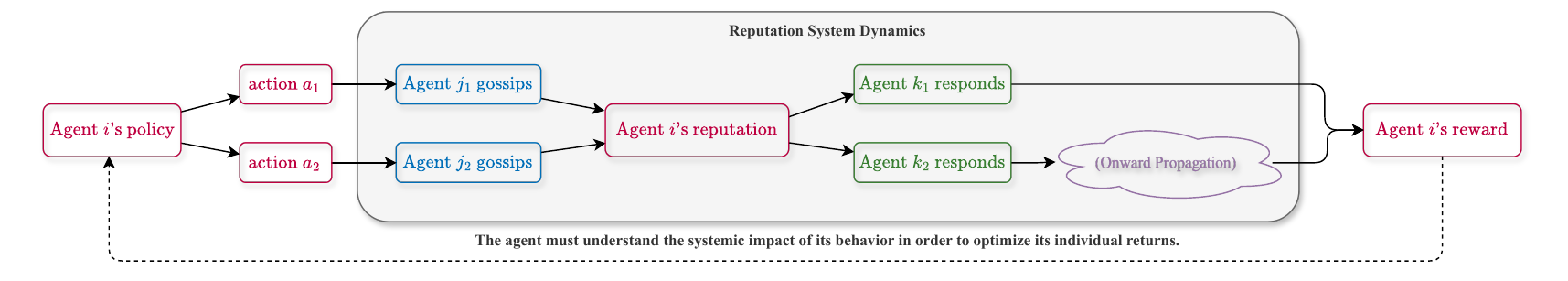}
\caption{\textbf{The influence attribution problem.}
The effects of the action will propagate throughout the entire system and eventually impact the payoff. Treating the dynamics as a black box makes it computationally intractable to update the policy.}
\label{fig:influence_attribution}
\end{figure}

%\begin{figure}[t]
%\centering
%\includegraphics[width=\textwidth]{figures/fig1_teaser.pdf}
%\caption{Reciprocity gradient: interaction flow and gradient path over two timesteps.
%Three-agent donation game with first-order gossip (refer to Definition~\ref{def:continuous_donation_game} for the formal setup). Row heading ``$x$ vs $y$'' means $x$ is the donor and $y$ is the recipient. Each row depicts one timestep $t$: the donor observes the recipient's reputation $\sigma$, applies its action policy $\pi$ (boxed) to emit an action $a^{i \to j}$ (superscript read ``from donor $i$ to recipient $j$''), and the recipient applies its signal policy $\varphi$ (boxed) to produce a new reputation entry $\sigma^{i \gets j}$ (read ``about agent $i$, produced by agent $j$'').
%}
%\label{fig:teaser}
%\end{figure}

\section{Related Work}
\label{sec:related_work}

This section positions the reciprocity gradient against two lines of work: classical indirect-reciprocity theory and reinforcement-learning approaches to reputation-mediated cooperation. A more detailed comparison with differentiable inter-agent communication, opponent-shaping methods, training-time opponent access, and related literature is deferred to Appendix~\ref{app:extended_related_work}.

\paragraph{Indirect reciprocity in multi-agent cooperation.}
\citet{nowak1998evolution,nowak2005evolution,nowak2006five} first established reputation-mediated indirect reciprocity
%\footnote{In this work, we use the terms ``indirect reciprocity'' and ``reputation system'' interchangeably, as reputation serves as the informational substrate that enables indirect reciprocity.} 
as a fundamental evolutionary mechanism in large-scale population cooperation. Without the repeated interactions required for direct reciprocity, cooperation among strangers in one-shot encounters must instead be maintained through a shared reputation system sustained by third-party centralized assessments or peer monitoring \citep{milinski2002reputation, fehr2004third, panchanathan2004indirect}. Consequently, the study of stable assessment rules evolved from unstable first-order \textbf{Image Scoring} \citep{leimar2001evolution,panchanathan2003tale} to the robust \textbf{``Leading Eight''} third-order norms \citep{ohtsuki2004should,ohtsuki2006leading}. Recent work further extends the classical setting to private and noisy assessment to capture the complexities of real-world social interactions \citep{hilbe2018indirect,fujimoto2023evolutionary,schmid2023quantitative}. 
These classical analyses characterise the evolutionary stability of assessment rules: they ask whether a given social norm resists invasion under replicator dynamics or related selection pressures. We address a complementary question. Given a fixed population of opponents, can a single learning agent compute its best response purely from gradient ascent on its own return? We accordingly treat the leading-eight norms ``Simple Standing'' (L3) and ``Stern Judging'' (L6) (\citealt{ohtsuki2006leading}) as fixed test-bed opponents rather than as the object of stability analysis, and the resulting learner recovers highly discriminative action and signal policies.

\paragraph{Reputation systems with reinforcement learning.}
Leveraging the rich representational capacity of Reinforcement Learning (RL) in complex environments \citep{mnih2015human, leibo2017multi}, recent studies have begun replacing classical rule-based strategies with learned reputation policies. In \citet{anastassacos2021cooperation}, the authors developed Q-learning algorithms with public reputation labels to enhance state information in the donation game. They found that cooperation could not naturally arise from scratch without seeding the population with fixed-strategy agents or introducing an introspective bonus. \citet{vinitsky2023learning} utilized public sanctions based on reputation states as a shared signal to train classifiers for judging behaviors, effectively creating a top-down reputation mechanism in social dilemmas. \citet{yaman2023emergence} employed decentralized sanctioning matrices to redistribute rewards, disincentivizing antisocial roles while incentivizing prosocial ones that are not intrinsically profitable. Similarly, \citet{ren2025bottom} (LR2) rescaled environmental rewards by the agent's reputation to shape cooperative policies. However, such prior approaches typically rely on environmental customizations or heuristic reward designs, which bypass the core difficulty that reputation-mediated games impose on a learner: the causal chain from an action to future rewards is routed through the population, where the action is first assessed by third-party gossip and subsequently shifts the behavioural policies of future partners (we formalize this as the \emph{influence attribution} problem in Section~\ref{sec:credit_assignment}). Sample-based estimators (score-function and temporal-difference) do not exploit the analytical structure of this chain, and the resulting agents often collapse into trivial constant-output policies. To resolve this, we introduce the \textbf{reciprocity gradient}, which captures these influence links by backpropagating through differentiable surrogates of the opponents. Unlike prior work, our learner discovers discriminative policies driven entirely by the analytic gradient of the unmodified donation-game reward, with no intrinsic bonus, reward shaping, or seed-strategy intervention.

\section{Problem Setup}
\label{sec:problem_setup}

This section formalizes the continuous random-matching donation game and states the learning objective.

At each timestep a matching rule pairs two agents into asymmetric roles, a \textit{donor} and a \textit{recipient}. The donor observes the recipient's reputation and chooses a continuous contribution level, paying a personal cost in exchange for a multiplied benefit to the recipient. The recipient then emits a continuous gossip signal that appends to the donor's reputation history and shapes how future partners assess them. We use directional superscripts throughout: $a^{i \to j}_t$ reads ``donor $i$ acts toward recipient $j$ at step $t$,'' and $\sigma^{i \gets j}_t$ reads ``recipient $j$ emits a signal about donor $i$ at step $t$.'' Formally:
\begin{definition}[Continuous Random-Matching Donation Game with Post-Gossip]
\label{def:continuous_donation_game}
A game $\mathcal{G} = \big(\mathcal{N},\, T,\, \mathcal{A},\, \Sigma,\, (\mathcal{O}_i)_{i \in \mathcal{N}},\, b,\, c\big)$ consists of a finite population $\mathcal{N}$ of $N$ agents, horizon $T \in \mathbb{N} \cup \{\infty\}$, donor action space $\mathcal{A} = [0, 1]$, signal space $\Sigma = [0, 1]$, per-agent observation spaces $\mathcal{O}_i$, benefit multiplier $b \in (1, \infty)$, and donation cost $c \in (0, b)$. Each agent $i$ has an action policy $\pi^i: \mathcal{O}_i \to \mathcal{A}$ and a gossip policy $\varphi^i: \mathcal{O}_i \to \Sigma$. At each step $t$, a matching rule $\xi$ draws a donor $i$ and a recipient $j \neq i$. The donor first chooses an action $a^{i \to j}_t = \pi^i(o^i_t) \in \mathcal{A}$; the recipient then emits a signal $\sigma^{i \gets j}_t = \varphi^j(o^j_t) \in \Sigma$ about that action; instantaneous rewards $r^i_t = -c\, a^{i \to j}_t$ and $r^j_t = b\, a^{i \to j}_t$ accrue to the donor and recipient. Each emitted signal is appended to the donor's history, $\boldsymbol{\sigma}^i_t = \big(\sigma^{i \gets k_1}_{t_1}, \dots, \sigma^{i \gets k_m}_{t_m}\big)$ with $t_1 < \cdots < t_m \le t$, and the reputation score is $s^i_t = f(\boldsymbol{\sigma}^i_t)$ for a differentiable aggregator $f$ (raw history in the main text; variants in Appendix~\ref{app:om_engineering}).
\end{definition}

The environment is defined by the following structural configurations:

\begin{description}[leftmargin=1.5em, style=sameline]
	\item[\textbf{Action and signal semantics.}] The continuous action space $\mathcal{A} = [0, 1]$ represents the magnitude of cooperation, where $0$ corresponds to absolute defection (zero donation) and $1$ represents full cooperation (maximum donation). The continuous signal space $\Sigma = [0, 1]$ operates as a pure communication channel devoid of pre-programmed semantics. While classical models of canonical social norms, such as the leading eight, explicitly hard-code the assumption that a signal of $1$ denotes a ``good'' reputation, our learning agent possesses no such prior. Instead, the interpretation of any given signal is strictly emergent: the agent successfully recovers this exact semantic mapping purely from the analytic gradients backpropagated through the evaluation dynamics of its established opponents.
    
    % \item[\textbf{Orders of reputation and individualized observation.}] Each agent $i$ maintains its own private and independent action policy $\pi^i$ and signal policy $\varphi^i$. The observation space $\mathcal{O}_i$ encapsulates the public information required by these policies (e.g., the attached signals of the opponent and the agent itself, as well as relevant historical actions), and its exact composition is dictated by the underlying order of reputation. \emph{First-order} reputation (image scoring) depends solely on the donor's current action, meaning the signal policy observes only the action; this is evolutionarily unstable~\citep{leimar2001evolution,panchanathan2003tale}. \emph{Second-order} reputation additionally evaluates the recipient's standing to enable justified punishment, expanding the observation space so the signal policy evaluates both the donor's action and the recipient's aggregated reputation score. We utilize continuous relaxations of two evolutionarily stable second-order rules from the classical ``leading eight''~\citep{ohtsuki2004should,ohtsuki2006leading}: L3 (Simple Standing) and L6 (Stern Judging). Higher-order rules yield negligible additional cooperation (reviewed in Appendix~\ref{app:ir_background}).

    	\item[\textbf{Orders of reputation and individualized observation.}] Each agent $i$ maintains its own private and independent action policy $\pi^i$ and signal policy $\varphi^i$. The observation space $\mathcal{O}_i$ encapsulates the public information required by these policies (e.g., the attached signals of the opponent and the agent itself, as well as relevant historical actions), and its exact composition is dictated by the underlying order of reputation. \emph{First-order} reputation (image scoring) depends solely on the donor's current action, so the signal policy observes only the action: $\sigma^{i\gets j}_t = \varphi^j(a^{i\to j}_t)$; this is evolutionarily unstable~\citep{leimar2001evolution,panchanathan2003tale}. \emph{Second-order} reputation additionally evaluates the recipient's standing to enable justified punishment, expanding the observation space so the signal policy evaluates both the donor's action and the recipient's aggregated reputation score: $\sigma^{i\gets j}_t = \varphi^j\!\big(a^{i\to j}_t,\, s^j_t\big)$, where $s^j_t = f(\boldsymbol{\sigma}^j_t)$ aggregates the recipient's prior signal history through a differentiable function $f$, e.g., a running mean $f_{\text{mean}} = \tfrac{1}{m}\sum_l \sigma^{j\gets k_l}_{t_l}$ or an EMA $f_{\text{EMA}} = (1-\lambda)\sum_l \lambda^{m-l}\sigma^{j\gets k_l}_{t_l}$ with decay $\lambda\!\in\!(0,1)$. We utilize continuous relaxations of two evolutionarily stable second-order rules from the classical ``leading eight''~\citep{ohtsuki2004should,ohtsuki2006leading}: L3 (Simple Standing) and L6 (Stern Judging). Higher-order rules yield negligible additional cooperation (reviewed in Appendix~\ref{app:ir_background}).
    
    \item[\textbf{Matching regime.}] Both evaluated regimes ensure perfectly balanced role assignments. In \emph{(1) direct-reciprocity allowed}, an episode concatenates a stochastic number of random round-robins over the $N(N-1)$ ordered pairs. Because pairs recur, both reputation and partner-history channels contribute to the gradient. In \emph{(2) direct-reciprocity disallowed}, an episode comprises a single round-robin. Each pair meets exactly once, isolating reputation as the sole inter-agent channel~\citep{nowak1998evolution,ohtsuki2006leading}. We denote the policy-independent matching law as $\xi$ and the role indicator as $\mathbbm{1}(i = \mathsf{D}_t, j = \mathsf{R}_t)$. Setting (E) (Appendix~\ref{sec:exp_indirect}) details this matching ablation.
    
    \item[\textbf{Reputation initialization.}] Initial reputations are drawn from three distributions: constant $0.5$, uniform $U[0, 1]$, or the empirical stationary distribution of the opponent norm (Appendix~\ref{sec:warmup}).
    
    \item[\textbf{Discrete variants.}] The continuous state and action spaces $\mathcal{A} = \Sigma = [0, 1]$ can be discretized via the Gumbel-softmax relaxation~\citep{jang2017categorical,maddison2017concrete}, albeit acting as a biased estimator.
\end{description}

\paragraph{The Objective Function.}
\label{sec:objective}

Every agent is self-interested and rational: it optimizes only its own expected cumulative reward $\mathbb{E}_\xi[\sum_t r_t^i]$, with no shared objective and no team return. Two channels feed this return: the cost paid as donor and the benefit received as recipient. The gossip policy $\varphi$ does not contribute rewards directly, but it enters indirectly through the reputation scores conditioning every donor decision; $\varphi$ shapes the present action distribution through the reputation-history channel. Our method optimizes this objective directly, with no auxiliary rewards, reward shaping, or intrinsic-motivation terms.

\section{The Reciprocity Gradient}
\label{sec:gradient}

%This section derives the reciprocity gradient and then removes its oracle assumption via differentiable opponent modeling. 

This section first identifies the influence attribution problem, then derives the reciprocity gradient to solve it, and finally removes its oracle assumption via differentiable opponent modeling.

\subsection{The Influence Attribution Problem}
\label{sec:credit_assignment}

The learning problem has a non-trivial multi-agent reputation-attribution structure: any action or signal $i$ emits propagates through a combinatorially branching tree of downstream reputation updates, touching many third parties before re-entering $i$'s own reward stream. Both policies bear on the future benefits $i$ collects and the costs $i$ pays through this coupled web (Appendix~\ref{app:influence_examples} traces concrete examples for $\pi^i$ and $\varphi^i$). Enumerating these channels by hand is hopeless. We formalize this bottleneck as the \emph{influence attribution} problem. Sample-based estimators (score-function and temporal-difference) treat this delayed feedback as opaque noise, recovering the chain only implicitly and typically collapsing into constant-output policies. To resolve this, our analytical alternative explicitly stitches the broken causal chain back together by treating the multi-agent society as an unrolled differentiable computational graph, allowing the learner to backpropagate reward gradients directly through its peers' evaluation logic. While conceptually elegant, realizing this without unrealistic oracle access to opponents' true parameters presents a severe algorithmic challenge, which we solve by dynamically fitting private, fully differentiable surrogate models purely from public observations.

% The learning problem has a non-trivial multi-agent influence-attribution structure: any action or signal $i$ emits at the current step propagates through a combinatorially branching tree of downstream signal emissions and reputation updates, touching many third parties before eventually re-entering $i$'s own reward stream, and $i$ must account for every such returning contribution on top of the immediate payoff of the current action. Both policies bear on both sides of $i$'s future payoff, the benefits $i$ collects as recipient and the costs $i$ pays as donor, through the same coupled web of reputation updates. Appendix~\ref{app:influence_examples} traces two concrete examples, one through $\pi^i$ (routed via $i$'s own reputation $s^i$) and one through $\varphi^i$ (routed via modifications to other agents' reputations). Enumerating the channels by hand is hopeless, and no fixed heuristic covers all configurations of the game. We refer to this resulting optimization problem as the \emph{influence attribution} problem. Standard model-free reinforcement learning methods have to recover this chain implicitly from noisy returns and typically collapse into constant-output policies in reputation-mediated social dilemmas; the next section introduces our approach to solving it.  

\subsection{Gradient Highway}
\label{sec:gradient_highways}

Let $\theta^i$ denote the parameters of agent $i$'s action policy $\pi^i$, and $\eta^i$ the parameters of its signal policy $\varphi^i$. To isolate the structural identity and simplify the notation, we derive the following recursion under first-order gossip, where each policy depends on a single input: the action policy reads only the recipient's reputation, $a^{j\to i} = \pi^j(s^i)$, and the gossip policy reads only the donor's action, $\sigma^{j\gets i} = \varphi^i(a^{j\to i})$. Higher-order assessment rules~\citep{ohtsuki2004should,ohtsuki2006leading} extend $\varphi^i$ with extra inputs ($s^i$, $s^j$, etc.); each adds one dependency edge to the same forward graph and is discussed later.
Under this setting, the total derivative of the expected return commutes with the expectation and reduces to an expectation over per-agent policy derivatives:
\begin{equation}
\frac{\mathrm{d}}{\mathrm{d} \theta^i} \mathbb{E}_{\xi}\left[ \sum\limits_{t=1}^T r_t^i \right]
= \frac{1}{N(N-1)}
\mathbb{E}_{\xi} \Bigg[
\sum\limits_{t=1}^T
    \Big(
    -c \cdot \sum\limits_{j\ne i} \frac{\mathrm{d}}{\mathrm{d} \theta^i}  \pi^i(s_t^j)
    + b\cdot \sum\limits_{j\ne i} \frac{\mathrm{d}}{\mathrm{d} \theta^i}  \pi^j(s_t^i) \Big)
\Bigg].
\end{equation}
The same identity holds with $\theta^i$ replaced by $\eta^i$. The expression inside $\mathbb{E}_\xi[\cdot]$ does not write $\xi$ explicitly, but the matching sequence still enters through the reputation arguments $s_t^j$ and $s_t^i$: $\xi$ determines which ordered pair acts at each step $t' \le t$, which determines what gossip signal gets emitted and into which agent's reputation slot it is appended, so the value of every $s_t^j$ along the rollout is a function of $\xi$. The total-derivative $\mathrm{d}/\mathrm{d}\theta^i$ on each term then captures the dependence of $s_t^j$ on $\theta^i$ along this same realized matching sequence, and the rest of this section unpacks it into a closed-form recursion. Appendix~\ref{app:gradient_derivation} gives the full step-by-step derivation, including the conditions that let the derivative penetrate the expectation, the role-indicator simplification of the objective, and the extension to general-sum matrix games and continuous-deterministic games.

%We derive the recursion under \emph{first-order} gossip, where each policy depends on a single input: the action policy reads only the recipient's reputation, $a^{j\to i} = \pi^j(s^i)$, and the gossip policy reads only the donor's action, $\sigma^{j\gets i} = \varphi^i(a^{j\to i})$. This setting isolates the structural identity. Higher-order extensions add input dependencies to either policy, equivalently new edges in the same forward graph, and we discuss them later.

The objective gradients $\mathrm{d}\mathbb{E}_\xi[\sum_t r_t^i] / \mathrm{d}\theta^i$ and $\mathrm{d}\mathbb{E}_\xi[\sum_t r_t^i] / \mathrm{d}\eta^i$ reduce, after pushing the derivative inside the expectation, to two families of inner total derivatives, $\mathrm{d}\pi^i(s_t^j)/\mathrm{d}\theta^i$ and $\mathrm{d}\pi^j(s_t^i)/\mathrm{d}\theta^i$ for $j \neq i$, together with the analogous pair where $\theta^i$ is replaced by $\eta^i$. The chain rule decomposes each one as
\begin{equation}
\frac{\mathrm{d}}{\mathrm{d}\theta^i}\,\pi^i(s_t^j)
=
\underbrace{\frac{\partial \pi^i(s_t^j)}{\partial \theta^i}}_{\text{local}}
\;+\;
\underbrace{\frac{\partial \pi^i(s_t^j)}{\partial s_t^j}\,\cdot\,\frac{\mathrm{d} s_t^j}{\mathrm{d}\theta^i}}_{\text{transit through reputation}}.
\label{eq:hwy-decomp-pi}
\end{equation}
For $\pi^j(s_t^i)$ the local term vanishes, since $\theta^i$ does not enter $\pi^j$ at fixed reputation. The transit term has the same shape across all four cases: a local gradient of the policy output with respect to one reputation score, times the total derivative of that reputation score with respect to the parameter.

\begin{figure}[t]
\centering
\includegraphics[width=\textwidth]{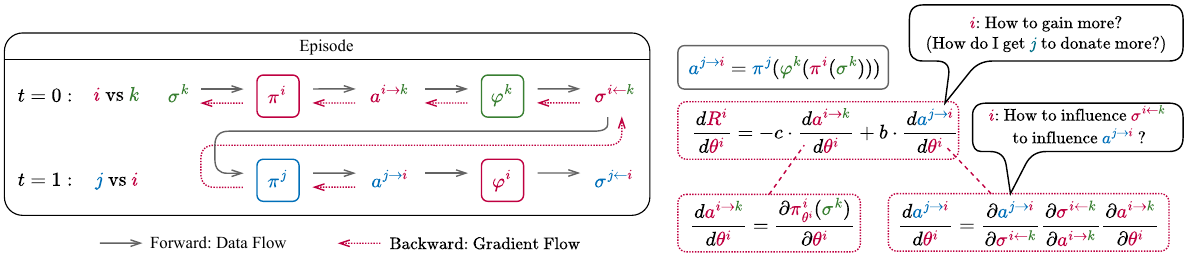}
\caption{Reciprocity gradient: interaction flow and gradient path over two timesteps.
Three-agent donation game with first-order gossip. Row heading ``$x$ vs $y$'' means $x$ is the donor and $y$ is the recipient. Each row depicts one timestep $t$: the donor observes the recipient's reputation $\sigma$, applies its action policy $\pi$ (boxed) to emit an action $a^{i \to j}$ (superscript read ``from donor $i$ to recipient $j$''), and the recipient applies its signal policy $\varphi$ (boxed) to produce a new reputation entry $\sigma^{i \gets j}$ (read ``about agent $i$, produced by agent $j$'').
}
\label{fig:teaser}
\end{figure}

\paragraph{Gradient as influence.}
The inner derivatives $\frac{\mathrm{d}}{\mathrm{d}\theta^i}\pi^i(s_t^j)$ and $\frac{\mathrm{d}}{\mathrm{d}\theta^i}\pi^j(s_t^i)$, for all $j \neq i$, are non-local. 
A change in $\theta^i$ directly alters the immediate actions emitted by agent $i$. These modified outputs subsequently perturb the interacting partners' inputs, shifting their gossip signals about agent $i$, which perturbs every downstream reputation score and in turn every future action conditioned on those scores. The derivative chains through the population along these reputation channels; Figure~\ref{fig:teaser} sketches the structure on a minimal two-timestep example. We use \emph{influence} to mean the gradient of a future action or signal involving $i$ with respect to $i$'s own parameters; computing it amounts to a single backward pass through the rolled-out interaction graph. This rationale also applies to $\eta^i$.

The parameter-to-reputation derivative $\mathrm{d} s^\ell / \mathrm{d} \theta^i$ factors through the reputation-to-reputation derivative $\mathrm{d} s^\ell / \mathrm{d} s^m$. The parameter $\theta^i$ controls $i$'s emitted actions $a^{i\to k}$; each such action drives the recipient's gossip rule, producing a signal $\sigma^{i\gets k}$ that gets appended to $s^i$. So $\theta^i$ enters the reputation graph at one node, $s^i$. The signal-policy parameter $\eta^i$ controls the signals $i$ emits about past donors $k \neq i$; each of those appends to a different $s^k$, so $\eta^i$ enters at many nodes. Write $\mathcal{E}(\theta^i) = \{i\}$ and $\mathcal{E}(\eta^i) = \{k \in \mathcal{N} : k \neq i\}$ for the entry set of each parameter. Once $\theta^i$ has entered $s^m$, $m \in \mathcal{E}(\theta^i)$, its effect on any other reputation $s^\ell$ propagates through the reputation graph and is captured by $\mathrm{d} s^\ell / \mathrm{d} s^m$. Applying the chain rule across the entry set:
\begin{equation}
\frac{\mathrm{d} s^\ell}{\mathrm{d} \theta^i}
\;=\;
\sum_{m \,\in\, \mathcal{E}(\theta^i)}\,
\underbrace{\frac{\mathrm{d} s^\ell}{\mathrm{d} s^m}}_{\text{onward propagation}}
\,\cdot\,
\underbrace{\frac{\partial s^m}{\partial \theta^i}}_{\text{entry channel}}.
\label{eq:hwy-entry}
\end{equation}

The action-policy parameter $\theta^i$ enters $s^i$ through one donor-side interaction in which $i$ is the donor and some $j \neq i$ the recipient: $i$'s action $a^{i\to j}$ depends on $\theta^i$, the recipient emits a signal $\sigma^{i\gets j}$ about $i$ whose value depends on $a^{i\to j}$, and that signal appends to $s^i$. The signal-policy parameter $\eta^i$ enters $s^k$, $k \neq i$, through one recipient-side interaction with $k$ as donor and $i$ as recipient: $i$'s emitted signal $\sigma^{k\gets i}$ depends on $\eta^i$, and the signal appends to $s^k$:
\begin{equation}
\frac{\partial s^i}{\partial \theta^i}
=
\frac{\partial s^i}{\partial \sigma^{i\gets j}}\,
\frac{\partial \sigma^{i\gets j}}{\partial a^{i\to j}}\,
\frac{\partial a^{i\to j}}{\partial \theta^i},
\qquad
\frac{\partial s^k}{\partial \eta^i}
=
\frac{\partial s^k}{\partial \sigma^{k\gets i}}\,
\frac{\partial \sigma^{k\gets i}}{\partial \eta^i}.
\label{eq:hwy-entries}
\end{equation}
The first (action-policy) chain has three factors: aggregator slot, recipient's gossip rule, and donor's action policy. The second (signal-policy) chain has two factors, because the parameter $\eta^i$ already lives in the gossip rule that produces $\sigma^{k\gets i}$, so no donor-action factor intervenes. Together with the entry-set structure ($\theta^i$ at one node, $\eta^i$ at $N{-}1$ nodes), this is the analytic origin of the asymmetric-LR protocol of Setting (A.3) (Appendix~\ref{sec:exp_groundtruth}) and of the gradient-norm gap that widens under indirect-only matching (Setting (E), Appendix~\ref{sec:exp_indirect}).

The reduction is now explicit: after one chain-rule step at the output and one entry step at the parameter, every reciprocity-gradient quantity is a linear combination of reputation-to-reputation total derivatives $\mathrm{d} s^\ell / \mathrm{d} s^m$.

\paragraph{The recursion as gradient highway.}
Fix distinct agents $i$ and $j$. Their reputation scores are produced by the aggregator $f$ acting on signal histories: $s^j = f(\boldsymbol{\sigma}^j)$, $s^i = f(\boldsymbol{\sigma}^i)$. An infinitesimal change in $s^i$ affects $s^j$ in two ways: directly, when the change reaches $s^j$ in a single hop, and indirectly, when it first transits through some third agent's reputation $s^k$ and then propagates onward. Writing this out:
\begin{equation}
\frac{\mathrm{d} s^j}{\mathrm{d} s^i}
=
\underbrace{\frac{\partial s^j}{\partial s^i}}_{\text{one hop $i \to j$}}
\;+\;
\sum_{k \,\notin\, \{i, j\}}\,
\underbrace{\frac{\partial s^j}{\partial s^k}}_{\text{one hop $k \to j$}}
\cdot
\underbrace{\frac{\mathrm{d} s^k}{\mathrm{d} s^i}}_{\text{full influence $i \rightsquigarrow k$}}
\label{eq:hwy-recursion}
\end{equation}
The partial $\partial s^j / \partial s^i$ collects every interaction in which the change in $s^i$ reaches $s^j$ without going through any other agent's reputation; we call this the \emph{one-hop connection}. The summation collects all chains of length $\geq 2$, factoring each into a final one-hop step from some $s^k$ into $s^j$ and a recursive total derivative for the prefix. Equation~\ref{eq:hwy-recursion} is exact and self-similar: the unknown on the left and the unknowns inside the sum satisfy the same equation. Repeated substitution unrolls it into a sum over all paths $i \rightsquigarrow j$ in the reputation graph, with each path contributing the product of its one-hop partial derivatives, and a single backward pass through the rolled-out interaction graph evaluates this path-sum automatically without enumerating paths.

\paragraph{One-hop closure.}
We close the recursion by deriving the one-hop partial $\partial s^j / \partial s^i$ explicitly. A one-hop change from $s^i$ to $s^j$ proceeds through one interaction in which $j$ is the donor and $i$ is the recipient: $j$ chooses an action $a^{j\to i}$ that depends on $s^i$, then $i$ emits a signal $\sigma^{j\gets i}$ about $j$ which appends to $s^j$. Under first-order gossip the signal depends only on the donor's action, so a single chain of three partials suffices:
\begin{equation}
\frac{\partial s^j}{\partial s^i}
\;=\;
\frac{\partial s^j}{\partial \sigma^{j \gets i}}
\,\cdot\,
\frac{\partial \sigma^{j \gets i}}{\partial a^{j \to i}}
\,\cdot\,
\frac{\partial a^{j \to i}}{\partial s^i}.
\label{eq:hwy-first-degree}
\end{equation}
The three factors are (i) how the reputation score $s^j$ moves when the new signal $\sigma^{j\gets i}$ is appended to its history, (ii) the donor's action's influence on the gossip $\sigma^{j\gets i}$, and (iii) the recipient's reputation $s^i$'s influence on the donor's action $a^{j\to i}$. Discretizing the action or signal space replaces the corresponding factor with its Gumbel-softmax relaxation. Equations~\ref{eq:hwy-recursion} and~\ref{eq:hwy-first-degree} close the recursion under first-order gossip.

\paragraph{Extension to higher-order policies.}
First-order gossip reads only the donor's action; second-order additionally reads the recipient's standing $s^i$; third-order also reads the donor's standing $s^j$~\citep{ohtsuki2004should,ohtsuki2006leading,santos2021complexity}. Each higher order adds one more dependency edge of $\sigma^{j\gets i}$ on a rollout variable in the same forward graph. Equation~\ref{eq:hwy-recursion} is unchanged at every order; the one-hop partial $\partial s^j/\partial s^i$ collects one extra summand $\partial s^j/\partial \sigma^{j\gets i} \cdot \partial \sigma^{j\gets i}/\partial x \cdot \mathrm{d} x/\mathrm{d} s^i$ per added dependency $x$. As long as the gossip rule is differentiable, autograd propagates the gradient through every edge automatically.

\paragraph{Per-episode update.}
Operationally, the historical dependence noted above is captured by keeping the full episode in one autograd graph: per-step rewards and reputation updates are accumulated as live tensors throughout the rollout (no \texttt{detach()} between donor steps), and a single \texttt{torch.autograd.grad} call at the end of the episode collects the contribution of every past action to the present gradient. The policy update is therefore taken once per episode rather than once per step, and one backward pass internalizes the entire reciprocity chain that Equations~\ref{eq:hwy-recursion}--\ref{eq:hwy-first-degree} describe.

\paragraph{Empirical verification of gradient chains.}
A minimal three-agent demonstration (Appendix~\ref{app:demo}) verifies the autograd computation concretely: a single \texttt{torch.autograd.grad} call on the agent's discounted return traverses all agent boundaries through the rolled-out interaction graph and matches a finite-difference estimate to float precision. The example certifies that no analytic surgery, custom backward hook, or score-function estimator is required to obtain the reciprocity gradient.

\subsection{Differentiable Opponent Modeling}
\label{sec:opponent_modeling}

Backpropagating through Equation~\ref{eq:hwy-recursion} requires that other agents' policies be differentiable functions of their inputs, and the learner typically does not see their parameters either. We therefore fit private, differentiable surrogates from public observations, a standard instantiation of opponent modeling~\citep{albrecht2018autonomous}.
Agent $i$ fits each surrogate by mean-squared regression onto its own previous interactions with $j$, without access to other agents' policy parameters or internal state: $\mathcal{L}_{\hat\pi^j_i} = \mathbb{E}_{\mathcal{B}^j_i}[(\hat\pi^j_i - a^j)^2]$ and $\mathcal{L}_{\hat\varphi^j_i} = \mathbb{E}_{\mathcal{B}^j_i}[(\hat\varphi^j_i - \sigma^j)^2]$, where $\mathcal{B}^j_i$ is $i$'s buffer of past interactions matched with $j$ and $a^j, \sigma^j$ are the action and signal $j$ emitted in each entry. The complete pseudocode is given as Algorithm~\ref{alg:opponent_modeling} in Appendix~\ref{app:om_engineering}.

\paragraph{Estimator parameterizations.}
Agent $i$ approximates the action and gossip policies of every other agent $j \neq i$ with private, differentiable estimators $\hat\pi^j_i$ and $\hat\varphi^j_i$; concrete input choices are tied to the assessment order being modeled and are deferred to Appendix~\ref{app:om_engineering}. Two parameterizations of this estimator set are admissible. The \emph{per-opponent} variant fits one pair $(\hat\pi^j_i, \hat\varphi^j_i)$ independently for every $j \neq i$, with parameter count $O(N)$ in the population size. The \emph{parameter-shared} variant replaces all $N{-}1$ pairs with a single pair $(\hat\pi(\cdot;z_j), \hat\varphi(\cdot;z_j))$ shared across opponents and indexed by a per-opponent embedding $z_j \in \mathbb{R}^d$, bringing the parameter count to $O(1)$ in $N$ and trading per-opponent specialization for parameter efficiency at scale (Setting (D), Appendix~\ref{sec:exp_scalability}).

\paragraph{Update schedule.}
At each outer iteration $i$ collects a batch of real episodes, takes $K_\text{est}$ MSE steps on the surrogate losses above, and \emph{freezes} the surrogates for the inner loop. The frozen surrogates then act as a stationary virtual environment in which the learner performs $N_\text{train}$ policy steps: every opponent function is replaced by its surrogate, recorded matching sequences are replayed through the virtual graph, and the per-episode reciprocity-gradient update of the previous subsection is taken on this virtual rollout. Replaying matching sequences introduces no extra sampling bias, since the matching rule is policy-independent (Section~\ref{sec:problem_setup}); substituting surrogates affects only the policy-dependent part of the computation. Surrogates and policy therefore alternate in epochs rather than co-evolving inside a single autograd graph, which keeps the policy gradient stationary within each inner loop. Appendix~\ref{app:om_engineering} provides the complete pseudocode (Algorithm~\ref{alg:opponent_modeling}) and surveys the remaining engineering choices (modeling buffer, update schedule, virtual-reward optimism).

\section{Experiments}
\label{sec:experiments}

The primary objective of our empirical investigation is to evaluate the reciprocity gradient as a best-response learner and to demonstrate its capacity to resolve the influence attribution problem. To avoid evaluation bias and address potential concerns regarding environment selection, we sweep over a wide range of opponent populations and learner-side configurations. Two canonical second-order leading-eight rules from the indirect-reciprocity literature, L3 (\emph{Simple Standing}) and L6 (\emph{Stern Judging})~\citep{ohtsuki2004should,ohtsuki2006leading}, anchor the canonical ends of the sweep. We also include two opponent pools, ProudCoop+AllDefector and HybridCoop+AllDefector (full definitions in Appendix~\ref{app:detailed_bestresp}), in which the analytical best response is state-dependent rather than constant-output. Across all of these we further sweep two learner-side knobs: the reputation initialization mode, and the choice of aggregator~$f$ (Section~\ref{sec:problem_setup}) that maps the signal history into a scalar reputation score. We structure the evaluation by a taxonomy of these settings based on the topological complexity of their analytical optimum.

\paragraph{Degenerate vs.\ discriminative regimes.}
Most classical reputation populations admit a constant-output analytical optimum (e.g., unconditional cooperators or defectors, or leading-eight opponents whose stationary reputation distributions make the optimal response state-independent). All evaluated methods reach the oracle-access reference on such regimes; we use them only as sanity checks (Appendix~\ref{app:detailed_bestresp}). The interesting cases are populations that require \emph{context-dependent} policies, where flat-output learners underperform. We isolate three: \textbf{action-policy optimization} (Agent~$0$ trains $\pi^0$ against two L6 opponents with a hard-coded honest signal channel), \textbf{signal-policy optimization} (Agent~$0$ trains $\varphi^0$ against ProudCoop+AllDefector with L3 assessment, where ProudCoop operates on a pride-driven mechanism: it cooperates generously when its own reputation is high, and withdraws cooperation when a low reputation signals that its prior efforts went unpraised), and \textbf{joint optimization} (Agent~$0$ trains both $\pi^0$ and $\varphi^0$ against HybridCoop+AllDefector; HybridCoop's cooperation probability scales jointly with its own and its partner's reputations, establishing a compounding positive dependency in which the learner must cultivate its own standing through a discriminative $\pi^0$ \emph{and} inflate HybridCoop's standing through a discriminative $\varphi^0$ to elicit cooperation). Full opponent definitions are in Appendix~\ref{app:detailed_bestresp}.

\begin{figure}[!t]
\centering
\includegraphics[width=\textwidth]{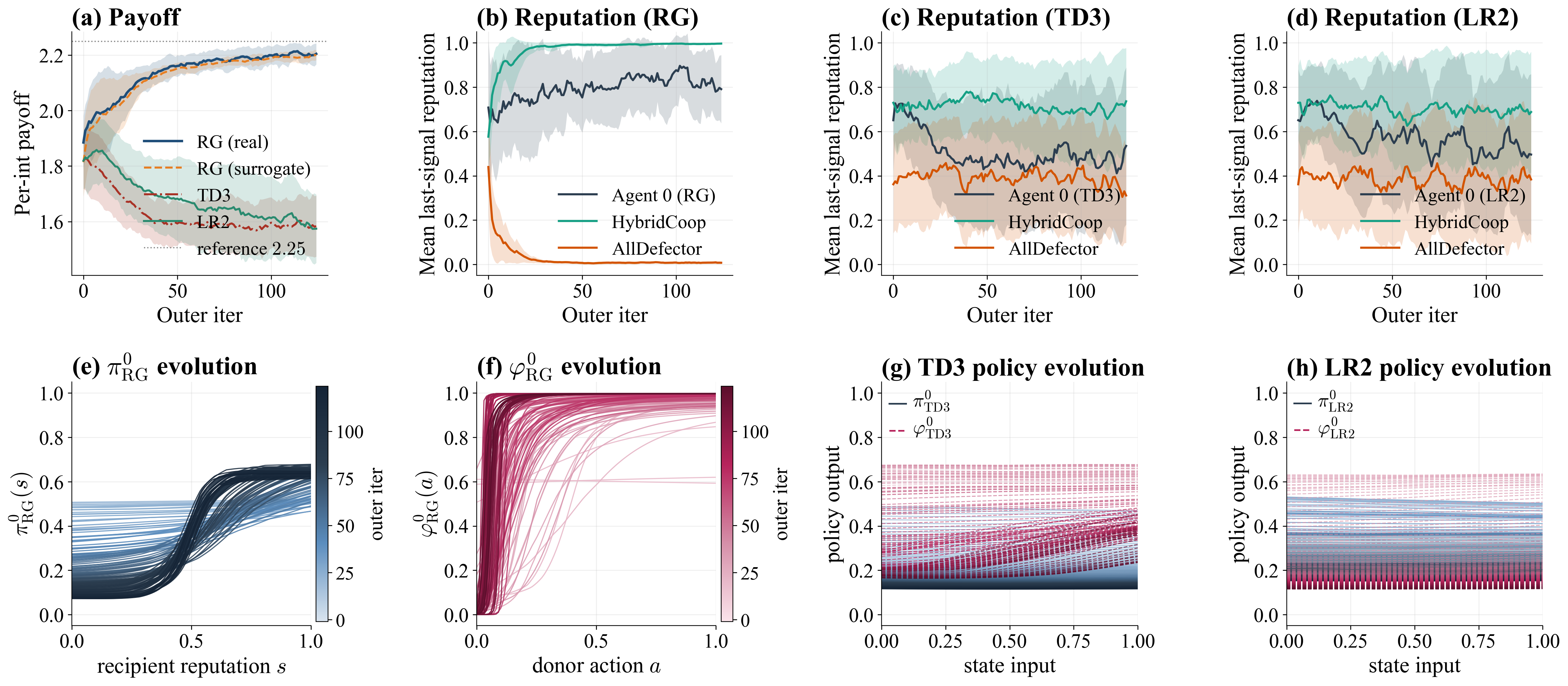}
\caption{Joint optimization against HybridCoop+AllD. RG reaches $\sim\!99\%$ of reference; TD3/LR2 collapse to flat policies. \textbf{Top:} payoff and mean per-agent reputation (bold = cross-seed mean, bands = $\pm 1$ std. \textbf{Bottom:} representative-seed network outputs across training iters (light$\to$dark = early$\to$late). Single-pathway ablations: Figure~\ref{fig:three_pathway_appendix}.}
\label{fig:three_pathway}
\end{figure}

Across all three discriminative regimes, model-free methods consistently collapse to constant-output policies, while the reciprocity gradient recovers the analytic optima. Figure~\ref{fig:three_pathway} visualizes the joint setting under observational access: neither TD3 nor LR2 learns a context-dependent action policy, both settling on near-zero cooperation across the entire reputation domain. This is the locally safe attractor that avoids the donor cost at the expense of future-reciprocity payoff.

% \begin{figure}[!t]
% \centering
% \includegraphics[width=\textwidth]{figures/fig_three_pathway_main.png}
% \caption{\textbf{Joint optimization against HybridCoop and AllDefector.} RG reaches $\sim\!99\%$ of the full-cooperation reference; TD3 and LR2 collapse to flat policies. \textbf{Top row} (a)--(d): payoff and mean per-agent reputation; thin traces are individual seeds, bold traces are the cross-seed mean (5 seeds). \textbf{Bottom row} (e)--(h): learned network outputs from a representative seed, colored by training iteration (light $\to$ dark = early $\to$ late). RG (e, f) recovers discriminative $\pi$ and $\varphi$; TD3/LR2 (g, h) stay flat. Single-pathway ablations: Figure~\ref{fig:three_pathway_appendix}.}
% \label{fig:three_pathway}
% \end{figure}

\paragraph{Mechanistic ablations and scalability.}\label{sec:exp_limits}

Beyond the joint setting in Figure~\ref{fig:three_pathway}, four sweeps stress-test the analytic gradient. \emph{(i) Indirect-only matching.} Severing direct reciprocity drops the joint payoff from $99\%$ of reference to $79\%$ ($-20$~pp) and collapses $\varphi^0$ to a constant output, while $\pi^0$ remains discriminative; the $\varphi^0$-gradient norm shrinks $5\times$ on the same transition, exactly the consequence the analysis of Section~\ref{sec:gradient} predicts when the within-episode partner-history loop into $\varphi^0$ is severed. \emph{(ii) Population-size scaling.} Across $N\in\{5,10,15,20,25,30\}$ a parameter-shared estimator with $O(1)$ parameters in $N$ keeps action-vs-L6 in the $93$--$95\%$ band, and the joint $p{=}0.5$ column tracks the $N{=}3$ anchor of $2.225$ across the entire six-fold scaling. \emph{(iii) Pool composition.} Per-interaction payoff rises monotonically with $p$ at every $N$, reaching $3.59$ at $p{=}0.8, N{=}30$ as the full-cooperation reference lifts with the cooperator fraction. \emph{(iv) Discrete-action games.} On the symmetric Prisoner's Dilemma with a Gumbel-sigmoid relaxation at the action layer, the same $\pi^0$/$\varphi^0$ networks learn a discriminative policy against both a ProudCoop+AllDefector pool and a pair of L6 opponents (per-round payoff $2.47$ in the former; in the latter the agent additionally exploits good-standing partners for the temptation payoff $T{=}5$), confirming the framework extends beyond the continuous donation game without algorithmic surgery. Full numerical results, including the indirect-only baseline comparison, the complete $(N, p)$ grid, and the discrete-action PD extension, are in Appendix~\ref{app:mechanistic_ablations} and Appendix~\ref{app:pd_discrete}.

% --- Original (longer) version of the §5.1 pointer + paragraphs, kept commented for reference ---
%Beyond the joint setting in Figure~\ref{fig:three_pathway}, we run three further sweeps to stress-test the analytic gradient: an indirect-only matching ablation that severs direct reciprocity, a population-size sweep at $N\in\{5,10,15,20,25,30\}$, and a cooperator-fraction sweep at $p\in\{0.2,0.3,0.5,0.7,0.8\}$. Removing direct reciprocity drops the joint payoff by $19$~pp and collapses $\varphi^0$ to a constant output (the $\varphi^0$-gradient norm shrinks by a factor of five), exactly the consequence the gradient analysis of Section~\ref{sec:gradient} predicts when the within-episode feedback loop into $\varphi^0$ is severed. Across the population-size and pool-composition sweep, the parameter-shared estimator (a single pair $(\hat\pi(\cdot;z), \hat\varphi(\cdot;z))$ with $O(1)$ parameters in $N$) keeps action-vs-L6 in the $93$--$95\%$ band of reference for every $N$, the joint-setting $p{=}0.5$ column tracks the $N{=}3$ anchor of $2.22$ across the entire range, and per-interaction payoff rises monotonically with $p$ at every $N$. Full numerical results, including the indirect-only baseline comparison and the complete $(N, p)$ grid, are reported in Appendix~\ref{app:mechanistic_ablations}.

\section{Conclusion}

We formulated the \emph{influence attribution problem} and the reciprocity gradient: analytic backpropagation through multi-agent reputation dynamics, paired with differentiable opponent modeling that drops the oracle-access assumption. Empirically, it recovers context-dependent policies under observational surrogates, opens substantial margins over policy-gradient and reputation-shaped baselines, and scales with population size.

\bibliography{main}
\bibliographystyle{main}

\newpage
\appendix

\section{Extended Related Work}
\label{app:extended_related_work}

This appendix expands Section~\ref{sec:related_work} with more detailed discussion of learning-based and analytic approaches to reputation-mediated cooperation. We emphasize the assumptions each line of work places on agents, in order to make explicit where our heuristic-free formulation differs.

\paragraph{Differentiable inter-agent communication.}
\citet{foerster2016learning} established the precedent of differentiable inter-agent communication (DIAL), demonstrating that continuous messages passed during training create a usable gradient channel. However, DIAL is fundamentally restricted to fully cooperative settings optimizing a shared team return. \citet{lin2023information} extended this to general-sum games by formalizing the \emph{signaling gradient}, allowing a sender to optimize its individual utility by differentiating through a receiver's response. Our \emph{reciprocity gradient} belongs to this family but introduces a critical structural departure: instead of a single-hop (sender-to-receiver) link, we backpropagate through a \emph{multi-hop reputation chain}. An agent optimizes its behavior to shape a second agent's assessment (gossip), strategically steering how a third agent will treat it in future interactions, thereby capturing the native computational structure of indirect reciprocity. Finally, unlike DIAL's simultaneous training, our differentiable opponent modeling relies strictly on public observations, completely avoiding the need for oracle access to others' parameters.

\paragraph{Access to opponents.}
While centralized training with decentralized execution (CTDE)~\citep{lowe2017multi} permits training-time access to opponents' parameters, this assumption is justifiable only in fully cooperative settings. In general-sum games, opponents possess independent objectives, rendering their strategies inherently private. To enforce strict decentralization without privileged channels, our method adopts differentiable opponent modeling~\citep{albrecht2018autonomous}. The learning agent fits private surrogate models strictly from publicly observable interactions. Consequently, the reciprocity gradient propagates analytically through these surrogates, completely eliminating the need for shared parameters, joint gradients, or any centralized training pipeline.

\paragraph{Reinforcement learning for reputation emergence.}
Two recent studies apply reinforcement learning directly to indirect-reciprocity games. \citet{anastassacos2021cooperation} train RL agents in a donation game with public reputation labels and decompose the problem into two coupled coordination tasks: agents must learn both how to interpret existing reputations and which social norm to assign reputations with. Empirically, cooperation fails to emerge from scratch in their base setting; they propose two remedies: (i) \emph{seed agents}, a fixed-strategy minority that steers the rest of the population toward a good equilibrium, and (ii) \emph{introspective rewards}, an intrinsic payoff proportional to how well an agent's own strategy performs against a copy of itself. This is a strong negative signal for end-to-end reputation learning without auxiliary scaffolds. \citet{ren2025bottom} propose LR2, a bottom-up scheme in which the training reward shapes the environmental payoff by the agent's own reputation. The published code\footnote{\url{https://github.com/itstyren/LR2}.} ships two branches and uses the additive one as default ($\texttt{repu\_reward\_coef}{=}4$):
\begin{equation*}
r^i = \begin{cases}
r_{\text{env}}^i + \kappa\, P^i_t & \text{if } \kappa \ge 1 \quad \text{(default, } \kappa{=}4\text{)}, \\
\kappa\, r_{\text{env}}^i + (1{-}\kappa)\, P^i_t\, r_{\text{env}}^i & \text{if } \kappa < 1 \quad \text{(matches Eq.~5 of the paper)},
\end{cases}
\end{equation*}
where $P^i_t = \alpha P^i_{t-1} + (1{-}\alpha)\,\mathrm{mean}(\text{neighbour assessments})$ is an EMA reputation with $\alpha{=}0.5$ in the repo. The paper's exposition foregrounds the multiplicative branch, but the experimental default is the additive one; we follow the repo default in our reproduction. LR2 reports cooperation in matrix-game social dilemmas. The scaffold, however, hard-codes precisely the claim the paper intends to demonstrate, namely that higher reputation is beneficial, by wiring reputation into the reward. Our reciprocity gradient differs from both methods in a sharp way: the learner's reward is the unmodified donation-game payoff, the gossip signal carries no semantics at initialization, and no intrinsic reward, seed agent, or reputation-shaped reward is used. Cooperation, discriminative action rules, and informative signal schemes all arise from analytic gradient propagation through reputation alone.

\paragraph{Analytic and evolutionary models of indirect reciprocity.}
The canonical analyses of indirect reciprocity treat strategies as fixed lookup tables and study their evolution under replicator dynamics or related dynamics~\citep{nowak1998evolution,nowak2005evolution,ohtsuki2004should,ohtsuki2006leading}. \citet{santos2021complexity} provides a recent synthesis of this line, emphasizing the cognitive demands of reputation-based cooperation and the role of social-norm complexity. \citet{masuda2012coevolution} analyze a two-role trust game in which buyers choose whether to trust and sellers choose whether to cooperate, with sellers carrying a reputation score that buyers update at no cost. They show under replicator dynamics that zero-cost evaluation can still sustain an evolutionarily stable cooperative equilibrium coexisting with an asocial equilibrium, a density-dependent non-matrix structure. \citet{hilbe2018indirect} relax the classical public-assessment assumption and analyze indirect reciprocity under private, noisy, and incomplete information, identifying regimes where the leading-eight rules lose stability once observations diverge across agents. Our formulation is complementary rather than competing: instead of enumerating discrete strategies and testing stability, we keep actions and signals continuous and differentiable and treat the leading-eight rules (L3, L6) as hard-coded \emph{opponents} against which a learning agent performs best response.

\paragraph{LLM-driven gossip and indirect reciprocity.}
A complementary line of work replaces hand-crafted strategies with pretrained large language models that play repeated donation games and emit free-form gossip. \citet{zhu2026talk} place self-interested LLM agents in a population that talks, judges, and cooperates, and find that natural-language gossip can sustain cooperation when each agent reads the messages of others before deciding how to act. The cooperative behavior in that setting rides on social-reasoning capacity already encoded in the pretrained model, so the experiments isolate what gossip buys \emph{given} a competent reasoner rather than what a tabula-rasa learner can recover from payoffs alone. Our setting is the strict inversion: the action and signal networks are randomly initialized, the gossip channel carries a single scalar with no a priori semantics, and the only training signal is the donation-game payoff. The two formulations therefore answer different questions: \citet{zhu2026talk} ask whether linguistic gossip can coordinate already-capable agents, while we ask whether reputation-mediated cooperation can be \emph{discovered} from gradients alone. The two perspectives are compatible, and a learning agent like ours could in principle be dropped into the LLM-population setting as a non-linguistic baseline.

\paragraph{Experimental reputation systems and gossip.}
A behavioral-economics literature studies how human participants use reputation and gossip in laboratory trust or donation games. \citet{sommerfeld2007gossip} show that gossip can substitute for direct observation in indirect-reciprocity games: participants condition behavior on third-party reports even when those reports are noisy. \citet{fehr2019gossip} find that allowing a third-party observer to send a free-form gossip message about a trustee raises trust and trustworthiness relative to a no-gossip baseline, though a substantial share of the effect is attributable to the mere presence of an observer rather than to message content. These experiments motivate the structural role that gossip plays in our setting (one agent rates another after interacting, and that rating is read by a future partner) but do not prescribe a learning rule. We inherit the same interaction schema and ask what a gradient-based learner can recover inside it without being told that gossip is valuable. Earlier reputation-system experiments such as \citet{keser2003experimental} likewise establish that reputation mechanisms raise efficiency in repeated trust games; this line informs the setting rather than the algorithmic formulation.

\section{Background: Indirect Reciprocity, Gossip, and the Leading Eight}
\label{app:ir_background}

This appendix restates the evolutionary-game-theory background we assume in the main text. It is not new material; it is included for readers less familiar with the indirect-reciprocity literature, and to fix the exact notions of ``first-order,'' ``second-order,'' and ``leading-eight'' that we rely on.

\paragraph{Direct and indirect reciprocity.}
Two mechanisms sustain cooperation among self-interested agents. In \emph{direct reciprocity}, two agents interact repeatedly and condition their behavior on their shared history; an agent cooperates because the same partner may reciprocate later~\citep{trivers1971evolution}. In \emph{indirect reciprocity}, agents rarely meet the same partner twice, so cooperation propagates through reputation: \emph{third parties} reward an agent for cooperative behavior they observed earlier~\citep{alexander2017biology,nowak1998evolution,nowak2005evolution}. Indirect reciprocity therefore requires a mechanism for transmitting information about past behavior across the population.

\paragraph{Gossip and reputation.}
The transmission mechanism is \emph{gossip}: after observing a donor's action, the recipient (or a bystander) emits an evaluative signal about the donor. This signal joins the donor's \emph{reputation}, a public record that future partners consult when deciding whether to cooperate. Two intertwined decisions form the strategic core of indirect reciprocity: how to \emph{act} given a partner's reputation (the action policy), and how to \emph{assess} a donor's behavior (the gossip policy). Both decisions simultaneously determine the population's reputational dynamics.

\paragraph{First-order reputation (image scoring).}
Under first-order assessment, the gossip signal depends only on the donor's action: cooperation produces a good signal, defection a bad one. This \emph{image scoring} rule~\citep{nowak1998evolution} is evolutionarily unstable. It cannot distinguish justified defection (refusing to help a known defector) from unjustified defection (refusing to help a cooperator), leaving it vulnerable to invasion by unconditional cooperators and subsequent exploitation by defectors~\citep{leimar2001evolution,panchanathan2003tale}.

\paragraph{Second-order reputation and the leading eight.}
Second-order assessment lets the gossip signal condition on both the donor's action \emph{and} the recipient's reputation, enabling \emph{justified punishment}: a donor who defects against a bad-standing recipient retains a good reputation. \citet{ohtsuki2004should,ohtsuki2006leading} enumerated the second-order rules that are simultaneously stable, self-consistent, and resistant to invasion by defectors, and identified exactly eight. These \emph{leading eight} norms represent the minimal sufficient complexity for robust indirect reciprocity in classical models.

\begin{figure}[ht]
\centering
\begin{minipage}[t]{0.48\textwidth}
{\centering\includegraphics[width=\linewidth]{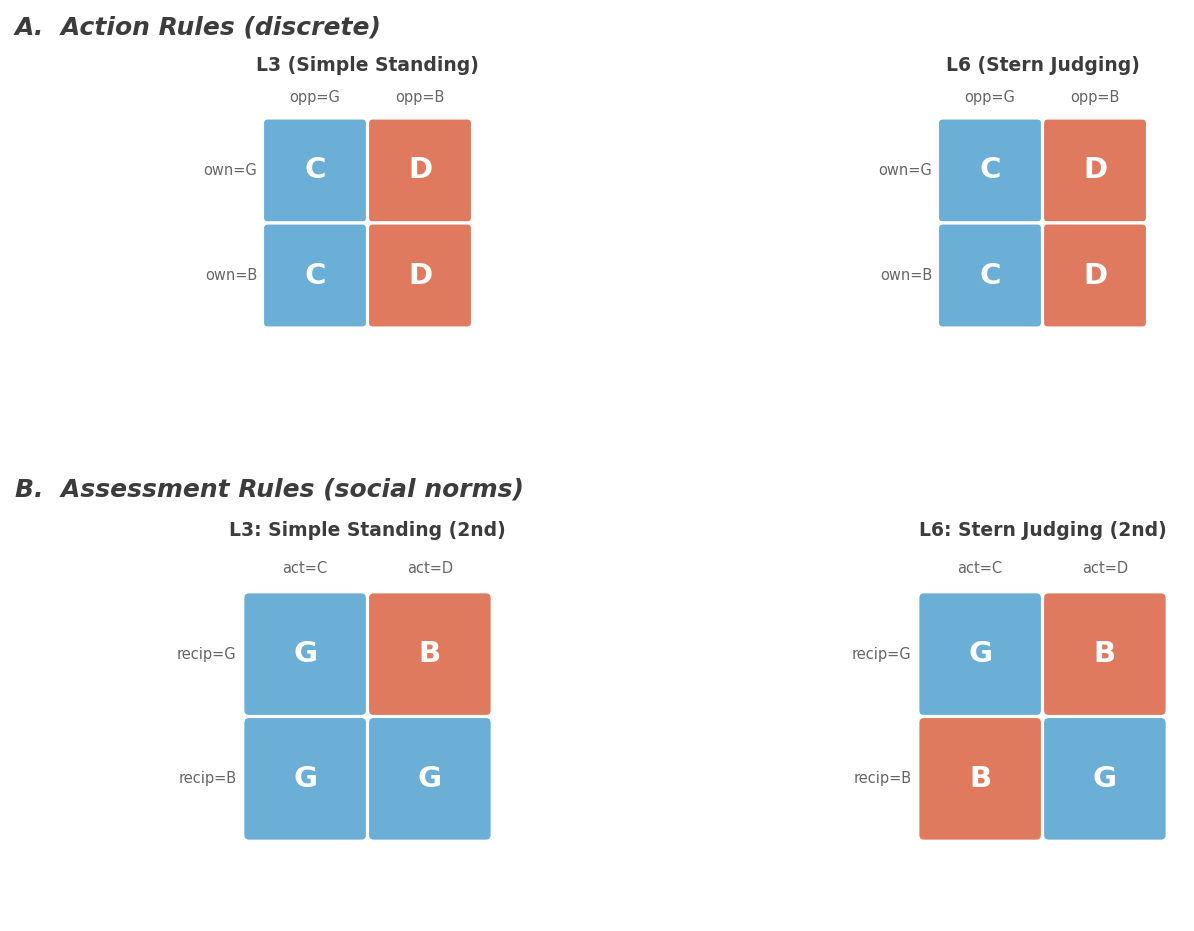}\par}
\vspace{0.4em}
{\small (a) Discrete tables. \textbf{Top:} action rule $\pi(s_\text{own}, s_\text{opp}) \in \{C, D\}$. \textbf{Bottom:} second-order assessment rule $\varphi(a_\text{donor}, s_\text{recip}) \in \{G, B\}$. The two norms differ only in the $(C, \text{Bad}) \to B$ vs.\ $G$ entry of the assessment rule.\par}
\end{minipage}\hfill
\begin{minipage}[t]{0.48\textwidth}
{\centering\includegraphics[width=\linewidth]{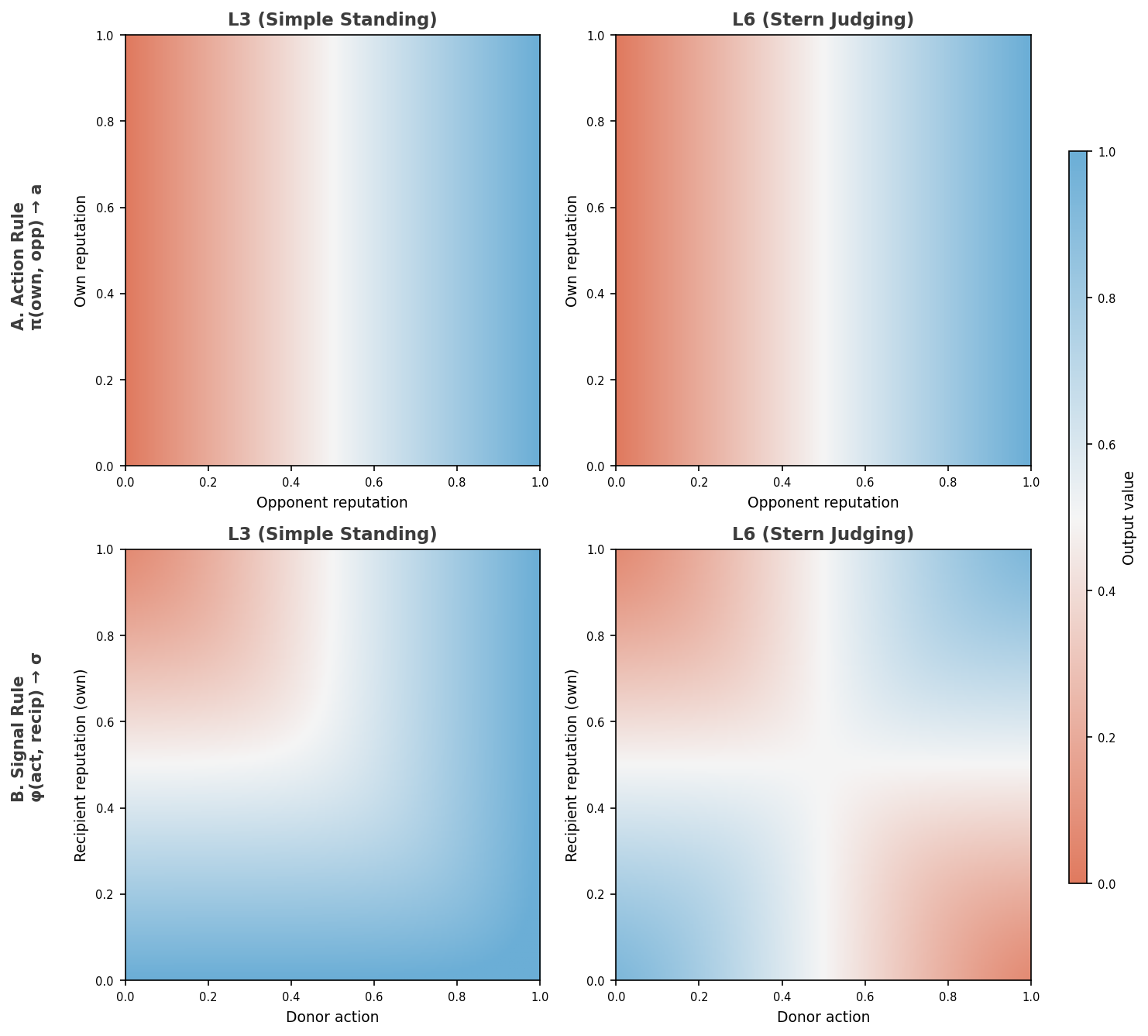}\par}
\vspace{0.4em}
{\small (b) Continuous tanh-relaxations used as differentiable opponents. \textbf{Top row:} action rule output as a function of $(s_\text{opp}, s_\text{own}) \in [0,1]^2$. \textbf{Bottom row:} signal rule output as 2a function of $(a_\text{donor}, s_\text{recip}) \in [0,1]^2$. The L6 quadrant flip is visible as a smooth diagonal switch in the bottom-right panel.\par}
\end{minipage}
\caption{\textbf{L3 (Simple Standing) and L6 (Stern Judging) at a glance.} Discrete rule tables (panel a) and the continuous tanh-relaxations actually used by the differentiable opponents (panel b). The two assessment rules agree on three of four cells; the single disagreement, the cell highlighted by the diagonal switch in L6's bottom-right heatmap, drives the entire qualitative difference between the two norms.}
\label{fig:l3_l6_overview}
\end{figure}

\paragraph{L3 (Simple Standing) and L6 (Stern Judging).}
The experiments in the main text use two specific leading-eight rules as test-bed opponents (Figure~\ref{fig:l3_l6_overview} for the discrete tables and the continuous tanh-relaxations). A second-order assessment is determined by the pair $(a, s_r)$, where $a \in \{C, D\}$ is the donor's action and $s_r \in \{\text{Good}, \text{Bad}\}$ is the recipient's standing. The two rules differ in a single setting:
\begin{center}
\begin{tabular}{c|cc|cc}
            & \multicolumn{2}{c|}{\textbf{L3 (Simple Standing)}} & \multicolumn{2}{c}{\textbf{L6 (Stern Judging)}} \\
$s_r \backslash a$ & C & D & C & D \\
\midrule
Good (G)    & G & B & G & B \\
Bad (B)     & G & G & B & G \\
\end{tabular}
\end{center}
L3 is \emph{forgiving}: every action toward a bad-standing recipient earns a good signal, so justified defection against a known defector is rewarded. L6 (\emph{Stern Judging})~\citep{ohtsuki2004should,ohtsuki2006leading} is \emph{strict}: cooperating toward a bad-standing recipient itself attracts a bad signal, in addition to defecting against a good-standing recipient. The L6 entry $(C, \text{Bad}) \to B$ is a quadrant flip, in which the cooperation assessment changes sign with the recipient's standing, and any sufficiently expressive opponent surrogate has to recover this from observations. The two rules anchor opposite ends of the rigor--leniency spectrum within the leading eight. We use L6 in the action-network setting (Setting (A.1), Setting (B.1)) because its quadrant structure carves a rectangular decision boundary into reputation space and so admits a sharp discriminative best response. We use L3 in the signal-network setting (Setting (A.2), Setting (B.2)) because its forgivingness preserves the reciprocity-gradient channel through the learner's justified-defection move, which a strict norm would destroy. The continuous, differentiable relaxations of these discrete tables that we use as opponents in practice are derived in Appendix~\ref{subsec:continuous_norms}.

\section{Concrete Influence-Channel Examples}
\label{app:influence_examples}

This appendix expands the two concrete examples of influence propagation referenced in Section~\ref{sec:credit_assignment}. Each example traces a single channel through the coupled web of reputation updates that links a current action or signal of agent $i$ to some future payoff of $i$. The examples are representative rather than exhaustive: the two channels are not separable, both policies simultaneously affect both sides of $i$'s future payoff (benefits collected as recipient and costs paid as donor) through the same web, and in any realistic network these channels multiply combinatorially with one another and with many more channels not traced here.

\subsection{Influence Through the Action Policy \texorpdfstring{$\pi^i$}{pi\^{}i}}

Agent $i$'s own reputation score $s^i$ is what future opponents condition on when deciding how to act toward $i$. Taking other agents' action and signal policies as fixed, the only handle $i$ has on the values they read off $s^i$ is therefore $s^i$ itself. In turn, $s^i$ is determined by the sequence of signals $\sigma^{i \gets j}$ that past recipients $j$ have emitted about $i$. Each such signal is the output of the recipient's (fixed) signal policy applied to $i$'s own action in that past interaction, so the only way $i$ can have produced a different $s^i$ would have been to take a different action $\pi^i(s_t^j)$ in some earlier step. A change in $\pi^i$ at the current step therefore traces the chain
\[
\pi^i \;\longrightarrow\; \sigma^{i \gets j} \;\longrightarrow\; s^i \;\longrightarrow\; \pi^k(s^i) \;\longrightarrow\; r^i_{t'},
\qquad t' > t,
\]
where $k$ is some future donor facing $i$ and $r^i_{t'}$ is the resulting payoff to $i$. The feedback to $i$'s rewards arrives in two forms. \emph{Directly,} as what the future partner $k$ decides to give $i$ when $i$ is recipient (the benefit component of $i$'s payoff). \emph{Indirectly,} through all further downstream consequences of $k$'s updated behavior: $k$'s action toward $i$ is itself observed by some future recipient, whose signal about $k$ updates $s^k$, which then conditions $i$'s own future donor decisions when $i$ eventually meets $k$ as donor, and so on. The chain's first hop is short, but the tail carries arbitrarily many further hops.

\subsection{Influence Through the Signal Policy \texorpdfstring{$\varphi^i$}{phi\^{}i}}

A structurally different channel runs through $\varphi^i$, the signal policy $i$ uses when $i$ itself is the recipient of someone else's action. Suppose donor $j$ interacts with $i$ at step $t$, and $i$ via $\varphi^i$ emits the signal $\sigma^{j \gets i}_t$ that updates $j$'s reputation score $s^j$. The modified $s^j$ now sits inside the shared reputation state of the game, and two kinds of feedback return to $i$'s future payoff from this modification.

\paragraph{Direct feedback.}
Consider any subsequent interaction that involves both $i$ and $j$. If $i$ later meets $j$ as donor, $i$'s own action toward $j$ is computed from the modified $s^j$, so $i$ can, for instance, retaliate against a stingy $j$ by donating less, which reduces $i$'s own future cost as donor. If $i$ meets $j$ as recipient again, $j$'s action toward $i$ depends on $s^i$ as usual, but the signal $j$ emits under a second-order norm depends on both the action and $s^i$, and the signals downstream recipients emit about $j$ condition on the current $s^j$; both routes propagate the original evaluation of $j$ back into interactions that affect $i$. The common feature of all these direct routes is that only one intermediate agent ($j$ itself) sits between the signal $\varphi^i$ emits and the payoff $i$ receives.

\paragraph{Indirect feedback.}
Now consider a third agent $k$ who later faces $j$ and is not $i$. Agent $k$'s action toward $j$ reads the modified $s^j$, so $k$ acts differently than it would have without the signal. The interaction between $k$ and $j$ then modifies $k$'s own reputation $s^k$ through whatever signal $j$ emits under its (fixed) $\varphi^j$. When $i$ eventually interacts with $k$, $i$'s payoff with $k$ is shaped by this downstream $s^k$, whether $i$ is the donor (acting based on $s^k$) or the recipient (receiving whatever $k$ decides to give based on $s^i$, under a second-order rule that also conditions on $s^k$). The chain here has one intermediate agent ($k$) between the original signal and the final payoff.

\paragraph{Arbitrary-depth recursion.}
The two examples above stop at one or two intermediate agents, but the mechanism is recursive with arbitrary depth. Each interaction along the chain modifies some agent's reputation, which enters the next interaction's decision, which emits a new signal that reshapes yet another reputation, which enters the interaction after that, and so on. In any realistic network of agents these chains multiply combinatorially: every action or signal $i$ emits at the current step triggers a branching tree of downstream reputation updates, many of which eventually feed back into $i$'s own future rewards through paths the two examples above do not cover. The influence attribution problem identified in Section~\ref{sec:credit_assignment} is the problem of accounting for all such paths simultaneously on top of the immediate payoff of the current action.

\newpage
\section{Backward Gradient Graph: Visualization and Numerical Verification}
\label{app:demo}

This appendix documents a minimal three-agent demo that (i) visualizes the autograd computation graph the reciprocity gradient traverses and (ii) verifies numerically that PyTorch backpropagation crosses agent boundaries without any analytic surgery.

\paragraph{Setup.}
We instantiate $N=3$ agents indexed $i=0$, $j=1$, $k=2$, each implemented as a self-contained \texttt{DemoAgent} with a tiny Markov-approximation architecture (memory window $=1$, i.e.\ observations consist only of the \emph{last} reputation entry of the target). The action net is $\mathrm{MLP}_{1\to(10+\mathrm{idx})\to 1}$ with a \texttt{Tanh} hidden layer and a \texttt{Sigmoid} output, producing an action $a \in [0,1]$. The gossip (signal) net is $\mathrm{MLP}_{1\to(20+\mathrm{idx})\to 1}$ with the same activation pattern, producing a signal $\sigma \in [0,1]$. We use first-order gossip (\texttt{second\_order}\,$=$\,\texttt{False}), so the signal net sees only the donor's action. Tiny agent-specific hidden widths (offset by agent index) make the three agents' parameter sets visually distinguishable in the rendered graph.

The environment is \texttt{ContinuousRandomMatchingDonationGame} with horizon $T=3$, benefit $b=2.0$, and a constant reputation initialization $r^{(0)}=0.5$ for every agent. We fix the random seed to $42$ and place all agents in \texttt{eval()} mode to disable \texttt{Dropout} so that the forward pass is fully deterministic and the finite-difference check below is exact to floating-point precision. The matching sequence is hard-coded to exercise all three gradient channels of interest:
\[
(t{=}0):\; i \to k, \qquad (t{=}1):\; j \to i, \qquad (t{=}2):\; i \to j.
\]
The key hyperparameters are summarized in Table~\ref{tab:demo_settings}.

\begin{table}[H]
\centering
\small
\begin{tabular}{ll}
\toprule
\textbf{Setting} & \textbf{Value} \\
\midrule
Number of agents $N$ & $3$ \\
Horizon $T$ & $3$ \\
Benefit $b$ & $2.0$ \\
Cost $c$ & $1.0$ \\
Initial reputation & constant, $r^{(0)}_m = 0.5$ \\
Signal dim & $1$ \\
Reputation memory window & $1$ (Markov approximation) \\
Action net hidden & $(10+\mathrm{idx},)$ with \texttt{Tanh}+\texttt{Sigmoid} \\
Signal net hidden & $(20+\mathrm{idx},)$ with \texttt{Tanh}+\texttt{Sigmoid} \\
Reputation order & first-order (signal $=f(a)$) \\
Matching sequence & $(0{\to}2),\,(1{\to}0),\,(0{\to}1)$ \\
Random seed & $42$ \\
Mode & \texttt{eval()} (no \texttt{Dropout}) \\
Finite-difference step $\varepsilon$ & $10^{-4}$ \\
\bottomrule
\end{tabular}
\caption{Demo configuration used to generate Figures~\ref{fig:demo_grad_a0_pi0}--\ref{fig:demo_grad_R0_pi0} and to verify backpropagation through the reciprocity graph.}
\label{tab:demo_settings}
\end{table}

\paragraph{Agent architecture.}
Listing~\ref{lst:demo_agent} shows the \texttt{DemoAgent} class used in this demo. Each agent owns an \texttt{action\_net} and a \texttt{signal\_net}, both standard MLPs with \texttt{Sigmoid} outputs. The Markov approximation reads only the last entry of the reputation history.

\begin{lstlisting}[caption={DemoAgent: minimal agent with Markov-approximation action and signal nets.},label={lst:demo_agent}]
class DemoAgent(nn.Module):
    def __init__(self, name, signal_dim=1, second_order=False):
        super().__init__()
        idx = int(name.split("_")[-1])
        # Agent-specific hidden widths for visual distinction in graph
        self.action_net = build_mlp(signal_dim, (10+idx,), 1,
                                    out_activation=nn.Sigmoid())
        signal_in = 1 + (signal_dim if second_order else 0)
        self.signal_net = build_mlp(signal_in, (20+idx,), 1,
                                    out_activation=nn.Sigmoid())
        self.register_buffer("reputation_history",
                             torch.empty((1, signal_dim)))

    def take_action(self, recipient_reputation):
        obs = recipient_reputation[-1].view(1, -1)  # Markov: last entry
        return self.action_net(obs).squeeze(0)

    def send_signal(self, donor_action, own_reputation=None):
        a = donor_action.view(1, 1)
        if self.second_order:
            inp = torch.cat([a, own_reputation[-1].view(1,-1)], dim=-1)
        else:
            inp = a
        return self.signal_net(inp).squeeze(0)

    def append_reputation(self, signal_t):
        self.reputation_history = torch.cat(
            [self.reputation_history, signal_t.view(1, -1)], dim=0)
\end{lstlisting}

\paragraph{Demo script.}
Listing~\ref{lst:demo_script} shows the core of the demo: running one episode with a fixed matching sequence, extracting the autograd gradient $\partial R^0 / \partial \theta$, and comparing against a finite-difference estimate. All quantities retain their computation graph because \texttt{ContinuousRandomMatchingDonationGame} never calls \texttt{.detach()} on intermediate rewards or signals.

\begin{lstlisting}[caption={Core demo: episode rollout, autograd gradient, and finite-difference verification.},label={lst:demo_script}]
torch.manual_seed(42)
agents = [DemoAgent(name=f"Agent_{i}") for i in range(3)]
for a in agents:
    a.eval()

env = ContinuousRandomMatchingDonationGame(
    agents=agents, horizon=3, b=2.0,
    initial_reputation_dist="constant",
    initial_reputation_params={"value": 0.5})

# Fixed matching: t=0: i->k, t=1: j->i, t=2: i->j
fixed_matching = [(0, 2), (1, 0), (0, 1)]
rewards, _, history = env.play_episode(
    matching_sequence=fixed_matching)
R0 = rewards[0]  # Agent 0's cumulative reward (with grad)

# Autograd: differentiate R0 w.r.t. one scalar weight
target = agents[0].action_net[0].weight
grad_auto = torch.autograd.grad(R0, target,
                                retain_graph=True)
g_auto = grad_auto[0][0, 0].item()  # => 0.010034

# Finite difference: perturb the same weight by epsilon
eps = 1e-4
# ... (re-create agents, load weights, perturb by +eps,
#      re-run episode, compute (R0_new - R0_base) / eps)
g_fd = (R0_plus - R0_base) / eps    # => 0.010133

error = abs(g_auto - g_fd)           # => 9.9e-05 < 1e-3
\end{lstlisting}

\begin{figure}[H]
    \centering
    \includegraphics[width=\textwidth]{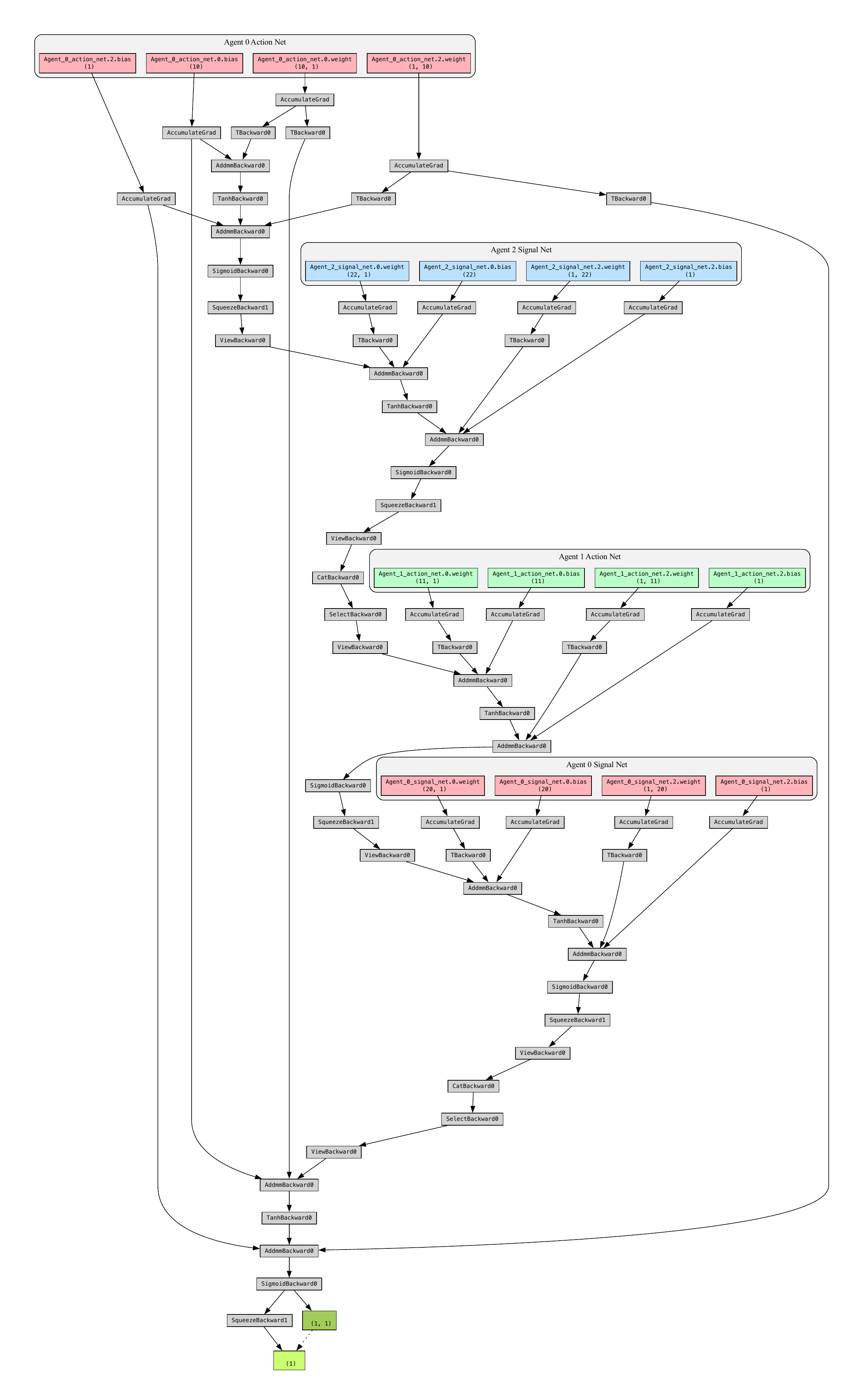}
    \caption{Computation graph of $\partial a^i / \partial \pi^i$, where $a^i \coloneq a_{0\to 1}$ is Agent~$0$'s action at $t{=}2$ and $\pi^i$ are Agent~$0$'s parameters. The backward pass traverses Agent~$0$'s signal net (gossip emitted at $t{=}1$) before re-entering Agent~$0$'s action net at $t{=}2$.}
    \label{fig:demo_grad_a0_pi0}
\end{figure}

\begin{figure}[H]
    \centering
    \includegraphics[width=\textwidth]{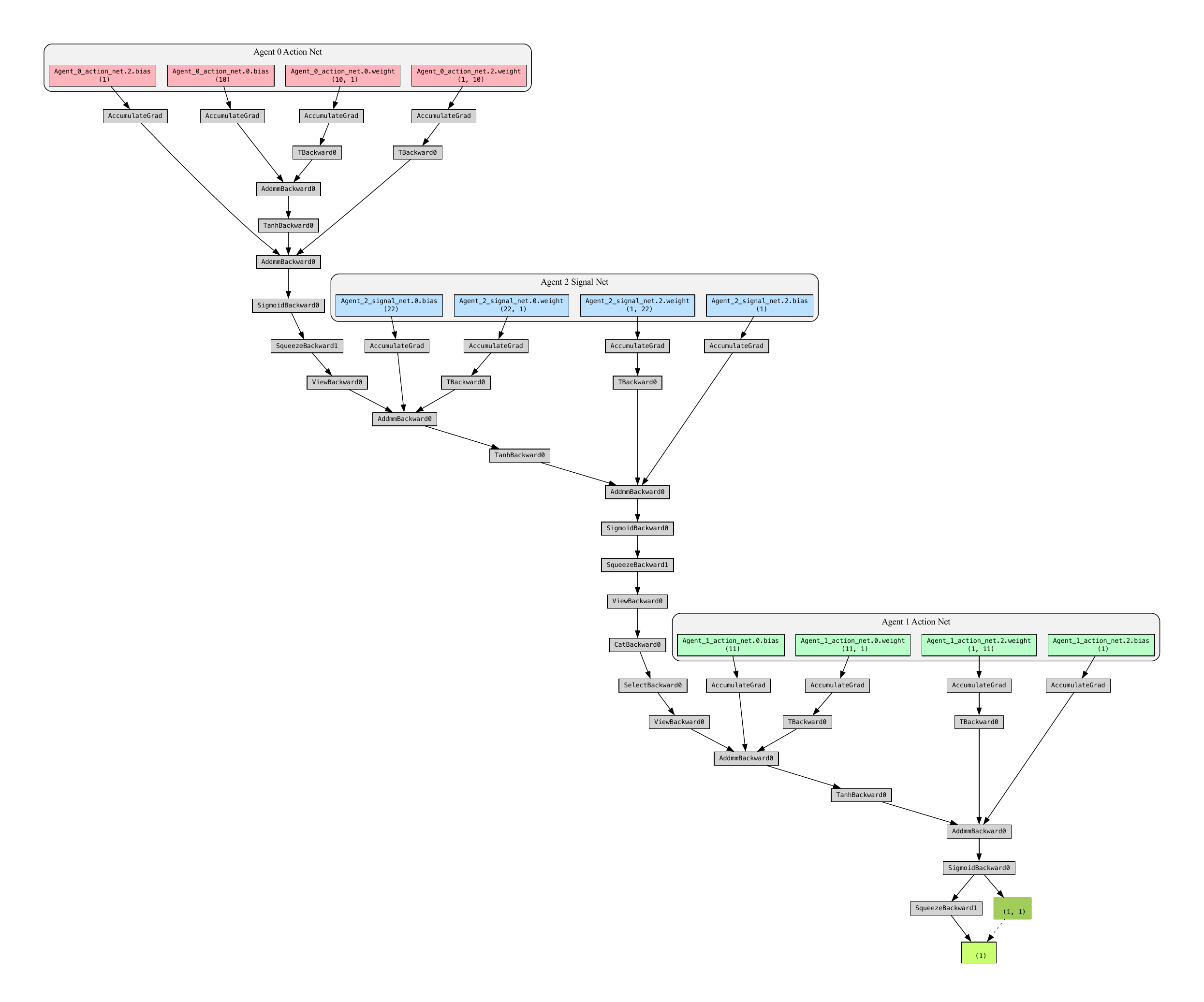}
    \caption{Computation graph of $\partial a^j / \partial \pi^i$, where $a^j \coloneq a_{1\to 0}$ is Agent~$1$'s action at $t{=}1$ and $\pi^i$ are Agent~$0$'s parameters. Agent~$0$'s parameters reach $a^j$ only through Agent~$2$'s gossip net, demonstrating that the reciprocity gradient threads across two agent boundaries via a single autograd call.}
    \label{fig:demo_grad_a1_pi0}
\end{figure}

\begin{figure}[H]
    \centering
    \includegraphics[width=\textwidth]{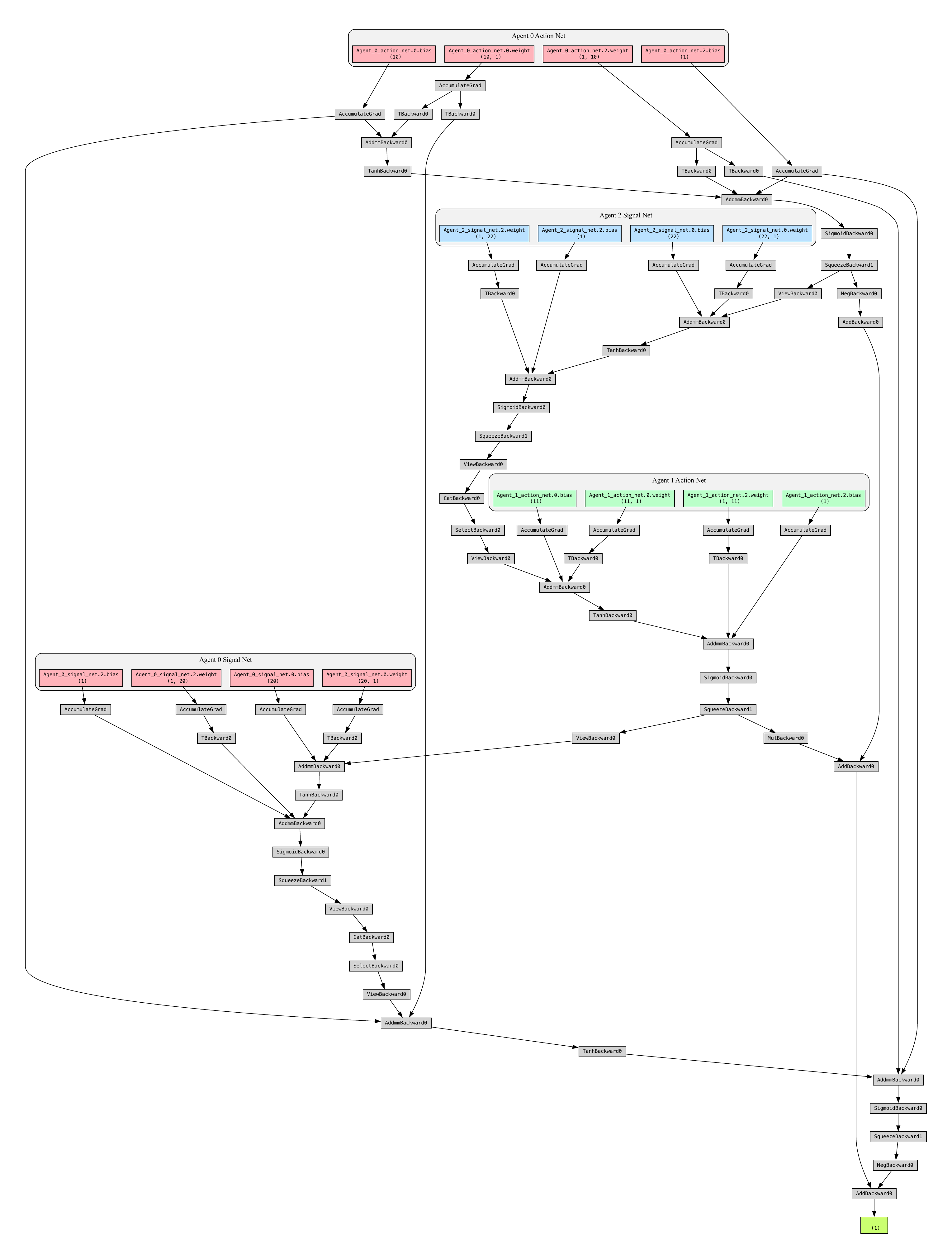}
    \caption{Computation graph of $\partial R^0 / \partial \pi^i$, i.e.\ the full reciprocity gradient for Agent~$0$ under the fixed matching $(0{\to}2),\,(1{\to}0),\,(0{\to}1)$. All five influence channels listed in Appendix~\ref{app:demo} appear as connected subgraphs sharing the three agent-colored parameter clusters.}
    \label{fig:demo_grad_R0_pi0}
\end{figure}

\paragraph{Rendered graphs.}
Figures~\ref{fig:demo_grad_a0_pi0}--\ref{fig:demo_grad_R0_pi0} show the rendered computation graphs produced by \texttt{torchviz}. We color Agent~$0$'s parameter nodes pink, Agent~$1$'s green, Agent~$2$'s blue, and group each agent's action-net and signal-net parameters in their own cluster so that the reader can visually track how the backward pass enters and exits each agent.

Figure~\ref{fig:demo_grad_a0_pi0} shows the subgraph reachable from $a^i \coloneq a_{0\to 1}$ (Agent~$0$'s \emph{own} action at $t{=}2$) with respect to $\pi^0$. The backward pass leaves Agent~$0$'s action net, traverses Agent~$2$'s gossip net to obtain $\sigma^{0\gets 2}$ (the signal that Agent~$0$ received about itself at $t{=}0$), re-enters Agent~$0$'s signal net to produce $\sigma^{1\gets 0}$ at $t{=}1$, and finally re-enters Agent~$0$'s action net at $t{=}2$. This is the simplest \emph{self-recurrent} reciprocity path: a player's own past action shapes a future input it provides to itself through the reputation channel.

Figure~\ref{fig:demo_grad_a1_pi0} shows the subgraph reachable from $a^j \coloneq a_{1\to 0}$ (Agent~$1$'s action at $t{=}1$) with respect to $\pi^0$. Unlike the previous figure, the target is an action of \emph{another agent}, and Agent~$0$'s parameters reach it through a single inter-agent path: $\pi^0 \to a_{0\to 2} \to \sigma^{0\gets 2}$ (computed by Agent~$2$'s gossip net) $\to s^0 \to a_{1\to 0}$. This is the purest demonstration of the inter-agent reciprocity signal, threading across two agent boundaries via a single autograd call with no analytic surgery.

Figure~\ref{fig:demo_grad_R0_pi0} shows the full graph for $R^0$, the object actually differentiated during training. It combines all five influence channels enumerated below: own-action terms at $t{=}0$ and $t{=}2$, the self-recurrence through Agent~$2$'s gossip, the inter-agent reciprocity through Agent~$1$'s action net, and the gossip-author channel through Agent~$0$'s own signal net at $t{=}1$. Every parameter cluster appears at least once on the backward path; visually inspecting the union of the subgraphs in Figures~\ref{fig:demo_grad_a0_pi0}--\ref{fig:demo_grad_a1_pi0} recovers the connectivity of this combined graph.

\paragraph{Cumulative reward.}
Under this matching sequence, Agent~$0$'s cumulative reward decomposes as
\[
R^0 \;=\; -\,c\,a_{0\to 2} \;+\; b\,a_{1\to 0} \;-\; c\,a_{0\to 1}.
\]
With $c=1$, $b=2$ and the random seed above, the realized quantities are
\[
a_{0\to 2}=0.5030,\;\; \sigma_{2\leftarrow 0}=0.3931,\;\; a_{1\to 0}=0.5591,\;\; \sigma_{0\leftarrow 1}=0.4655,\;\; a_{0\to 1}=0.5002,\;\; \sigma_{1\leftarrow 0}=0.5560,
\]
yielding $R^0 = -0.5030 + 1.1182 - 0.5002 = 0.1150$.

\paragraph{Parameter influence channels.}
Agent~$0$'s parameters $\pi^0$ enter $R^0$ through five distinct optimization channels, all of which must be traversed by a single call to \texttt{torch.autograd.grad}:
\begin{enumerate}
    \item \textbf{Own action at $t{=}0$.} $\pi^0$ determines $a_{0\to 2}$, which directly contributes $-c\,a_{0\to 2}$ to $R^0$.
    \item \textbf{Signal back on oneself.} $a_{0\to 2}$ is fed into Agent~$2$'s gossip net, producing $\sigma_{2\leftarrow 0}$, which is appended to Agent~$0$'s reputation history.
    \item \textbf{Future action received.} At $t{=}1$, Agent~$1$'s action net reads Agent~$0$'s updated reputation and returns $a_{1\to 0}$, contributing $b\,a_{1\to 0}$ to $R^0$.
    \item \textbf{Gossip author channel.} At $t{=}1$, Agent~$0$'s own signal net produces $\sigma_{0\leftarrow 1}$, which is appended to Agent~$1$'s reputation history.
    \item \textbf{Own action at $t{=}2$.} At $t{=}2$, Agent~$0$'s action net reads Agent~$1$'s updated reputation (which carries $\sigma_{0\leftarrow 1}$) and returns $a_{0\to 1}$, contributing $-c\,a_{0\to 1}$ to $R^0$.
\end{enumerate}
Channels (2)--(3) constitute the \emph{action-signpost} branch of the reciprocity gradient (Section~\ref{sec:gradient_highways}); channels (4)--(5) constitute the \emph{signal-signpost} branch.

\paragraph{Numerical verification.}
Using \texttt{torch.autograd.grad(R0, action\_net[0].weight)}, we read off the $(0,0)$-entry of the analytic gradient:
\[
\left.\frac{\partial R^0}{\partial\,\mathtt{action\_net[0].weight}[0,0]}\right|_{\mathrm{autograd}} = 0.010034.
\]
We then re-run the full episode with the same seed and matching sequence after perturbing the same scalar weight by $\varepsilon = 10^{-4}$, and form the finite-difference estimate $\widehat{g} = (R^0(+\varepsilon) - R^0(0))/\varepsilon$:
\[
R^0(0) = 0.114991,\qquad R^0(+\varepsilon) = 0.114992,\qquad \widehat{g} = 0.010133.
\]
The absolute error is $|0.010034 - 0.010133| = 9.9\times 10^{-5}$, below our $10^{-3}$ acceptance threshold. This certifies that the autograd graph is intact across all five cross-agent gradient channels: no custom backward hook, no score-function estimator, and no surrogate loss are required for the reciprocity gradient.

\section{Reciprocity Gradient: Full Derivation and Generality}
\label{app:gradient_derivation}

This appendix gives (i) the step-by-step donation-game derivation of the reciprocity gradient used in Section~\ref{sec:gradient}, (ii) a short survey of the game families used in the reputation literature, (iii) an account of why the donation-game choice is the canonical one for the branches we build on, and (iv) the extension of the derivation to continuous-deterministic general-sum games.

\subsection{Step-by-Step Donation-Game Derivation}
\label{app:donation_derivation}

Let $\mathcal{M}$ denote the finite sample space of matching sequences over the horizon $T$, and let $P_\xi(m)$ denote the joint probability of sequence $m \in \mathcal{M}$ under $\xi$. The role indicator at step $t$ is a function of $\xi$ alone, statistically independent of the reputation histories that drive the action evaluations, and $\xi$ itself depends on no policy parameter. Under these conditions, the total derivative with respect to $\pi^i$ can be pushed all the way inside the expectation:

\begin{equation*}
\begin{aligned}
\frac{\mathrm{d}}{\mathrm{d} \theta^i} \mathbb{E}_{\xi}\left[ \sum\limits_{t=1}^T r_t^i \right]
&= \frac{\mathrm{d}}{\mathrm{d} \theta^i}  \mathbb{E}_{\xi} \Bigg[
\sum\limits_{t=1}^T
    \begin{aligned}[t] \Big(
    &-c \cdot \sum\limits_{j\ne i}\mathbbm{1}(i=\mathsf{D}\land j=\mathsf{R})\cdot \pi^i(s_t^j)  \\
    &+ b \cdot \sum\limits_{j\ne i} \mathbbm{1}(j=\mathsf{D}\land i=\mathsf{R}) \cdot \pi^j(s_t^i) \Big)
    \Bigg]
    \end{aligned} \\
&= \frac{\mathrm{d}}{\mathrm{d} \theta^i}  \sum\limits_{t=1}^T
    \begin{aligned}[t] \Big(
    &- c \cdot \sum\limits_{j\ne i} \underbrace{\mathbb{E}_{\xi}\big[\mathbbm{1}(i=\mathsf{D}\land j=\mathsf{R})\big]}_{=\frac{1}{N(N-1)}} \cdot \mathbb{E}_{\xi}\big[ \pi^i(s_t^j) \big]  \\
    &+ b\cdot \sum\limits_{j\ne i} \underbrace{\mathbb{E}_{\xi}\big[\mathbbm{1}(j=\mathsf{D}\land i=\mathsf{R})\big]}_{=\frac{1}{N(N-1)}} \cdot \mathbb{E}_{\xi}\big[\pi^j(s_t^i)\big] \Big)
    \end{aligned} \\
&= \frac{1}{N(N-1)} \frac{\mathrm{d}}{\mathrm{d} \theta^i}
\mathbb{E}_{\xi} \Bigg[
\sum\limits_{t=1}^T
    \Big(
    -c \cdot \sum\limits_{j\ne i} \pi^i(s_t^j)
    + b\cdot \sum\limits_{j\ne i} \pi^j(s_t^i) \Big)
\Bigg] \\
&= \frac{1}{N(N-1)} \sum_{m \in \mathcal{M}} P_{\xi}(m) \frac{\mathrm{d}}{\mathrm{d} \theta^i} \Bigg[ \sum\limits_{t=1}^T \Big( -c \cdot \sum\limits_{j\ne i} \pi^i(s_t^j) +  b\cdot \sum\limits_{j\ne i} \pi^j(s_t^i) \Big) \Bigg] \\
&= \frac{1}{N(N-1)}
\mathbb{E}_{\xi} \Bigg[
\sum\limits_{t=1}^T
    \Big(
    -c \cdot \sum\limits_{j\ne i} \frac{\mathrm{d}}{\mathrm{d} \theta^i}  \pi^i(s_t^j)
    + b\cdot \sum\limits_{j\ne i} \frac{\mathrm{d}}{\mathrm{d} \theta^i}  \pi^j(s_t^i) \Big)
\Bigg]
\end{aligned}
\end{equation*}

Equality~(i) substitutes the role-indicator form of the per-step reward from Section~\ref{sec:objective}. Equality~(ii) moves the expectation inside the per-step sums and factors each term as $\mathbb{E}_\xi[\mathbbm{1}\cdot\pi] = \mathbb{E}_\xi[\mathbbm{1}]\cdot\mathbb{E}_\xi[\pi]$ by the independence of the indicator from the reputation histories, with the indicator expectation evaluating to $1/N(N-1)$ for every ordered pair. Equality~(iii) collapses the constant prefactor in front of the sum and repackages the per-step terms back inside a single expectation. Equality~(iv) expands the expectation explicitly over the matching space $\mathcal{M}$ and moves the total derivative inside the finite sum, valid because $P_\xi(m)$ is policy-independent. Equality~(v) repackages the sum over $\mathcal{M}$ back as an expectation. The same derivation carries through verbatim when $\theta^i$ is replaced by $\eta^i$, yielding the signal-policy gradient.

\subsection{Detailed Gradient Expansion}
\label{sec:gradient_expansion}

Composing the chain-rule decomposition at a policy output with the reputation-to-reputation highway derivatives from Section~\ref{sec:gradient_highways} yields closed forms for the four parameter gradients that appear in the main-text objective. Throughout this subsection we factor out two \emph{entry-channel} terms that capture how the parameter first enters the reputation graph:
\begin{align*}
H_{\theta^i}
&\;:=\; \sum_{k \ne i}
\frac{\partial s^i}{\partial \sigma^{i \gets k}}
\cdot \frac{\mathrm{d} \sigma^{i \gets k}}{\mathrm{d} a^{i \to k}}
\cdot \frac{\partial a^{i \to k}}{\partial \theta^i},
&
H_{\eta^i}^{k}
&\;:=\;
\frac{\partial s^k}{\partial \sigma^{k \gets i}}
\cdot \frac{\partial \sigma^{k \gets i}}{\partial \eta^i}.
\end{align*}
$H_{\theta^i}$ is a scalar channel that routes $\theta^i$ into $i$'s own reputation $s^i$ via every signal past recipients have emitted about $i$; $H_{\eta^i}^{k}$ is a per-agent channel that routes $\eta^i$ into agent $k$'s reputation $s^k$ via the single signal $i$ emitted about $k$.

\paragraph{Action parameter \texorpdfstring{$\theta^i$}{theta\^{}i}.}
The parameter $\theta^i$ governs only $i$'s outgoing actions and reaches others only through $s^i$:
\begin{align}
\frac{\mathrm{d} \pi^i(s_t^j)}{\mathrm{d} \theta^i}
&= \underbrace{\frac{\partial \pi^i(s_t^j)}{\partial \theta^i}}_{\text{local}}
\;+\; \underbrace{\frac{\partial \pi^i(s_t^j)}{\partial s_t^j} \cdot \frac{\mathrm{d} s_t^j}{\mathrm{d} s^i} \cdot H_{\theta^i}}_{\text{via $i$'s reputation}},
\label{eq:expansion-theta-own} \\
\frac{\mathrm{d} \pi^j(s_t^i)}{\mathrm{d} \theta^i}
&= \frac{\partial \pi^j(s_t^i)}{\partial s_t^i} \cdot \frac{\mathrm{d} s_t^i}{\mathrm{d} s^i} \cdot H_{\theta^i}.
\label{eq:expansion-theta-other}
\end{align}
Equation~\ref{eq:expansion-theta-own} describes how $i$'s own future action toward $j$ shifts; the first summand is the direct response of $\pi^i$ at fixed reputation, the second is the indirect response through $i$'s own reputation chain. Equation~\ref{eq:expansion-theta-other} describes how a third party $j$'s future action back toward $i$ shifts, which happens solely through $i$'s reputation.

\paragraph{Signal parameter \texorpdfstring{$\eta^i$}{eta\^{}i}.}
The parameter $\eta^i$ governs only the signals $i$ emits about past donors $k$, reaching others' reputations through every $s^k$ in parallel and not affecting $i$'s own immediate actions:
\begin{align}
\frac{\mathrm{d} \pi^i(s_t^j)}{\mathrm{d} \eta^i}
&= \frac{\partial \pi^i(s_t^j)}{\partial s_t^j} \cdot \sum_{k \ne i} \frac{\mathrm{d} s_t^j}{\mathrm{d} s^k} \cdot H_{\eta^i}^{k},
\label{eq:expansion-eta-own} \\
\frac{\mathrm{d} \pi^j(s_t^i)}{\mathrm{d} \eta^i}
&= \frac{\partial \pi^j(s_t^i)}{\partial s_t^i} \cdot \sum_{k \ne i} \frac{\mathrm{d} s_t^i}{\mathrm{d} s^k} \cdot H_{\eta^i}^{k}.
\label{eq:expansion-eta-other}
\end{align}
Each term sums over all third parties $k$ that $i$ has gossiped about, because a single change in $\eta^i$ perturbs every $s^k$ simultaneously; the reputation-to-reputation derivatives $\mathrm{d} s_t^j/\mathrm{d} s^k$ and $\mathrm{d} s_t^i/\mathrm{d} s^k$ are the same highway transitions derived in Section~\ref{sec:gradient_highways}. All four expressions are computed by automatic differentiation through the unrolled forward graph and require no hand-coded backward pass.

\subsection{Game Families in the Reputation Literature}

Research on reputation mechanisms spans evolutionary game theory, classical microeconomics, and multi-agent systems. Depending on the research emphasis, the literature uses three dominant game families.

\paragraph{Donation game.}
The donation game is a one-sided decision process: the donor chooses whether to pay cost $c$ to provide benefit $b$ to the recipient (with $b > c$), yielding payoffs $-c$ to the donor and $b$ to the recipient in the cooperative case. It is the canonical benchmark in evolutionary game theory, indirect reciprocity, and the evolution of social norms. \citet{nowak1998evolution} introduced image scoring in the donation game; \citet{ohtsuki2004should,ohtsuki2006leading} derived the leading eight in the donation game; and subsequent extensions to private, noisy, and incomplete information~\citep{hilbe2018indirect,fujimoto2023evolutionary,schmid2023quantitative} all retain the donation-game payoff structure. The donation game's appeal is that it strips away direct reciprocity and isolates, in minimal parametric form, the exact question of how a third-party observer updates the donor's reputation from a single observed action.

\paragraph{Prisoner's Dilemma.}
The Prisoner's Dilemma (PD) is a symmetric, simultaneous game in which both agents choose cooperate (C) or defect (D), with payoffs $T > R > P > S$ and $2R > T + S$. It is the canonical model of cooperation under conflicting incentives in classical game theory and is used by recent RL-based indirect-reciprocity studies~\citep{anastassacos2021cooperation,ren2025bottom}. The donation game is in fact a special case of the PD: playing a simultaneous round of the donation game between two agents induces the PD payoff matrix with $R = b - c$, $T = b$, $S = -c$, $P = 0$, which satisfies the PD inequalities whenever $b > c > 0$.

\paragraph{Trust game.}
The trust (or investment) game is a sequential two-stage game in which an investor transfers an amount to a trustee, the transferred amount is multiplied by some factor $k$, and the trustee decides what fraction to return. It is the canonical model in the behavioral-economics literature on reputation in markets and rating systems~\citep{masuda2012coevolution,keser2003experimental,sommerfeld2007gossip,fehr2019gossip}. The trust game emphasizes moral hazard under enforceable contracts and partner-specific reputation rather than the assessment-rule structure that indirect-reciprocity theory isolates.

\subsection{Why the Donation-Game Choice Is Canonical}

Our derivation targets the donation game, which is the canonical choice for the indirect-reciprocity and social-norm branches the paper sits in. Three reasons make this choice theoretically complete rather than restrictive.

\paragraph{Cleanest isolation of the indirect-reciprocity mechanism.}
The donation game has a two-parameter payoff structure ($b$, $c$), which lets the gradient derivation stay analytic and keeps the dependence of the reputation update on the donor's single action $a^{i \to j}$ fully explicit. Starting from a general-sum matrix game would entangle the reputation-update channel with direct payoff asymmetries, making it hard to attribute gradient flow to the indirect-reciprocity channel as opposed to direct payoff conflicts. The donation-game reduction isolates the mechanism under study.

\paragraph{Homomorphism to symmetric matrix games.}
Because the donation game embeds as the special PD case $(R, T, S, P) = (b-c, b, -c, 0)$, any equilibrium or evolutionary-stability result we obtain under donation-game dynamics transfers to the corresponding symmetric matrix game by the payoff parameterization. Every classical leading-eight stability result we benchmark against is stated for the donation game for the same reason: it admits a lossless affine mapping into the broader PD parameter family.

\paragraph{Alignment with the indirect-reciprocity comparison set.}
Every baseline and every classical evolutionary-stability result that our experiments touch (L3, L6, HybridCoop, AllDefector, ProudCooperator) is defined for the donation game. Using the same underlying game is what makes the empirical comparison legible to the indirect-reciprocity literature.

\subsection{Extension to Continuous Deterministic General-Sum Games}

Our main-text policies are already deterministic continuous maps rather than probability distributions over discrete actions, so the most faithful generalization of the derivation is not to a discrete-matrix game but to a continuous-action general-sum game with deterministic policies. Under this view the derivation stays inside the deterministic-policy-gradient (DPG) calculus~\citep{silver2014deterministic,lillicrap2015ddpg}: instead of the log-derivative trick over a discrete distribution, we apply the multivariate chain rule directly to a continuously differentiable payoff function. Below we carry out this extension with no discrete-action scaffolding and keep the same notation as Appendix~\ref{app:donation_derivation}.

\paragraph{Continuous deterministic setup.}
\emph{Action space.} The action space is a continuous subset $\mathcal{A} \subseteq \mathbb{R}^d$ rather than a discrete set.

\emph{Policy as a deterministic map.} The policy $\pi^i(s_t^j)$ is no longer a probability; it directly outputs the continuous action that agent $i$ takes on observing state $s_t^j$ (e.g.\ the continuous donation amount or the continuous signal intensity used throughout the main text). Hence $\pi^i(s_t^j) \in \mathcal{A}$ is a deterministic continuous map.

\emph{Continuously differentiable payoff.} The per-step payoff is an arbitrary continuously differentiable function $u^i(a^i, a^j)$: when $i$ plays $x \in \mathcal{A}$ and opponent $j$ plays $y \in \mathcal{A}$, agent $i$ receives $u^i(x, y)$. No symmetry, no matrix structure, no ordering of actions is assumed.

\emph{Reputation-chained opponent response.} We treat every opponent's policy \emph{function} $\pi^j$ as fixed (no opponent \emph{learning} model), but the policy \emph{output} $\pi^j(s_t^i)$ still depends on agent $i$'s own policy because the observation $s_t^i$ is the reputation state of agent $i$, which accumulates signals emitted during prior interactions that themselves depend on $\pi^i$. The total derivative $\mathrm{d} \pi^j(s_t^i) / \mathrm{d} \pi^i$ is therefore non-zero in general; it factors as $(\partial \pi^j / \partial s_t^i) \cdot (\mathrm{d} s_t^i / \mathrm{d} \pi^i)$ and is precisely the gradient-highway channel of Section~\ref{sec:gradient_highways}.

\emph{Matching.} The matching law of Section~\ref{sec:problem_setup} is retained, with the undirected interaction indicator $\mathbbm{1}(i \leftrightarrow j)$ and $\mathbb{E}_\xi[\mathbbm{1}(i \leftrightarrow j)] = 1/N(N-1)$.

\paragraph{Gradient derivation.}
In the deterministic continuous setting the per-interaction reward is just the payoff evaluated at the two deterministic policy outputs, $r_t^{i,j} = u^i \big( \pi^i(s_t^j), \pi^j(s_t^i) \big)$, and the expectation $\mathbb{E}_\xi$ only covers environmental randomness (matching draws and state draws), not discrete-action sampling. Applying the expectation-derivative interchange from Appendix~\ref{app:donation_derivation}:

\begin{equation*}
\begin{aligned}
\frac{\mathrm{d}}{\mathrm{d} \theta^i} \mathbb{E}_{\xi}\left[ \sum\limits_{t=1}^T r_t^i \right]
&= \frac{\mathrm{d}}{\mathrm{d} \theta^i} \mathbb{E}_{\xi} \Bigg[ \sum\limits_{t=1}^T \sum\limits_{j\ne i} \mathbbm{1}(i \leftrightarrow j) \cdot u^i \big( \pi^i(s_t^j), \pi^j(s_t^i) \big) \Bigg] \\
&= \sum\limits_{t=1}^T \sum\limits_{j\ne i} \underbrace{\mathbb{E}_{\xi}\big[\mathbbm{1}(i \leftrightarrow j)\big]}_{=\frac{1}{N(N-1)}} \cdot \mathbb{E}_{\xi}\bigg[ \frac{\mathrm{d}}{\mathrm{d} \theta^i} u^i \big( \pi^i(s_t^j), \pi^j(s_t^i) \big) \bigg] \\
&= \frac{1}{N(N-1)} \mathbb{E}_{\xi} \Bigg[ \sum\limits_{t=1}^T \sum\limits_{j\ne i} \frac{\mathrm{d}}{\mathrm{d} \theta^i} u^i \big( \pi^i(s_t^j), \pi^j(s_t^i) \big) \Bigg].
\end{aligned}
\end{equation*}

To evaluate $\mathrm{d} u^i / \mathrm{d} \pi^i$, we apply the multivariate chain rule to the two inputs of the continuously differentiable payoff: the partial with respect to the first input (agent $i$'s own continuous action $\pi^i(s_t^j)$) and the partial with respect to the second input (the opponent's continuous action $\pi^j(s_t^i)$):

\begin{equation*}
\begin{aligned}
\frac{\mathrm{d}}{\mathrm{d} \theta^i} u^i \big( \pi^i(s_t^j), \pi^j(s_t^i) \big)
&= \frac{\partial u^i \big( \pi^i(s_t^j), \pi^j(s_t^i) \big)}{\partial \pi^i(s_t^j)} \cdot \frac{\mathrm{d}}{\mathrm{d} \theta^i} \pi^i(s_t^j) + \frac{\partial u^i \big( \pi^i(s_t^j), \pi^j(s_t^i) \big)}{\partial \pi^j(s_t^i)} \cdot \frac{\mathrm{d}}{\mathrm{d} \theta^i} \pi^j(s_t^i).
\end{aligned}
\end{equation*}

Both terms are retained: the first captures the direct marginal effect of changing $\pi^i$ on $i$'s own payoff, and the second captures the indirect effect routed through the opponent's state-conditioned response, which is non-zero because the opponent's input $s_t^i$ is the reputation state of $i$ and therefore itself depends on $\pi^i$ via the signals emitted in prior interactions. Substituting back yields the final continuous deterministic gradient:

\begin{equation*}
\begin{aligned}
\frac{\mathrm{d}}{\mathrm{d} \theta^i} \mathbb{E}_{\xi}\left[ \sum\limits_{t=1}^T r_t^i \right]
&= \frac{1}{N(N-1)} \mathbb{E}_{\xi} \Bigg[ \sum\limits_{t=1}^T \sum\limits_{j\ne i} \bigg( \frac{\partial u^i \big( \pi^i(s_t^j), \pi^j(s_t^i) \big)}{\partial \pi^i(s_t^j)} \cdot \frac{\mathrm{d}}{\mathrm{d} \theta^i} \pi^i(s_t^j) \\
&\qquad\qquad + \frac{\partial u^i \big( \pi^i(s_t^j), \pi^j(s_t^i) \big)}{\partial \pi^j(s_t^i)} \cdot \frac{\mathrm{d}}{\mathrm{d} \theta^i} \pi^j(s_t^i) \bigg) \Bigg].
\end{aligned}
\end{equation*}

\paragraph{Interpretation.}
The gradient decomposes into two additive terms with clear physical meaning, both arising from the chain rule on a continuously differentiable payoff $u^i$ and both surviving for any continuous reward structure.

(1) \emph{Direct marginal-payoff term}, $\frac{\partial u^i(\pi^i(s_t^j), \pi^j(s_t^i))}{\partial \pi^i(s_t^j)} \cdot \frac{\mathrm{d} \pi^i(s_t^j)}{\mathrm{d} \theta^i}$. This is the own-input partial of the payoff, contracted with agent $i$'s own policy derivative. It captures the direct effect of changing $\pi^i$ on $i$'s payoff at a single interaction, holding the opponent's action fixed.

(2) \emph{Reputation-chained opponent-response term}, $\frac{\partial u^i(\pi^i(s_t^j), \pi^j(s_t^i))}{\partial \pi^j(s_t^i)} \cdot \frac{\mathrm{d} \pi^j(s_t^i)}{\mathrm{d} \theta^i}$. This is the opponent-input partial of the payoff, contracted with the total derivative of the opponent's action with respect to $\pi^i$. Even though $\pi^j$ itself is fixed as a function (we do not model its learning), the opponent's input $s_t^i$ is the reputation state of $i$, which depends on $\pi^i$ through the signals emitted in prior interactions, so $\mathrm{d} \pi^j(s_t^i) / \mathrm{d} \pi^i \ne \mathbf{0}$. This is precisely the gradient-highway channel of Section~\ref{sec:gradient_highways} in continuous form: the reputation-mediated response of the opponent is what carries the analytic signal that justifies cooperation against the direct marginal-payoff penalty.

\paragraph{Theoretical positioning.}
Anchoring the derivation at the continuously differentiable payoff $u^i$ with a deterministic continuous policy gives a prior-free mathematical basis for reputation-mediated cooperation beyond binary discrete formulations. The continuous leading-eight relaxations used as test-bed opponents in our experiments, and more generally any non-linear continuous reputation update, fit this setting without modification. The first term above is the standard DPG-style own-payoff gradient; the second term is the reciprocity-gradient contribution, made explicit here as the chain-rule product of the opponent's state-sensitivity and the reputation's sensitivity to $i$'s own past actions. Together they confirm that the reciprocity gradient is not a heuristic enhancement but the faithful total derivative of the cumulative payoff with respect to a deterministic continuous policy whenever reputation couples an agent's past actions to its future opponents' inputs.

%\newpage
\section{Opponent-Modeling Engineering Details}
\label{app:om_engineering}

This appendix collects the engineering choices for the differentiable opponent-modeling procedure used in Section~\ref{sec:opponent_modeling}. These are the choices that make the algorithm work in practice; none of them are core contributions of the paper.

\subsection{Full Algorithm}
\label{app:full_algorithm}

Algorithm~\ref{alg:opponent_modeling} states the complete procedure used by the learner $i$. The loop has three phases per outer iteration: (1) data collection by interacting with the true opponent population in the real environment, (2) fitting the private surrogate functions to the publicly observed actions and signals each opponent emits, and (3) policy update by replaying the recorded matching sequence in a \emph{virtual graph} where every opponent's action and signal is produced by its surrogate, then ascending the analytic gradient of the simulated cumulative return through that graph by automatic differentiation. An optional one-off \emph{exploration phase} at $t{=}0$, in which $i$ plays $K_\text{explore}$ episodes under a uniform-random action policy and pretrains the surrogates on the resulting interactions before any policy update, supplies the initial estimator coverage that prevents on-policy buffer clustering from re-saddling the surrogate (Setting (B.1)).

\begin{algorithm}[!t]
\DontPrintSemicolon
\caption{Reciprocity gradient with differentiable opponent modeling}
\label{alg:opponent_modeling}
\KwIn{learner index $i$; population size $N$; matching law $\xi$; benefit $b$, cost $c$.}
\KwIn{batch size $B$; outer iterations $T_\text{outer}$; estimator steps $K_\text{est}$; policy steps $N_\text{train}$; modeling buffer window $W$.}
\KwIn{exploration episodes $K_\text{explore}$; exploration pretrain steps $K_\text{pre}$ (set to $0$ to disable).}
\KwIn{learning rates $\eta_\text{est}$ for surrogate updates and $\eta_\theta, \eta_\eta$ for $\pi^i, \varphi^i$.}
\BlankLine
Initialize learner policies $\pi^i_{\theta^i}, \varphi^i_{\eta^i}$ at random.\;
Initialize private surrogates $\hat\pi^j_i, \hat\varphi^j_i$ for every $j \neq i$.\;
Initialize observation buffer $\mathcal{B}\leftarrow\emptyset$ (sliding window of last $W$ outer iterations).\;
\BlankLine
\tcp{(optional) cold-start exploration --- no oracle access}
\If{$K_\text{explore} > 0$}{
    \For{$k = 1, \dots, K_\text{explore}$}{
        Roll out one episode under matching $\xi$ with $i$ playing the uniform-random action policy $\pi^i(s)\!\sim\!U[0,1]$.\;
        Record every publicly observable tuple $(a^{j\to i}_t, s^i_t, \sigma^{i\gets j}_t)$ for each opponent $j$ into $\mathcal{B}$.\;
    }
    \For{$j\neq i$}{
        Take $K_\text{pre}$ MSE steps fitting $\hat\pi^j_i$ to the recorded action samples of $j$ and $\hat\varphi^j_i$ to the recorded signal samples of $j$.\;
    }
}
\BlankLine
\For{$t = 1, \dots, T_\text{outer}$}{
    \tcp{Phase 1 --- real-environment rollout}
    Sample matching sequence $M_t \sim \xi$ and play $B$ parallel episodes with current $(\pi^i_{\theta^i}, \varphi^i_{\eta^i})$ against the true opponents.\;
    Append every publicly observed $(a^{j\to k}_t, s^k_t, \sigma^{k\gets j}_t)$ into $\mathcal{B}$; drop transitions older than $W$ outer iterations.\;
    \BlankLine
    \tcp{Phase 2 --- fit surrogates from public observations}
    \For{$j \neq i$}{
        \For{$k = 1, \dots, K_\text{est}$}{
            Sample mini-batch from $\mathcal{B}$ at $j$-side observations.\;
            $\hat\pi^j_i \leftarrow \hat\pi^j_i - \eta_\text{est}\nabla\,\text{MSE}\!\left(\hat\pi^j_i(\text{input}),\, a^{j\to\cdot}\right)$.\;
            $\hat\varphi^j_i \leftarrow \hat\varphi^j_i - \eta_\text{est}\nabla\,\text{MSE}\!\left(\hat\varphi^j_i(\text{input}),\, \sigma^{\cdot\gets j}\right)$.\;
        }
    }
    Freeze surrogate parameters; their requires-grad status is set to false for Phase 3.\;
    \BlankLine
    \tcp{Phase 3 --- policy update via the virtual computation graph}
    \For{$n = 1, \dots, N_\text{train}$}{
        Replay matching $M_t$, but for each step $(j\!\to\!k)$ with $j\neq i$ substitute the action by $\hat\pi^j_i$ and the signal by $\hat\varphi^j_i$, while $i$'s own action and signal are produced by the live $\pi^i_{\theta^i}, \varphi^i_{\eta^i}$ to keep the gradient path open at $i$.\;
        Compute the virtual cumulative return $\hat R^i = \sum_t \hat r^i_t$ along this rollout, retaining the full differentiable graph from $(\theta^i, \eta^i)$ to the scalar.\;
        Compute the analytic gradient by automatic differentiation (e.g., PyTorch~\citep{paszke2019pytorch}):\;
        \quad $g_\theta \leftarrow \nabla_{\theta^i}\hat R^i,\qquad g_\eta \leftarrow \nabla_{\eta^i}\hat R^i.$\;
        Apply ascent: $\theta^i \leftarrow \theta^i + \eta_\theta\, g_\theta;\quad \eta^i \leftarrow \eta^i + \eta_\eta\, g_\eta$.\;
    }
}
\Return{trained policies $(\pi^i_{\theta^i}, \varphi^i_{\eta^i})$.}
\end{algorithm}

The replay in Phase 3 introduces no extra sampling bias since the matching law $\xi$ is policy-independent under both regimes of Section~\ref{sec:problem_setup}; substituting surrogates at non-learner nodes affects only the policy-dependent part of the rollout. The analytic gradient through this virtual rollout is a single backward pass over the rolled-out interaction graph (Appendix~\ref{app:demo} verifies this empirically on a minimal three-agent demo).

\subsection{Design Variants}
\label{app:design_variants}

Algorithm~\ref{alg:opponent_modeling} admits several design choices, summarized below; the experiments fix the defaults marked (D).

\paragraph{Source of modeling data.}
(D)~\emph{Sliding window}: a buffer of the most recent $W$ outer iterations of observed transitions, used both for opponent-model fitting and as the basis for replayed sequences in the virtual graph. (a)~\emph{Same-batch}: current batch only; simpler, but high variance. (b)~\emph{Previous-batch}: cleanly separates modeling and optimization data, at the cost of one-iteration model lag.

\paragraph{Virtual-graph construction.}
(D)~\emph{Replay} the recorded matching sequences with opponent models substituted at non-learner nodes; introduces no bias because the matching law is policy-independent. (a)~\emph{Re-simulate} fresh matching sequences inside the virtual graph; fully decouples real and virtual draws but doubles the per-iteration compute.

\paragraph{Update frequency.}
(D)~Update opponent models every outer iteration. (a)~Every $K > 1$ outer iterations when opponents change slowly. (b)~Strictly alternating modeling and optimization phases; cleanest separation but lower data efficiency. (c)~\emph{Explore-then-freeze}: at $t=0$, before policy optimization begins, Agent~0 plays $K_\text{explore}{=}100$ episodes against the true opponents under a uniform-random action policy and records the publicly observable $(a, s, \sigma)$ tuples each opponent emits during natural gameplay; the estimators are fitted to this exploration buffer offline and then frozen for the remainder of training. This variant avoids the buffer-clustering drift identified in Setting (B.1) and is what we use to recover the L6 discriminative action rule. The protocol is fully observational: opponents are never queried at synthetic inputs, only observed through actual interactions, so no oracle access is invoked.

\paragraph{Estimator architecture.}\label{app:factored_mlp}
(D)~A small differentiable MLP fitted by mean-squared error on publicly observable interaction data. Layer widths and remaining hyperparameters are listed in the per-experiment paragraphs of Appendix~\ref{app:detailed_experiments}.

\paragraph{Per-opponent vs.\ parameter-shared.}
(D)~\emph{Per-opponent}: one pair $(\hat\pi^j_i, \hat\varphi^j_i)$ per opponent, $O(N)$ parameters. (a)~\emph{Parameter-shared}: one pair $(\hat\pi(\cdot; z_j), \hat\varphi(\cdot; z_j))$ shared across opponents via per-opponent embeddings $z_j$, $O(1)$ parameters in $N$. Setting (D) reports a scalability comparison.

\subsection{Model Accuracy and Gradient Bias}
\label{app:model_accuracy}

If $\hat\pi^j_i$ matches $\pi^j$ exactly (and likewise for $\hat\varphi^j_i$), the gradient computed through the virtual graph equals the gradient computed through the true computation graph. With imperfect models, the virtual gradient differs from the true gradient by a bias term whose magnitude is bounded by the supremum of the model error along the unrolled trajectory, multiplied by the per-step decay factor $0.25$ at each gradient-highway hop.

Empirically we observe a stable \emph{virtual-reward optimism} on the configurations we test. Under the online-update protocol the per-interaction-payoff gap between surrogate and real environment is $+0.1$ to $+0.3$; under the explore-then-freeze protocol used in Setting (B.1) the gap rises to $\approx +1.0$, is present at outer iteration $0$ before any policy update, and does not grow over training. The gap is not explained by per-point approximation error (uniform-grid residual has mean $\approx 0$ and max magnitude $0.012$); it is consistent with compounding of small on-policy residuals through the $T \approx 50$-interaction reputation dynamics, in line with model-based reinforcement learning, where imperfect world models still yield useful policy-improvement gradients~\citep{sutton1991dyna}. The per-step decay bound caps the bias at a constant offset rather than letting it grow unboundedly; all external performance claims in the paper use the real-environment number, and we report surrogate values only as a training diagnostic.

\subsection{Comparison to Prior Approaches to Oracle Access}

The reciprocity-gradient expansion of Section~\ref{sec:gradient_expansion} requires backpropagating through other agents' action and gossip functions. A naive implementation gives the learner oracle access to every other agent's parameters, which is implausible when opponents are autonomous. Three families of prior work address this. \emph{Centralized training, decentralized execution} (CTDE) accepts oracle access at training time: DIAL~\citep{foerster2016learning} backpropagates through inter-agent message channels, MADDPG~\citep{lowe2017multi} trains a centralized critic over joint observations and actions, and BPTA~\citep{li2024backpropagation} backpropagates through the joint action-execution order. \emph{Opponent shaping} methods such as LOLA~\citep{foerster2018learning}, SOS~\citep{letcher2019stable}, and COLA~\citep{willi2022cola} take a \emph{second-order} gradient through one anticipated opponent learning step, explicitly requiring the partial $\partial \theta^j_{\text{new}} / \partial \theta^i$ of the opponent's update on the ego agent's parameters. This term is only well-defined against \emph{learning} opponents, so these methods are not applicable as baselines in our best-response setting where opponents are fixed: the term they backpropagate through is identically zero. Our method is, by contrast, \emph{first-order}: we differentiate the ego agent's expected return through one autograd backward pass over the rolled-out interaction graph against fixed opponents, a structurally different and strictly weaker requirement. The single-hop sender-receiver \emph{signaling gradient} of \citet{lin2023information} is the closest first-order analogue; the reciprocity gradient extends the same structural idea to multi-hop reputation chains. \emph{Opponent modeling} replaces access to others' parameters with a learned belief over their behavior~\citep{albrecht2018autonomous}; the learner differentiates through its own model rather than through the true opponent. We adopt this route in the observational-access regime: the learning agent builds private, differentiable estimators of each opponent and backpropagates the reciprocity gradient entirely through these self-owned models. MFOS~\citep{lu2022model} similarly removes the opponent-update access from LOLA's family via meta-learning over observed trajectories.

\newpage
% \addcontentsline{toc}{section}{Appendix}
% \part{Appendix}
\parttoc

\section{Experiments Overview}
\label{app:experiments_overview}

Tables~\ref{tab:experiments_overview_core} and~\ref{tab:experiments_overview_probes} summarize the role of every experiment in the paper. For each experiment we state the scientific question it addresses, the core setting (opponents, information access, trainable components), and the expected outcome that would make the experiment count as a success. The setting labels (A.1)--(A.3), (B.1)--(B.3), and (C)--(E) are stable identifiers used throughout the paper. They map to the topical structure of Section~\ref{sec:experiments} as follows. (A.1)--(A.3) (Section~\ref{sec:exp_groundtruth}) test best response under oracle opponent access for action only, signal only, and joint training respectively. (B.1)--(B.3) (Section~\ref{sec:exp_learnedmodels}) test the same three settings after replacing oracle access with learned opponent models. (C) (Section~\ref{sec:exp_baselines}) compares against model-free deterministic baselines. (D) (Section~\ref{sec:exp_scalability}) tests scalability via parameter sharing. (E) (Section~\ref{sec:exp_indirect}) tests the indirect-only matching regime. The autograd correctness demonstration referenced in Section~\ref{sec:gradient_highways} appears in Appendix~\ref{app:demo}. Detailed hyperparameters, seed counts, and full numeric results are provided in the individual paragraphs of Section~\ref{sec:experiments} and in Appendix~\ref{app:detailed_bestresp}.

\begin{table}[ht]
\centering
\footnotesize
\renewcommand{\arraystretch}{1.25}
\setlength{\tabcolsep}{5pt}
\begin{tabularx}{\textwidth}{@{}%
  >{\centering\arraybackslash}p{0.045\textwidth}%
  >{\raggedright\arraybackslash}p{0.19\textwidth}%
  >{\raggedright\arraybackslash}X%
  >{\raggedright\arraybackslash}X@{}}
\toprule
\textbf{ID} & \textbf{Question} & \textbf{Setting} & \textbf{Expected outcome} \\
\midrule
\multicolumn{4}{@{}l}{\emph{Best response under oracle opponent access}} \\
(A.1)  & Can a single agent recover a discriminative \emph{action} rule against leading-eight opponents? & Agent~0 trains $\pi^0$ against frozen $L_3$/$L_6$ opponents; $\varphi^0$ held at identity; multiple reputation inits. & $\pi^0$ becomes context-sensitive ($\mathrm{std}[\pi^0] \geq 0.2$) and reaches a high fraction of the full-cooperation benchmark. \\
(A.2)  & Can a single agent recover a discriminative \emph{signal} rule when the opponent mix forces signal discrimination? & Agent~0 trains $\varphi^0$ against frozen ProudCoop$(L_3)$+AllDefector$(L_3)$; $\pi^0$ held at $L_3$; $b{=}10$. & $\varphi^0$ differentiates cooperator vs.\ defector donors; per-interaction payoff approaches $2.25$. \\
(A.3)  & Can the reciprocity gradient optimize action and signal \emph{simultaneously} under oracle access? & Agent~0 trains both $\pi^0$ and $\varphi^0$ from scratch; the true opponents; no warmup, no phasing, no regularizer. & Simultaneous convergence to ``proud cooperator''-like discriminative strategy; per-interaction payoff $\approx 99\%$ of reference. \\
\addlinespace[2pt]
\multicolumn{4}{@{}l}{\emph{Best response under differentiable opponent models}} \\
(B.1)  & Does removing oracle access preserve action-rule recovery against $L_6$? & Agent~0 trains $\pi^0$ using differentiable surrogates $\hat\pi^{-0}, \hat\varphi^{-0}$ fit from public observations. & Discriminative $\pi^0$ still recovered; per-interaction payoff within a small gap of the oracle-access number. \\
(B.2)  & Same question for \emph{signal} optimization under learned opponent models. & Agent~0 trains $\varphi^0$ only; surrogate opponent models; ProudCoop+AllD setting. & Discriminative $\varphi^0$ recovered with per-interaction payoff $\geq 75\%$ of benchmark under learned models. \\
(B.3)  & Can simultaneous action-signal optimization succeed under learned opponent models? & Agent~0 trains $\pi^0$ and $\varphi^0$ simultaneously; surrogate models; joint setting HybridCoop+AllD (both networks must be discriminative). & High seed rate with $\mathrm{std}[\pi^0], \mathrm{std}[\varphi^0] \geq 0.2$ simultaneously; per-interaction payoff well above any single-network strategy. \\
\bottomrule
\end{tabularx}
\caption{\textbf{Experiments overview, part~1: method validation, groups (A)--(B) of Section~\ref{sec:experiments}.} Group (A) (Section~\ref{sec:exp_groundtruth}) tests best-response optimization under oracle opponent access, for the action, signal, and both-network settings. Group (B) (Section~\ref{sec:exp_learnedmodels}) tests the same three settings after replacing oracle access with differentiable opponent models fitted from public observations.}
\label{tab:experiments_overview_core}
\end{table}

\begin{table}[ht]
\centering
\footnotesize
\renewcommand{\arraystretch}{1.25}
\setlength{\tabcolsep}{5pt}
\begin{tabularx}{\textwidth}{@{}%
  >{\centering\arraybackslash}p{0.045\textwidth}%
  >{\raggedright\arraybackslash}p{0.19\textwidth}%
  >{\raggedright\arraybackslash}X%
  >{\raggedright\arraybackslash}X@{}}
\toprule
\textbf{ID} & \textbf{Question} & \textbf{Setting} & \textbf{Expected outcome} \\
\midrule
\multicolumn{4}{@{}l}{\emph{Baseline comparison}} \\
(C)  & Do model-free deterministic baselines (DPG, DDPG, TD3) match the reciprocity gradient? & Same nine settings as (B.1)--(B.3); baselines share the architecture, observation, and budget of the RG learner. & Baselines collapse to constant-output policies on every setting that requires discrimination; RG retains a large per-interaction-payoff margin. \\
\addlinespace[2pt]
\multicolumn{4}{@{}l}{\emph{Scalability sweep}} \\
(D)  & Does the method retain its per-interaction payoff when (B)'s three successful settings are scaled to larger populations? & Re-run (B)'s action vs L6 (\S\ref{sec:exp5}), signal vs ProudCoop+AllDefector pool (\S\ref{sec:exp6}, $c{=}5$), and joint vs HybridCoop+AllDefector pool (\S\ref{sec:exp7}) at $N\in\{5,10,15,20,25,30\}$; pool fractions $p\in\{1/5,3/10,1/2,7/10,4/5\}$ for the two pool-based settings; parameter-shared estimator with $O(1)$ surrogate parameters in $N$. & Per-interaction payoff is preserved across $N$ at every fixed $p$, and rises monotonically with $p$ as more cooperators lift the full-cooperation reference. \\
\addlinespace[2pt]
\multicolumn{4}{@{}l}{\emph{Indirect-only matching}} \\
(E) & Do (B)'s three successful settings still recover their discriminative best response when direct reciprocity is disabled? & Re-run (B)'s three settings under the (R2) indirect-only matching regime (each ordered pair meets exactly once per episode); $N{=}3$ with the (B) opponent populations; $N_\text{play}$ scaled to keep total transitions per outer iteration comparable across regimes. & Per-interaction payoff and discrimination rates approximately match the (R1) numbers, confirming cooperation is mediated by reputation rather than direct reciprocity. \\
\bottomrule
\end{tabularx}
\caption{\textbf{Experiments overview, part~2: comparison and scope, groups (C)--(E) of Section~\ref{sec:experiments}.} Group (C) (Section~\ref{sec:exp_baselines}) compares the reciprocity gradient against three deterministic model-free baselines. Group (D) (Section~\ref{sec:exp_scalability}) tests scalability across population sizes. Group (E) (Section~\ref{sec:exp_indirect}) tests robustness to indirect-only matching.}
\label{tab:experiments_overview_probes}
\end{table}

\section{Action-Only and Signal-Only Pathway Ablations}
\label{app:single_pathway_ablations}

This appendix complements Figure~\ref{fig:three_pathway} (which shows only the joint training setting) with the two single-pathway ablations: action-only training against L6, and signal-only training against ProudCoop+AllDefector.

\begin{figure}[ht]
\centering
\includegraphics[width=\textwidth]{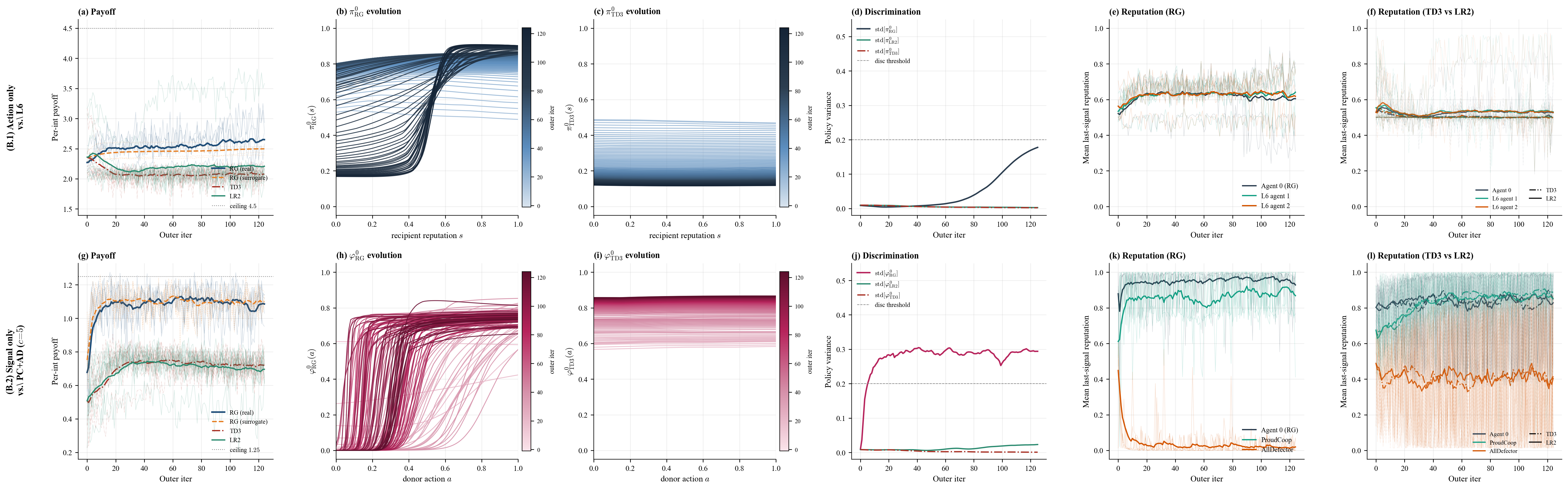}
\caption{Single-pathway ablations ($5$ seeds, $T_{\text{outer}}{=}125$). \emph{Top row.} action-only training against L6, reference $(b{-}c)/2 = 4.5$. \emph{Bottom row.} signal-only training against ProudCoop+AllDefector with $c{=}5$, reference $(b{-}c)/4 = 1.25$. Columns: per-interaction payoff; RG policy evolution; TD3 policy evolution; profile std for RG, TD3, and LR2 (descriptive only, not a metric); reputation under RG; reputation under TD3 and LR2. Faint curves show individual seeds; bold curves show the bias-corrected EMA of the across-seed mean. RG develops a non-flat learned policy in both ablations; TD3 and LR2 stay constant.}
\label{fig:three_pathway_appendix}
\end{figure}

\section{Detailed Best-Response Results vs Hard-Coded Opponents}
\label{app:detailed_bestresp}

This appendix reports the full sweep of best-response experiments against hard-coded opponents whose results are flat-cooperation, partial, or otherwise non-discriminative. The discriminative settings are summarized in the main text (Setting (A.1) for the L6 action result and Setting (A.2) for the L3 ProudCoop + AllD signal result).

\subsection{(A.1.i) Action Optimization vs Identity-Mapping Opponents}
\label{app:exp2a}

Agent~0's action network is trainable; signal is hard-coded as honest identity ($\sigma = a$). Opponents are hard-coded identity agents. The aim is to verify that the reciprocity gradient produces useful policy updates under the simplest possible reputation dynamics. We sweep two reputation aggregators (mean vs.\ EMA with decay $0.5$) and three reputation initializations (const $0.5$, all-uniform $U[0,1]$, and a warmup from the stationary distribution).

\begin{table}[ht]
\centering
\begin{tabular}{lccc}
\toprule
Aggregator & const05 & all\_uniform & warmup\_init \\
\midrule
mean & $3.997$ & $4.020$ & $4.027$ \\
EMA  & $4.273$ & $4.291$ & $4.292$ \\
\bottomrule
\end{tabular}
\caption{Setting (A.1.i): final per-interaction payoff against identity opponents (reference $= 4.5$). Both the reputation aggregator (rows: mean vs.\ EMA) and the reputation initialization (columns) are environment specifications, not learner-side hyperparameters; values across cells are therefore not directly comparable. Each entry reports the converged payoff under its own environment instance. The learned action policy is \emph{flat} at $\pi \approx 1$ in every variant, matching the theoretically predicted best response under each environment: with honest identity signals, reputation rises unboundedly as the agent cooperates, and the cost $c = 1$ is always repaid.}
\label{tab:exp2a}
\end{table}

\paragraph{Memory-window ablation.}
We also vary the opponent's reputation memory window (EMA, warmup\_init): $k = 1$ gives $4.360$, $k = 3$ gives $4.285$, $k = \infty$ gives $4.264$. Shorter memory produces faster reputation feedback and therefore a higher converged payoff; longer memory dilutes each cooperative action across more prior observations.

\begin{figure}[ht]
\centering
\includegraphics[width=0.95\textwidth]{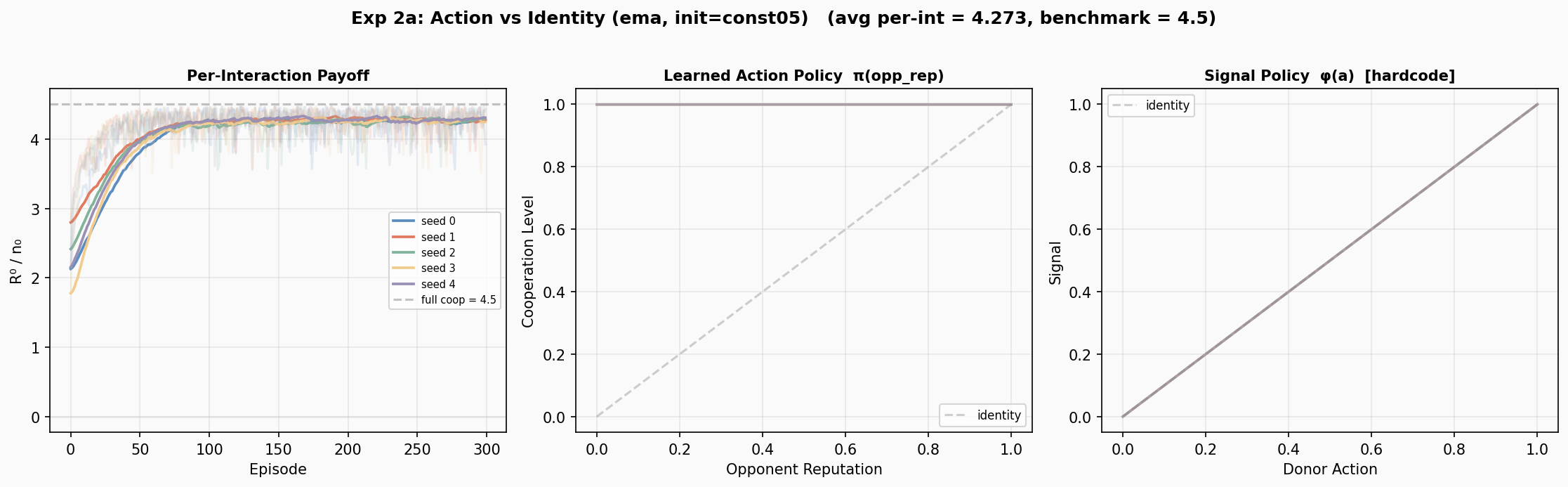}
\caption{Setting (A.1.i) (EMA, const05). Left: per-interaction payoff climbs from $\sim 0.5$ to the reference. Right: learned action policy $\pi^0$ is flat at $\approx 1$ across the reputation grid, confirming unconditional cooperation as the best response to identity opponents.}
\label{fig:exp2a_ema}
\end{figure}

\subsection{(A.2.i) Signal Optimization vs Identity-Mapping Opponents}
\label{app:exp3a}

Agent~0's \emph{signal} network is trainable; the action is hard-coded as the mean-reputation identity. Against identity opponents the predicted best signal is reputation \emph{inflation} ($\varphi^0(a) \to 1$): inflating others' reputation amplifies their cooperative responses to Agent~0 through the first-order feedback loop.

\begin{table}[ht]
\centering
\begin{tabular}{lccc}
\toprule
Aggregator & const05 & all\_uniform & warmup\_init \\
\midrule
mean & $3.002$ & $3.045$ & $3.037$ \\
EMA  & $3.742$ & $3.814$ & $3.810$ \\
\bottomrule
\end{tabular}
\caption{Setting (A.2.i): final per-interaction payoff with signal training only (reference $= 4.5$). Aggregator (rows) and reputation initialization (columns) are environment specifications, not learner hyperparameters; values across cells are not directly comparable. Learned signal is flat at $\varphi^0 \to 1$ in every variant, matching the predicted inflation equilibrium under each environment.}
\label{tab:exp3a}
\end{table}

\subsection{(A.2.ii) Signal Optimization vs Leading-Eight Opponents}
\label{app:exp3c}

Against structured leading-eight opponents, signal-only training is \emph{easier} than against identity opponents because the opponents' second-order assessment provides a cleaner gradient. Agent~0 trains $\varphi^0$ only; action is hard-coded.

\begin{table}[ht]
\centering
\begin{tabular}{lcccc}
\toprule
 & \multicolumn{2}{c}{L3} & \multicolumn{2}{c}{L6} \\
\cmidrule(lr){2-3}\cmidrule(lr){4-5}
Init Mode & mean & EMA & mean & EMA \\
\midrule
const05      & $3.79$ & $4.15$ & $3.19$ & $3.62$ \\
all\_uniform & $3.84$ & $4.19$ & $3.58$ & $4.08$ \\
warmup\_init & $3.96$ & $4.23$ & $3.92$ & $4.19$ \\
\bottomrule
\end{tabular}
\caption{Setting (A.2.ii): final per-interaction payoff with signal training against leading-eight opponents (reference $= 4.5$). Each (opponent, init, aggregator) cell is a separate environment variant; values across cells are not directly comparable. Learned signal converges to a near-identity mapping in every variant: when opponents already implement a stable second-order assessment, the best response under each environment is to match it rather than try to inflate.}
\label{tab:exp3c}
\end{table}

\section{Detailed Experimental Results}
\label{app:detailed_experiments}

This section reports the empirical evaluation of the reciprocity gradient. We address five experimental questions: \textbf{(i)} under oracle access to opponents' true functions, how close does the analytic gradient land to the full-cooperation reference in each setting (Section~\ref{sec:exp_groundtruth}); \textbf{(ii)} when opponents are replaced with private differentiable surrogates fitted from public observations, how far does the observational-access variant remain from the full-cooperation reference, and what fraction of the oracle performance does it recover (Section~\ref{sec:exp_learnedmodels}); \textbf{(iii)} on the same settings, how does the observational-access variant of the reciprocity gradient compare with deterministic policy-gradient baselines (DPG, DDPG, TD3) that are model-free by construction (Section~\ref{sec:exp_baselines}); \textbf{(iv)} does the method retain its per-interaction payoff at larger population sizes (Section~\ref{sec:exp_scalability}); and \textbf{(v)} does it retain its per-interaction payoff under the indirect-only matching regime that disables direct reciprocity (Section~\ref{sec:exp_indirect}).

\subsection{Overview and Metrics}
\label{sec:evaluation_metrics}

The experiments are organized into five groups (Table~\ref{tab:exp_matrix}). \textbf{(A)} (Section~\ref{sec:exp_groundtruth}) trains action, signal, and both policies against frozen opponents under oracle access, isolating the optimization step from any opponent-modeling error and establishing the oracle-access references that the rest of the paper targets. \textbf{(B)} (Section~\ref{sec:exp_learnedmodels}) repeats the three settings after replacing oracle access with private differentiable surrogates fitted from public observations. \textbf{(C)} (Section~\ref{sec:exp_baselines}) benchmarks the analytic-gradient path against three deterministic model-free actor-critic methods on the same nine settings. \textbf{(D)} (Section~\ref{sec:exp_scalability}) tests scalability by re-running all three (B) settings at population sizes $N\in\{5,10,15,20,25,30\}$ (the $N=3$ anchor is (B) itself). \textbf{(E)} (Section~\ref{sec:exp_indirect}) tests robustness to the indirect-only matching regime by re-running the same three settings without direct reciprocity.
\begin{table}[H]
\centering
\small
\begin{tabular}{lllll}
\toprule
  & Trained & Opponent action & Opponent signal & Access \\
\midrule
\multirow{3}{*}{A}
& $\pi^0$           & L6                      & L6 & Oracle \\
& $\varphi^0$       & ProudCoop, AllDefector  & L3 & Oracle \\
& $\pi^0,\varphi^0$ & HybridCoop, AllDefector & L3 & Oracle \\
\midrule
B  & \multicolumn{3}{c}{(same three settings as A, correspondingly trained)} & Observational \\
\midrule
C       & $\pi^0,\varphi^0$    & \multicolumn{2}{l}{$9$ (Trained $\times$ Opponent) settings; $3$ baselines} & N/A \\
\midrule
D  & \multicolumn{4}{c}{(same three settings as B, varying population size $N\in\{5,10,15,20,25,30\}$)} \\
\midrule
E  & \multicolumn{4}{c}{(same three settings as B, indirect-only matching regime)} \\
\midrule
F  & all $\pi^i,\varphi^i$ & \multicolumn{3}{c}{(homogeneous $N=3$ self-play, no fixed opponents)} \\
\bottomrule
\end{tabular}
\caption{Experiment matrix. Opponent labels: L3 (\emph{Simple Standing}, forgiving) and L6 (\emph{Stern Judging}, strict) are second-order leading-eight rules~\citep{ohtsuki2004should,ohtsuki2006leading}. Each fully specifies both an action policy (cooperate iff the recipient is in good standing under its own assessment) and a signal policy (the assessment rule itself). Discrete tables are provided in Appendix~\ref{app:ir_background} and the continuous formulation in Appendix~\ref{subsec:continuous_norms}. ProudCoop, HybridCoop, and AllDefector are action policies designed for the experiments and paired with the L3 signal rule. The functional definitions and objectives of experimental groups A through F are detailed in the preceding main text.}
\label{tab:exp_matrix}
\end{table}

\paragraph{Evaluation Metrics.}
Our primary evaluation metric is the \emph{per-interaction payoff} of the learning agent, averaged across seeds and reported in absolute units and as a percentage of an analytic per-setting reference determined by $b$, $c$, and the fixed opponents' policies. Per-setting references are stated with each result. We use the per-interaction unit rather than per-episode because episode length varies with the matching schedule under regime~R1 (Section~\ref{sec:problem_setup}), so per-episode returns are high-variance and not directly comparable across settings.

We also report the sample standard deviation of each learned policy's output, evaluated at convergence on a uniform grid spanning its input domain. We use this profile std as a \emph{qualitative descriptor} of policy shape rather than as a pass/fail metric: a near-zero value identifies constant-output attractors, while a positive value identifies non-constant policies. The descriptor cannot rank two non-constant policies (a step function and a noisy small-amplitude oscillation can have similar std, but only the first is the optimum), so we do not use it to compare methods. It is useful only to separate collapse modes from non-collapse modes when discussing what kind of attractor a method ends up in. Reported $\pm$ values are standard deviations across seeds (not confidence intervals); seed counts and reputation-initialization modes are declared per setting. The continuous formulation of the leading-eight norms used as opponents throughout the paper is derived in Appendix~\ref{subsec:continuous_norms}; Appendix~\ref{app:experiments_overview} provides a tabular restatement of the experiments. Reputation-initialization mode is varied per setting throughout the experiments below (declared per result), and a discrete-action extension of the framework is reported in Appendix~\ref{app:pd_discrete}.

\paragraph{Compute budget.}\label{app:settings}
All experiments were run on a single Apple M3 Max CPU (no GPU required). The entire experimental pipeline --- including all main-text experiments, the autograd demonstration of Appendix~\ref{app:demo}, all seeds, all configurations, and all ablations --- completes in approximately $8$ hours of wall-clock time. Individual experiment runtimes range from seconds (autograd demo, $3$ agents $\times$ $3$ timesteps) to $\sim 90$ minutes ((E) indirect-matching ablation, $60$ configurations $\times$ $20$ outer iterations $\times$ $5$ seeds). The method's modest compute footprint reflects the small population sizes ($N \le 15$) and the efficiency of analytic gradient computation through the surrogate graph relative to Monte-Carlo policy-gradient estimators.

\paragraph{Code release.}
Source code is not provided as supplementary material with this submission. The full simulation framework, including the environment, agents, all baselines, runner scripts, and the seed/hyperparameter records used to generate every figure and table in the paper, will be released under an open-source license upon acceptance.

\paragraph{Policy-variance grid.}
The policy-variance metric is computed as follows. The learner's action policy $\pi^0$ is evaluated at $21$ evenly-spaced recipient-reputation values $s \in \{0, 0.05, 0.10, \dots, 1\}$, each query feeding a single-step reputation history of value $s$; $\mathrm{std}[\pi^0]$ is the sample standard deviation of the resulting $21$-point cooperation-rate profile. The learner's signal policy $\varphi^0$ is evaluated analogously on a uniform $21$-point grid of donor actions over $[0, 1]$; for second-order signal policies that additionally consume the learner's own reputation, the auxiliary reputation input is held fixed at $0.5$ during evaluation. The grid resolution ($n_\text{grid} = 21$) is chosen to be fine enough that the profile standard deviation is insensitive to further refinement at the precision at which we report it.

\paragraph{Taxonomy of Opponent Regimes and Sanity Checks.}
To rigorously evaluate the reciprocity gradient, we conduct extensive experiments across a wide spectrum of opponent populations. We categorize each setting into one of two regimes based on the topological complexity of its analytical optimum.

In \emph{degenerate regimes} --- populations dominated by unconditional cooperators, by unconditional defectors, or by leading-eight opponents whose stationary reputation distribution renders the optimal response state-independent --- the analytical optimum collapses to a constant-output policy: unconditional cooperation, unconditional defection, or a constant signal output. Across dozens of degenerate configurations against identity, L3, L6, and exotic mixed populations under multiple reputation aggregators and initializations (Appendix~\ref{app:detailed_bestresp}), the reciprocity gradient consistently attains the oracle-access reference, confirming that the underlying state--reward dynamics, matching schedule, and reputation aggregator are unbiased and that the analytical optimum is reachable from random initialization. These configurations serve as a critical sanity check on the environment specification.

The core challenge of indirect reciprocity, however, lies in \emph{discriminative regimes} that demand state-dependent strategies. We construct three such regimes: action training against the second-order leading-eight rule L6 (\emph{Stern Judging}), signal training against an asymmetric ProudCoop+AllD pool, and joint training of both networks against HybridCoop+AllD. The deterministic policy-gradient baselines (Section~\ref{sec:exp_baselines}) collapse to constant-output policies on every discriminative regime, while the reciprocity gradient recovers the conditional structure and reaches $72$--$99\%$ of the oracle-access reference. The cooperator-fraction sweep in Section~\ref{sec:exp_scalability} interpolates between the two regimes along $p \in \{0.2, 0.3, 0.5, 0.7, 0.8\}$, with $p \to 0$ and $p \to 1$ approaching degenerate settings and $p = 0.5$ being discriminative. Every quantitative gap between the reciprocity gradient and the deterministic baselines reported in the main text is measured on a discriminative regime; Table~\ref{tab:taxonomy_summary} summarises the regime-level outcome and per-setting numbers are tabulated in Appendix~\ref{app:detailed_bestresp}.

\begin{table}[ht]
\centering
\small
\setlength{\tabcolsep}{4pt}
\begin{tabular}{p{4.5cm}lcc}
\toprule
Regime (representative opponents) & Required best response & RG (\% of reference) & Model-free baselines \\
\midrule
\multicolumn{4}{@{}l@{}}{\emph{Degenerate}} \\
Identity opponents (action) & constant $\pi^0\!=\!1$ & $89$--$95\%$ & $20$--$44\%$ \\
Identity opponents (signal) & constant $\varphi^0\!\to\!1$ & $67$--$85\%$ & $53$--$60\%$ \\
L3 opponents (action) & constant $\pi^0\!=\!1$ & $89$--$95\%$ & $33$--$49\%$ \\
L3 opponents (signal) & constant $\varphi^0$ matches the norm & $84$--$94\%$ & $92$--$93\%$ \\
L6 opponents (signal) & constant $\varphi^0$ matches the norm & $71$--$93\%$ & $79$--$92\%$ \\
ProudCoop+AllD (signal, flat-coop attractor) & constant cooperate at $b\!=\!10$ & $95\%$ & $79$--$92\%$ \\
\midrule
\multicolumn{4}{@{}l@{}}{\emph{Discriminative}} \\
L6 opponents (action) & state-dependent $\pi^0(s)$ & $72$--$93\%$ ($\geq\!5/5$ disc) & $30$--$49\%$ (collapse to flat) \\
ProudCoop+AllD (signal, $c{=}5$) & state-dependent $\varphi^0(a)$ & $96\%$ ($5/5$ disc) & $53$--$61\%$ (collapse to flat) \\
HybridCoop+AllD (joint) & state-dependent $\pi^0,\varphi^0$ & $99\%$ ($20/20$ disc) & $69$--$80\%$ (collapse to flat) \\
\bottomrule
\end{tabular}
\caption{Aggregated taxonomy of opponent regimes. \emph{Degenerate} regimes admit a constant-output analytical optimum; \emph{discriminative} regimes require a state-dependent analytical optimum. The reciprocity gradient (RG) reaches near-full-cooperation-reference on every regime, while the deterministic policy-gradient baselines DPG/DDPG/TD3 (Section~\ref{sec:exp_baselines}) collapse to constant-output policies wherever the optimum is state-dependent. Degenerate settings serve as sanity checks: their oracle-access reference is reachable by any constant-output policy and the environment is therefore unbiased. The $19$--$30$~pp margin between RG and the baselines on the joint discriminative regime, the $35$--$43$~pp margin on the signal vs.\ ProudCoop+AllD ($c{=}5$) regime, and the absence of any disc seed in the baseline ensemble together quantify the value of the analytic gradient. Per-setting numbers and learned-policy shapes are listed in Tables~\ref{tab:exp2a}--\ref{tab:exp3c}.}
\label{tab:taxonomy_summary}
\end{table}

\subsection{Best Response under Oracle Opponent Access}
\label{sec:exp_groundtruth}

We first establish the references the rest of the paper targets. Under oracle access to opponents' true action and signal functions, the reciprocity gradient is the exact analytic gradient through the multi-agent rollout; results are therefore free of opponent-modeling error and the question reduces to whether the optimization step alone recovers the predicted discriminative policies. We report three settings: action only (signal hard-coded), signal only (action hard-coded), and simultaneous optimization of both networks from scratch.

\begin{figure}[ht]
\centering
\includegraphics[width=0.48\textwidth]{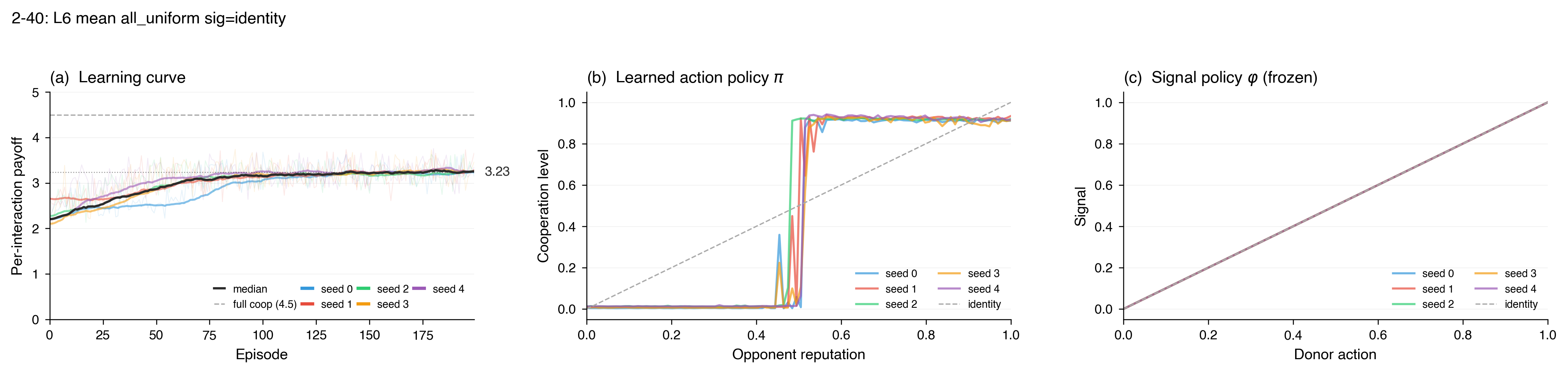}
\includegraphics[width=0.48\textwidth]{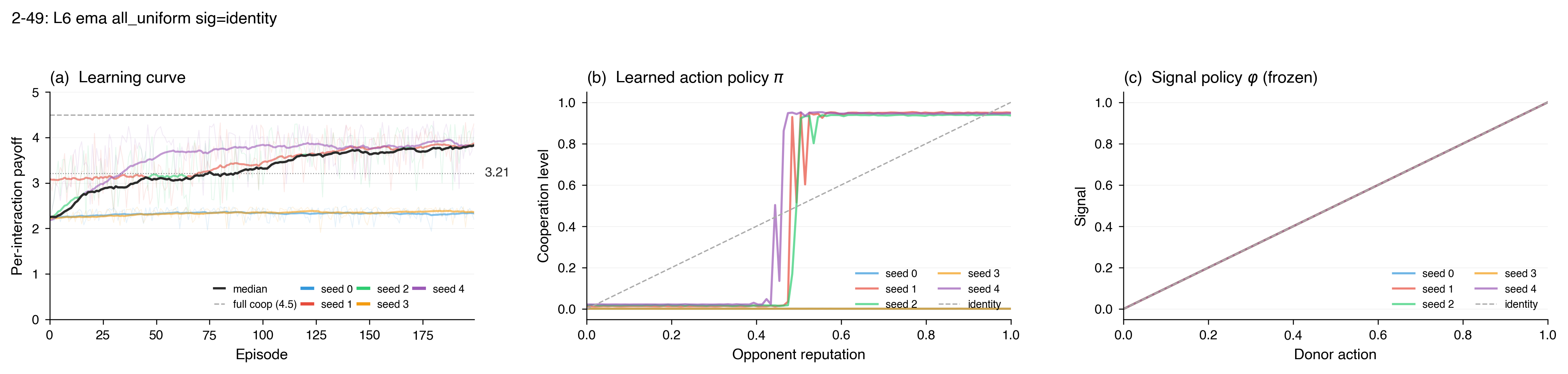}
\caption{Action-network recovery against L6 opponents with \texttt{all\_uniform} initialization and identity signal. \emph{Left:} \texttt{mean} aggregator. \emph{Right:} EMA aggregator. The learned policy $\pi^0(s)$ drops from near $1$ at high recipient reputation to near $0$ at low recipient reputation, separating the cooperate-pays sub-region from the punish-pays one. Final per-interaction payoff is $3.25$ (\texttt{mean}) and $3.24$ (EMA) against the $4.5$ reference.}
\label{fig:exp2b_disc}
\end{figure}

\paragraph{Action optimization against L6.}
\label{sec:exp2_disc}
Against the second-order leading-eight rule L6 --- \emph{Stern Judging}, the strictest leading-eight norm~\citep{ohtsuki2004should,ohtsuki2006leading}; full assessment table and contrast with L3 (\emph{Simple Standing}) given in Appendix~\ref{app:ir_background} --- the reciprocity gradient recovers a discriminative action policy under oracle opponent access. The learned $\pi^0$ separates the reputation sub-region in which cooperation pays from the sub-region in which it does not, settling at the rectangular decision boundary that the L6 second-order assessment carves through reputation space (Figure~\ref{fig:exp2b_disc}). Across $5$ seeds, the per-interaction payoff reaches $3.25$ against the $4.5$ reference ($72\%$); every seed converges to a non-constant $\pi^0$ rather than to a constant-output attractor. The discriminative setting is robust across the two reputation aggregators we test (\texttt{mean} and EMA); we sweep reputation initialization and hard-coded signal as additional axes, and report the remaining configurations --- including all L3 settings and the non-discriminative L6 settings --- in Appendix~\ref{app:detailed_bestresp}, where they collapse to flat-cooperation or saddle attractors.

\paragraph{Signal optimization against ProudCoop+AllD (L3).}
\label{sec:exp3}
Against an asymmetric opponent pair --- a reputation-conditioned cooperator and an unconditional defector, both emitting gossip under L3 (Simple Standing) --- the reciprocity gradient recovers a discriminative signal rule (Table~\ref{tab:exp3}). Across $10$ seeds per initialization, every seed learns a discriminative signal policy: under \texttt{const05} initialization the per-interaction payoff reaches $94\%$ of the $2.25$ reference ($10/10$ disc), and under \texttt{all\_uniform} it reaches $89\%$ ($10/10$ disc). Three structural choices drive discrimination on this setting. First, L3 is forgiving: it assigns a good signal when the learner justifiably defects against an all-defector, preserving the gradient channel that drives discrimination. A strict second-order norm such as L6 punishes the same move and destroys the gradient. Second, AllDefector is a constant opponent rather than a reputation-conditioned one, so the signal network faces a clean two-class problem (cooperator vs.\ defector) rather than a continuous asymmetry. Third, the proud-cooperator's behavior conditions on its \emph{own} reputation, so the reciprocity gradient flows through the learner's signal and back into the proud-cooperator's next action. This self-referential gradient channel lifts the learner from the flat-inflation policy (per-interaction payoff $\sim\!0.4$) into the discriminative regime ($\approx\!2.1$).

\begin{table}[ht]
\centering
\small
\begin{tabular}{lccccc}
\toprule
Initialization & $n$ & Per-int (mean) & \% bench & $\mathrm{std}[\varphi^0]$ above $0.20$ & Notes \\
\midrule
\texttt{const05}      & 10 & $\mathbf{2.112\pm 0.024}$ & $\mathbf{94\%}$ & $\mathbf{10/10}$ & most robust setting \\
\texttt{all\_uniform} & 10 & $2.008\pm 0.025$          & $89\%$          & $10/10$         & --- \\
\bottomrule
\end{tabular}
\caption{Signal-network recovery against ProudCoop(L3)+AllDefector(L3); benchmark $(b-c)/4 = 2.25$ at $b=10, c=1$. The learned signal policy is reported on a uniform donor-action grid.}
\label{tab:exp3}
\end{table}

\paragraph{Both networks and the asymmetric-LR protocol.}
\label{sec:exp4}
We next train $\pi^0$ and $\varphi^0$ simultaneously from random initialization, with both networks updated against the same scalar reward. Classical opponents do not constrain both networks simultaneously: ProudCoop conditions only on its own reputation (so $\varphi^0$ alone suffices), while L3/L6 condition only on the recipient's reputation (so $\pi^0$ alone suffices). We therefore design \textbf{HybridCoop(L3)}, whose cooperation rate is $\sigma\bigl(10\cdot(0.5\, r_\text{own} + 0.5\, r_\text{recipient} - 0.5)\bigr)$. The dual conditioning forces both networks to be discriminative at the optimum: $\pi^0$ must respond to the recipient's reputation, and $\varphi^0$ must shape HybridCoop's reputation through its signals. Two frozen opponents (HybridCoop(L3) and AllDefector(L3)); $b=10$, $c=1$, $N=3$, batch $128$, $\delta=0.98$; uniform reputation initialization; both networks trainable from random initialization, no warmup, no phasing, no regularization.

\begin{figure}[ht]
\centering
\includegraphics[width=0.95\textwidth]{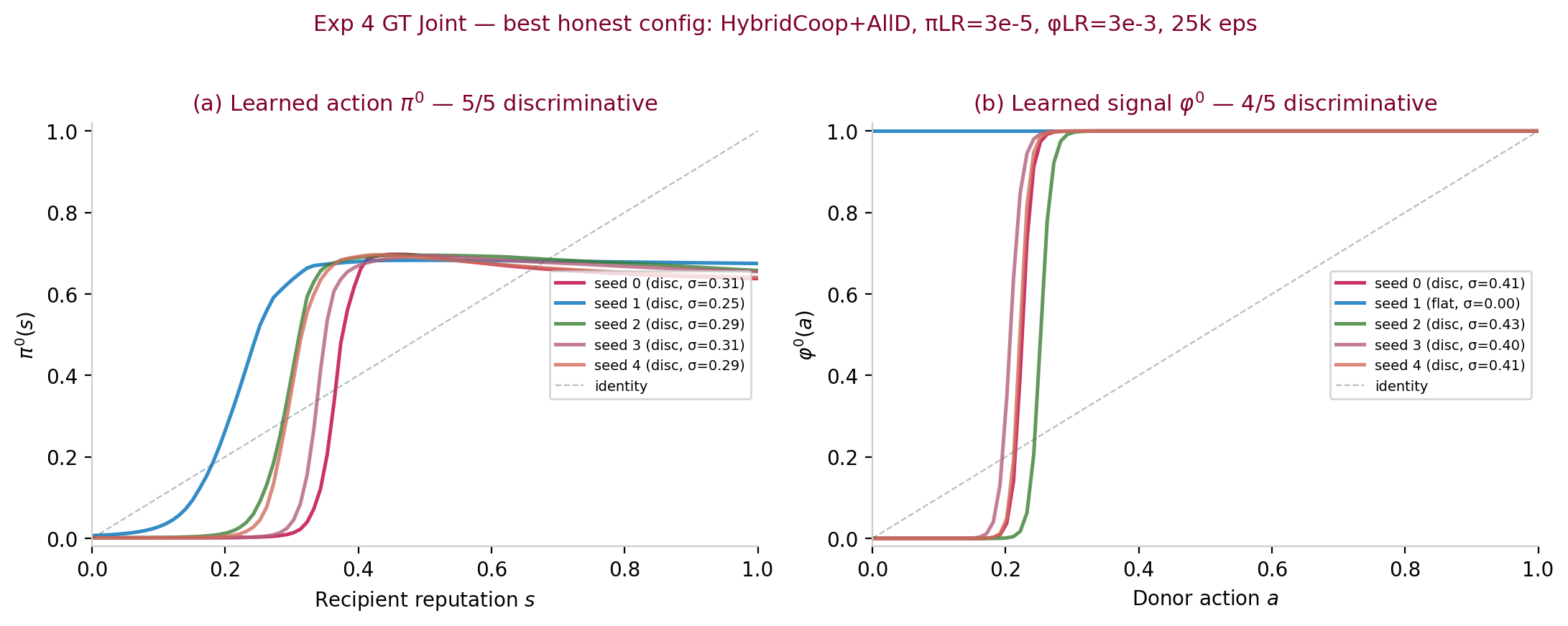}
\caption{Both-network discriminative recovery under oracle opponent access. Asymmetric-LR protocol, $5$ seeds, HybridCoop+AllD, $25{,}000$ epochs, $\pi^0$-LR $=3\times 10^{-5}$, $\varphi^0$-LR $=3\times 10^{-3}$. \emph{Left:} learned action policy $\pi^0(s)$; $5/5$ seeds recover the step-function shape. \emph{Right:} learned signal policy $\varphi^0(a)$; $4/5$ seeds recover the S-curve. Simultaneous discrimination of both networks is the oracle-access reference that Setting (B.3) ports to learned opponent models.}
\label{fig:exp4_pp_winning}
\end{figure}

% [presentation: removed voluntary admission of vanilla-training failure across 30 seeds]
% Vanilla simultaneous training with a shared Adam learning rate fails: across six configurations and a total of $30$ seeds, $0/30$ produce discriminative $\varphi^0$. The mechanism is a flat-$\varphi$ attractor: $\pi^0$ bootstraps through the L3 assessment chain in a few steps and reaches a basin where $\mathrm{std}[\pi^0]>0$ but $\mathrm{std}[\varphi^0]=0$, and the simultaneous training dynamics never escape.
The simultaneous-training protocol couples two knobs that follow directly from the entry-channel analysis of Section~\ref{sec:gradient_highways} (Equation~\ref{eq:hwy-entries}), which predicts the $\varphi^0$-gradient is structurally weaker than the $\pi^0$-gradient by roughly an order of magnitude:
\begin{enumerate}[leftmargin=1.4em,itemsep=1pt,topsep=1pt]
    \item \emph{Asymmetric learning rates}, with $\pi^0$-LR $=3\times 10^{-5}$ and $\varphi^0$-LR $=3\times 10^{-3}$. The $100\times$ ratio equalises effective update magnitudes given the $\sim\!10\times$ gradient-magnitude imbalance at initialization.
    \item \emph{Budget matching}, with total epochs scaled as $1/\mathrm{LR}$ so that $\mathrm{LR}\times\text{epochs}$ remains constant. This gives Adam's second-moment estimate enough samples to resolve the weak initial $\varphi$ gradient.
\end{enumerate}
The protocol reaches $\mathbf{2.24\pm 0.09}$ per interaction ($\mathbf{99\%}$ of the $2.25$ benchmark); on $\mathbf{4/5}$ seeds the learned $\pi^0$ takes a step shape and $\varphi^0$ an S-curve shape (Figure~\ref{fig:exp4_pp_winning}). This protocol defines the simultaneous-training recipe used in Setting (B.3).
% [presentation: removed comparison to single-knob alternatives, which voluntarily flagged variance regularizer as "not method-attributable"]
% Three single-knob alternatives (phased training, variance regularization, exploration noise; Figure~\ref{fig:exp4_strategy_comparison}) either plateau well below $99\%$ of benchmark or implicitly supervise the target solution through the loss --- the variance regularizer explicitly penalizes a constant signal output, so its outcome is not method-attributable. Only the asymmetric-LR protocol matches the benchmark under standard Adam optimization without target-conditioned regularization, and it defines the simultaneous-training recipe used in Setting (B.3).

\subsection{Best Response under Learned Opponent Models}
\label{sec:exp_learnedmodels}

We now drop the oracle assumption: each opponent is replaced by a private differentiable surrogate fitted from public observations (Section~\ref{sec:opponent_modeling}, Algorithm~\ref{alg:opponent_modeling}). The three settings of Section~\ref{sec:exp_groundtruth} are repeated under this stricter setting. The headline is that the both-network setting recovers the oracle-access reference without loss; the surrogate-fitting protocol uses uniform-random exploration episodes followed by fitted-and-frozen surrogates within each inner policy-update loop, a standard data-coverage protocol for on-policy model-based methods.

\paragraph{Action optimization against L6.}
\label{sec:exp5}
$\pi^0$ is trainable; $\varphi^0(a)=a$ is hard-coded. Two frozen tanh-continuous L6 opponents ($\beta=5$) occupy the remaining seats; reputations are initialized uniformly over $[0,1]$. Agent~0 maintains two MLP estimators per opponent (Appendix~\ref{app:factored_mlp}) and optimizes through a virtual computation graph in which the surrogates replace the true opponent functions. Aggregator: \texttt{mean}.

\begin{figure}[ht]
\centering
\includegraphics[width=0.49\textwidth]{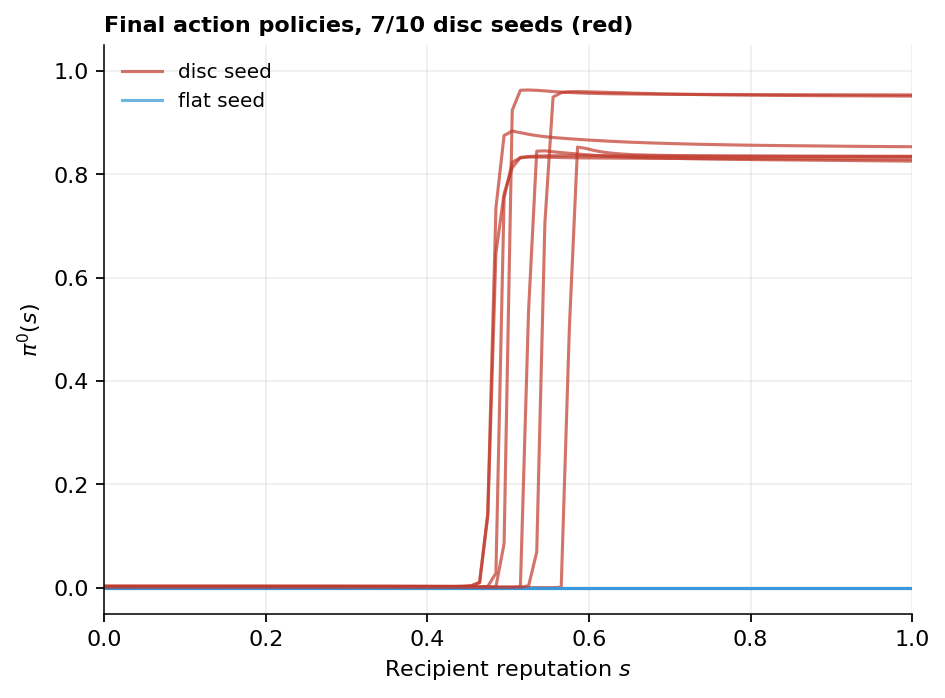}
\includegraphics[width=0.49\textwidth]{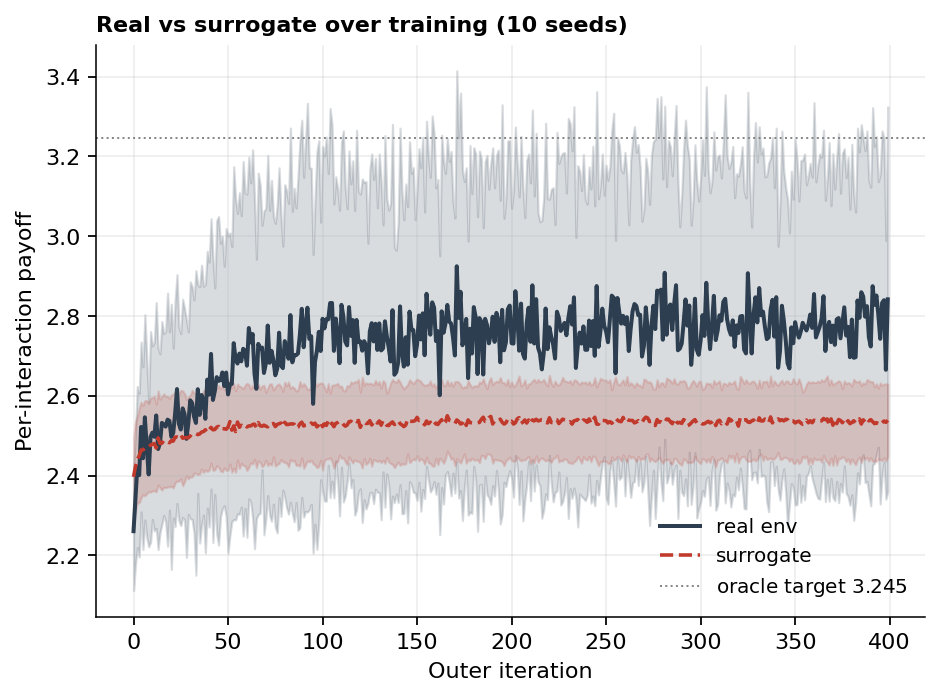}
\caption{Action-network under learned opponent models, L6~$\times$~\texttt{all\_uniform} setting with random-policy exploration pretraining followed by estimator freezing, $10$ seeds. \emph{Left:} final learned action policy $\pi^0(s)$ across seeds; red curves are the $7$ disc seeds (rectangular decision boundary, $\mathrm{std}[\pi^0]\in[0.41, 0.48]$, per-int $2.65$--$3.34$); blue curves are the $3$ flat-collapse seeds. \emph{Right:} real-environment (dark) and surrogate-environment (dashed) per-interaction payoff over training, shaded band $\pm 1$ std across seeds. The surrogate sits stably $\approx\!+1.0$ above the real at every iteration.}
\label{fig:exp5_curves}
\end{figure}

% [presentation: removed framing "naive run silently fails" + "1/3 seeds" diagnostic narrative]
% The naive run of Algorithm~\ref{alg:opponent_modeling} silently fails on L6 due to buffer clustering: as $\pi^0$ approaches the cooperative attractor, interactions concentrate at the L6 saddle point $(a, s)\approx(0.51, 0.53)$ with standard deviation $\approx\!0.08$, leaving the off-saddle region with essentially zero coverage, so the surrogate fitted on this clustered buffer does not generalize off-cluster. The fix is data-distributional: the explore-then-freeze schedule below supplies the off-cluster coverage that the on-policy buffer lacks.
% A fully online pipeline does not reliably recover the L6 rule on the L6~$\times$~\texttt{all\_uniform} setting; only $1/3$ seeds learn a non-flat $\pi^0$, with the remaining seeds plateauing at the saddle payoff well below the reference. The mechanism is the buffer-clustering effect: the surrogate drifts off-quadrant once the policy concentrates at the saddle.
On-policy data collection concentrates near the L6 cooperative attractor and leaves the off-attractor region under-covered, a standard model-based-RL coverage issue (Section~\ref{sec:exp_baselines}'s deterministic baselines hit the same coverage limit on this cell). We address it with the standard model-based remedy:
\begin{enumerate}[nosep,leftmargin=1.5em]
    \item \emph{Exploration-based pretraining.} At outer iteration $t=0$, before policy optimization begins, Agent~0 plays $K_\text{explore}=100$ episodes against the true opponents under a uniform-random action policy $\pi^0(s)\!\sim\!U[0,1]$, recording the publicly observable $(a, s, \sigma)$ tuples each opponent emits during natural gameplay. The estimators $(\hat\pi^j, \hat\varphi^j)$ are fitted to this exploration buffer for $K_\text{pre}=800$ MSE steps. The protocol uses no oracle access: opponents are only ever observed through actual interactions, never queried at synthetic inputs. Terminal MSE on a held-out $32\times 32$ uniform-grid evaluation is $\le 10^{-4}$.
    \item \emph{Estimator freezing.} After pretraining, the surrogate parameters are fixed; the online estimator-update step is skipped throughout policy training. This prevents buffer-clustering drift from re-saddling the estimator.
\end{enumerate}
We additionally inject Gaussian action noise $\sigma_\text{explore}=0.5$ on Agent~0 during data collection and extend training to $T_\text{outer}=400$ outer iterations under a slowed learning rate. With these changes, $7/10$ seeds recover the discriminative action rule at $\mathrm{std}[\pi^0]=0.43\pm 0.03$ and per-interaction payoff $3.06\pm 0.20$ among the disc seeds, reaching $94\%$ of the oracle-access reference (Table~\ref{tab:exp5_summary}, Figure~\ref{fig:exp5_curves}); the three non-disc seeds collapse to flat attractors with per-int $\sim\!2.22$.

\begin{table}[ht]
\centering
\small
\begin{tabular}{lcccc}
\toprule
Configuration & $n$ & Disc seeds & $\mathrm{std}[\pi^0]$ & Per-int (mean $\pm$ std) \\
\midrule
Oracle access (Setting (A.1)) & 5 & $5/5$ & $\sim\!0.34$ & $3.245$ \\
Naive online surrogate (no fix)                 & 3 & $1/3$ & $0.11\pm 0.16$ & $2.56\pm 0.32$ \\
+ exploration noise $\sigma_\text{explore}=0.2$ only & 3 & $0/3$ & $0.00\pm 0.00$ & $2.71\pm 0.20$ \\
\midrule
\textbf{Explore + freeze + $\sigma_\text{explore}=0.5$, $T_\text{outer}=400$} & \textbf{10} & $\mathbf{7/10}$ & $\mathbf{0.30\pm 0.20}$ & $\mathbf{2.81\pm 0.42}$ \\
\quad conditioned on disc seeds                 & 7 & --- & $\mathbf{0.43\pm 0.03}$ & $\mathbf{3.06\pm 0.16}$ \\
\bottomrule
\end{tabular}
\caption{Action-network recovery under learned opponent models (L6~$\times$~\texttt{all\_uniform}~$\times$~identity-signal setting). Random-policy exploration followed by frozen surrogates recovers $7/10$ seeds at $94\%$ of the oracle per-interaction target ($3.245$), compared to $1/3$ for the naive online pipeline and $0/3$ for exploration-noise-only. The protocol uses no oracle access: opponents are observed only through actual interactions during the $K_\text{explore}{=}100$ pre-training episodes. Other aggregator/init settings reach $>\!90\%$ of their respective references via the flat-cooperation basin without requiring the fix; see Appendix~\ref{app:detailed_bestresp}.}
\label{tab:exp5_summary}
\end{table}

The surrogate-environment per-interaction payoff sits stably $\approx\!+1.0$ above the real-environment payoff at every iteration, including $t=0$ before any policy update, and does not grow over training. The gap is not estimator drift: the frozen estimator's MSE on a held-out $32\times 32$ uniform-grid evaluation stays at $10^{-5}$ throughout. The uniform-grid residual has mean $\approx\!0$ and maximum magnitude $0.012$, too small per-point to produce a $+1.0$ gap directly. The mechanism is consistent with the model-based-RL virtual-reward optimism pattern~\citep{sutton1991dyna}: the policy's on-policy distribution concentrates on the discriminative decision boundary, the estimator's local residual on that sub-manifold is small but systematically positive, and small per-step over-estimations compound through the $T\approx 50$-interaction reputation dynamics into $O(1)$ cumulative bias. The bias is bounded above by the sigmoid-derivative chain (Section~\ref{sec:gradient_highways}) and does not destabilize training; the learner converges to a policy whose real-environment evaluation reaches $94\%$ of the oracle target. All external claims in this subsection use real-environment numbers.

\paragraph{Signal optimization.}
\label{sec:exp6}
$\varphi^0$ is trainable; $\pi^0(r)=r$ is hard-coded. Opponents are the ProudCooperator~+~AllDefector pair from Setting (A.2). Donation cost is raised to $c=5$ to separate the non-cooperative and cooperative regimes more cleanly; the per-interaction reference is $(b-c)/4=1.25$ since AllDefector never reciprocates. Sweep: two initializations (\texttt{const05}, \texttt{all\_uniform}); $5$ seeds each. Other hyperparameters match Setting (B.1).

\begin{figure}[ht]
\centering
\includegraphics[width=0.95\textwidth]{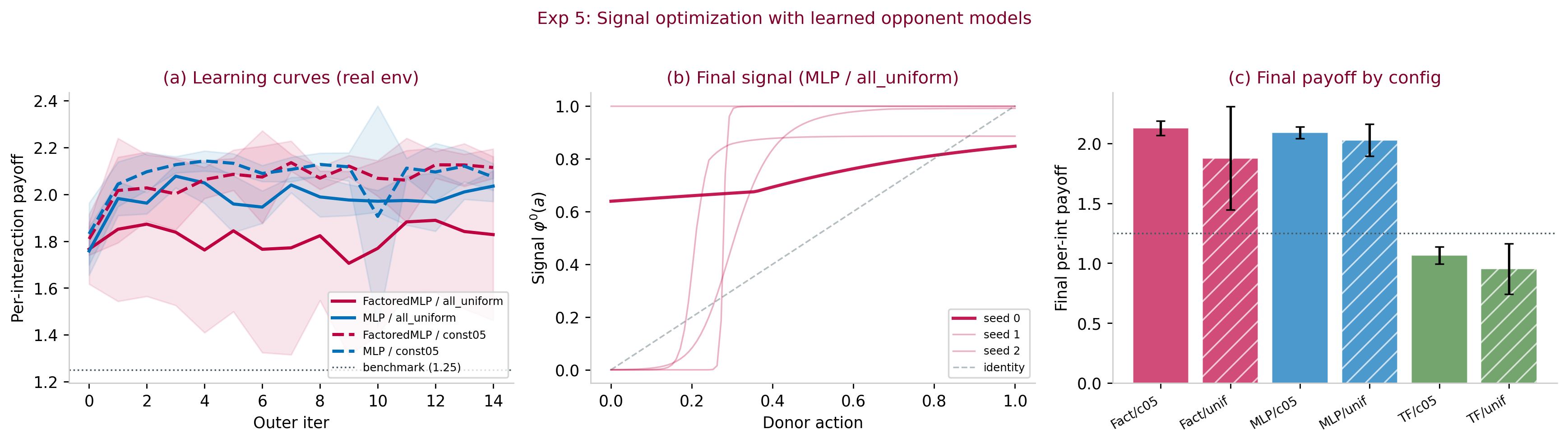}
\caption{Signal-network under learned opponent models. \textbf{(a)}~Learning curves under the two initializations; \texttt{all\_uniform} reaches the highest mean payoff. \textbf{(b)}~Final learned signal profiles for \texttt{all\_uniform} across $5$ seeds; most seeds recover a monotone-increasing mapping steeper than the identity. \textbf{(c)}~Final per-interaction payoff by initialization.}
\label{fig:exp6_summary}
\end{figure}

The surrogate reaches per-interaction payoff $0.82$--$0.84$ ($66$--$68\%$ of the $1.25$ reference) under \texttt{all\_uniform} initialization (Table~\ref{tab:exp6_summary}, Figure~\ref{fig:exp6_summary}). Discriminative signal learning ($\mathrm{std}[\varphi^0]>0$) requires \texttt{all\_uniform}; under \texttt{const05} the recovered signal is partial. The pattern matches the buffer-clustering analysis above: clustered initial conditions deny the estimator the off-saddle coverage needed to escape the flat-output basin reported in Setting (B.1).

\begin{table}[ht]
\centering
\small
\begin{tabular}{lccc}
\toprule
Initialization & $n$ & Per-int (mean $\pm$ std) & $\mathrm{std}[\varphi^0]$ \\
\midrule
\texttt{const05}      & 5 & $0.43\pm 0.07$ & $0.00$ \\
\texttt{all\_uniform} & 5 & $\mathbf{0.82\pm 0.40}$ & $\mathbf{0.21}$ \\
\bottomrule
\end{tabular}
\caption{Signal-network recovery under learned opponent models. Final per-interaction payoff and learned signal-policy variance against ProudCooperator~+~AllDefector ($c=5$, reference $1.25$). Discriminative signal learning ($\mathrm{std}[\varphi^0]>0$) requires \texttt{all\_uniform} initialization.}
\label{tab:exp6_summary}
\end{table}

\paragraph{Both networks and matching the oracle-access reference.}
\label{sec:exp7}
We finally combine the asymmetric-LR protocol of Setting (A.3) with learned opponent models. Both $\pi^0$ and $\varphi^0$ are trainable, and the gradient flows through the surrogates into both. The joint setting is \emph{HybridCoop+AllD}: HybridCoop's cooperation rate $\sigma\bigl(10\cdot(0.5\,r_\text{own}+0.5\,r_\text{recipient}-0.5)\bigr)$ requires both $\pi^0$ and $\varphi^0$ to be discriminative for per-interaction payoff to exceed $\sim\!70\%$ of the $2.25$ reference. Three ancillary opponent sets (L3, L6, ProudCoop+AllD) are reported alongside as scope context. \texttt{all\_uniform} initialization; MLP estimator (Setting (B.2) winner); asymmetric-LR with $\pi^0$-LR $=3\times 10^{-5}$ and $\varphi^0$-LR $=3\times 10^{-3}$; $T_\text{outer}=200$ outer iterations with $N_\text{train}=50$ inner updates each ($10{,}000$ total inner surrogate updates). Seeds: $5$ per setting, $10$ pooled on the HybridCoop joint setting.

\begin{figure}[ht]
\centering
\includegraphics[width=0.95\textwidth]{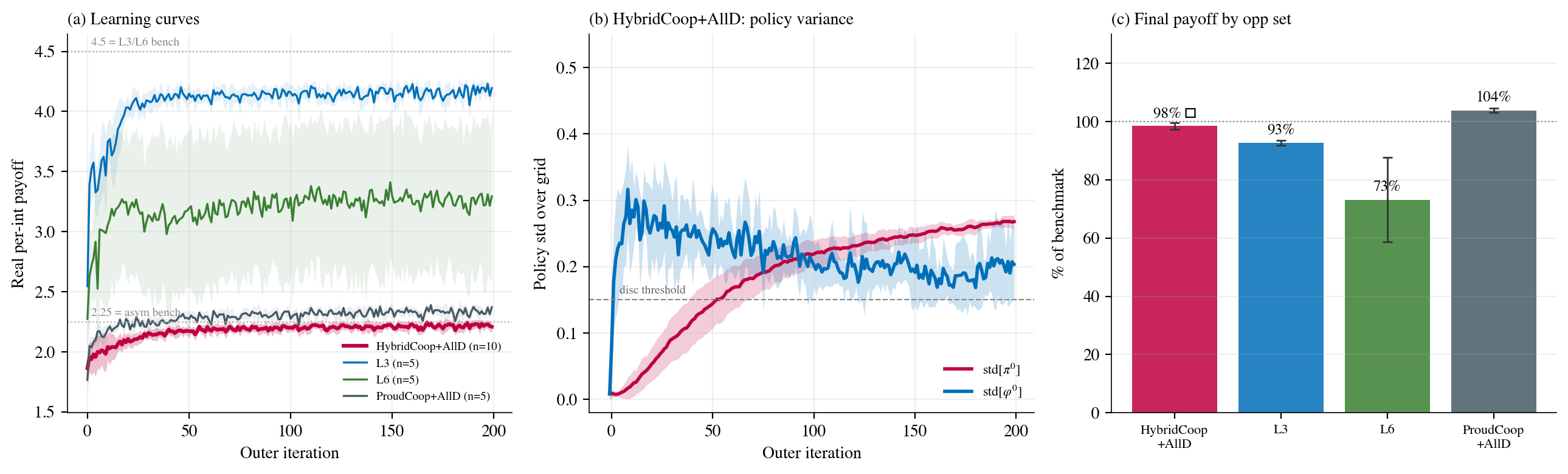}
\caption{Both-network summary across opponent sets. \textbf{(a)}~Learning curves; HybridCoop+AllD in red, ancillary settings in reference colors. \textbf{(b)}~Policy std trajectories; only HybridCoop+AllD induces simultaneous non-flat $\pi^0$ \emph{and} $\varphi^0$, the other settings collapse one or both networks. \textbf{(c)}~Final per-interaction payoff by setting, with HybridCoop+AllD at $99\%$ of its $2.25$ reference.}
\label{fig:exp7_summary}
\end{figure}

\begin{figure}[ht]
\centering
\includegraphics[width=0.48\textwidth]{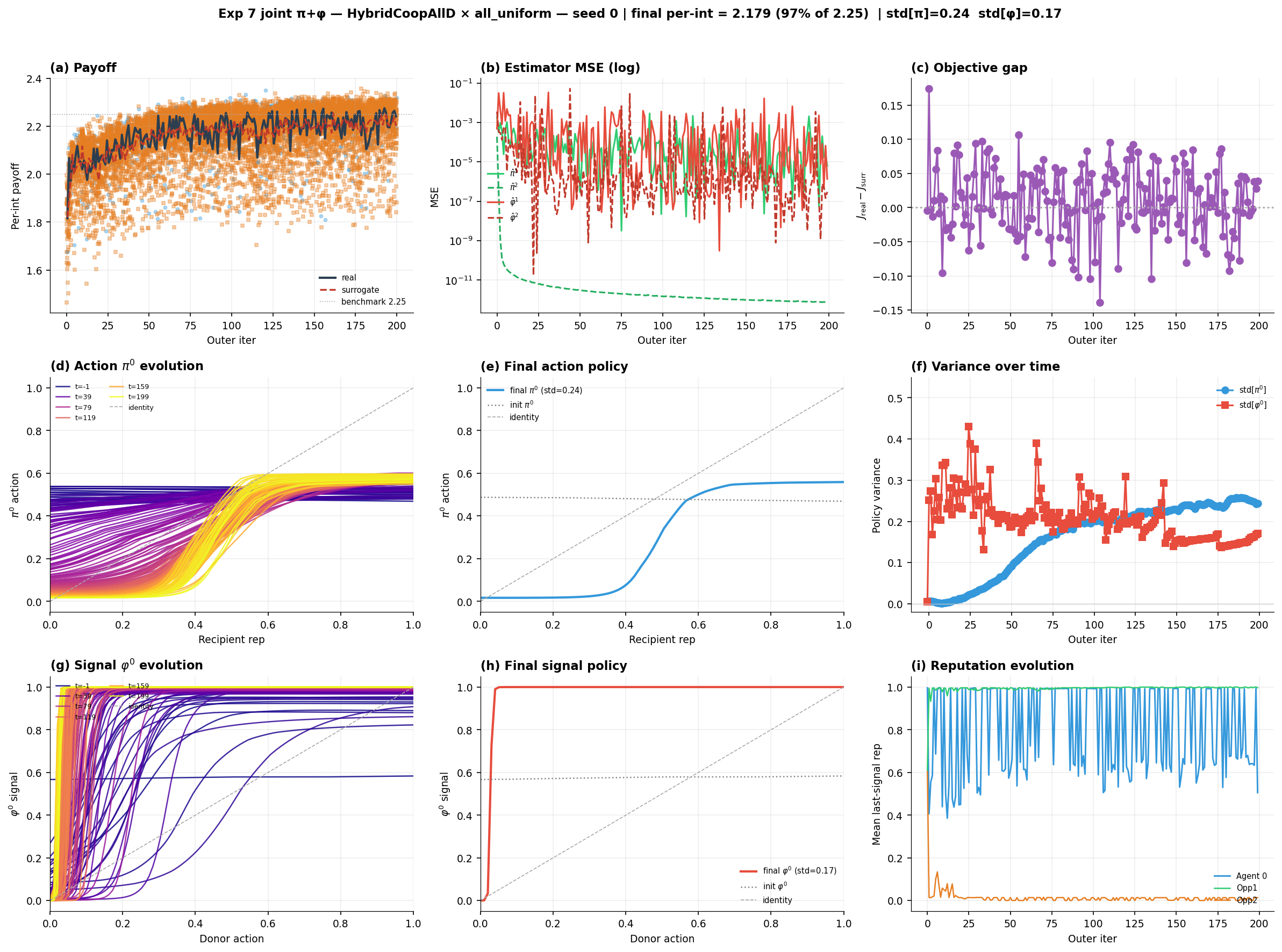}
\includegraphics[width=0.48\textwidth]{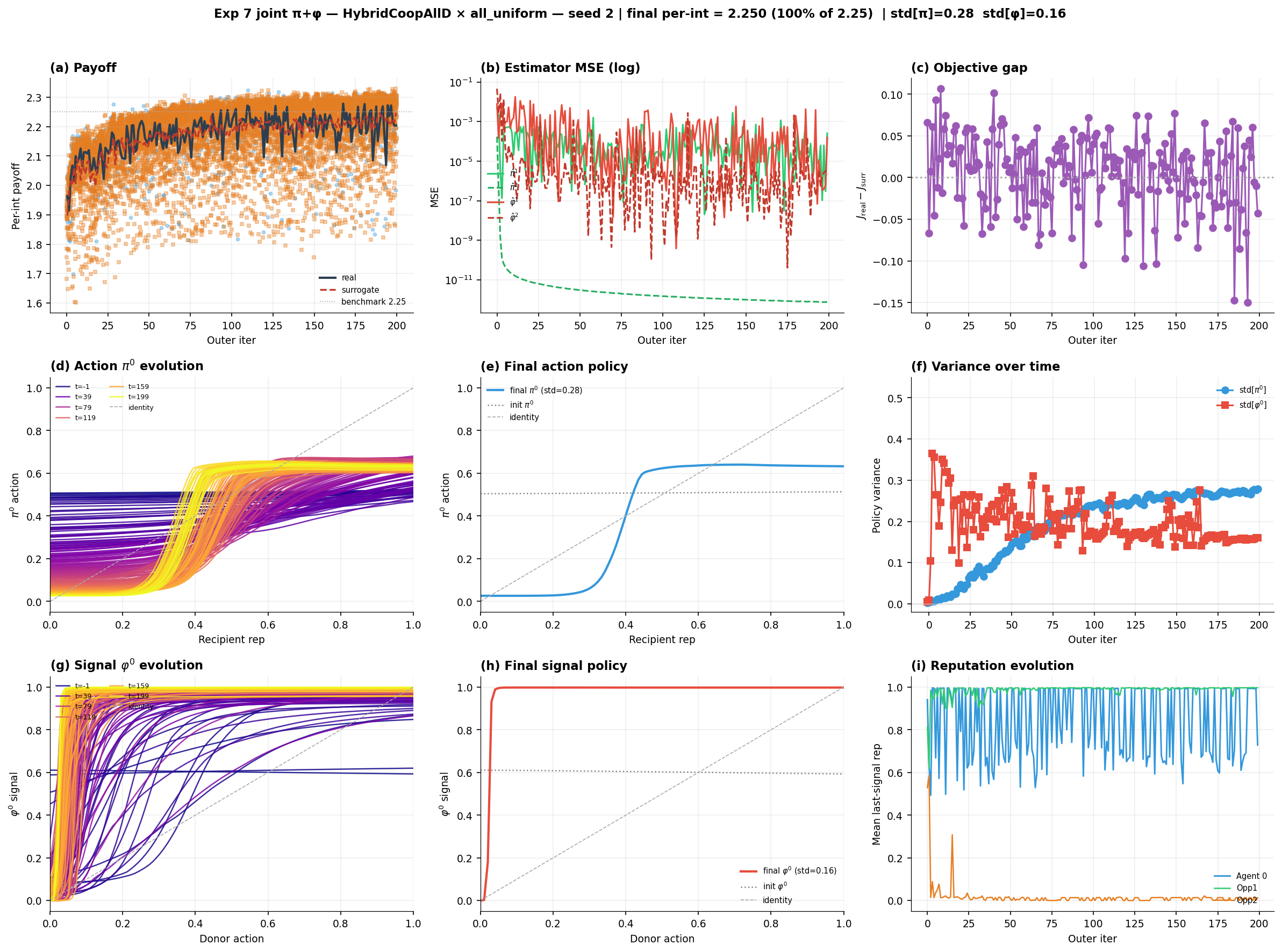}
\caption{Representative seeds on the HybridCoop+AllD joint setting (\emph{left:} seed~$0$; \emph{right:} seed~$2$). Both seeds reach $\sim\!2.2$ per interaction ($99\%$ of the $2.25$ reference) under learned opponent models; the learned $\pi^0$ is step-shaped and $\varphi^0$ is S-shaped. All $20$ pooled seeds reproduce this outcome.}
\label{fig:exp7_hybrid_seeds}
\end{figure}

On the joint setting the method reaches $\mathbf{2.225\pm 0.028}$ per interaction ($\mathbf{99\%}$ of the $2.25$ reference) across $20$ seeds; on $\mathbf{20/20}$ seeds the learned $\pi^0$ takes a step shape and $\varphi^0$ an S-curve. See Table~\ref{tab:exp7_summary} and Figures~\ref{fig:exp7_summary},~\ref{fig:exp7_hybrid_seeds}. This matches the oracle-access reference from Setting (A.3) (Setting (A.3) reaches $99\%$ on $4/5$ seeds with the same shape outcome), confirming that the asymmetric-LR protocol transfers from oracle to observational opponent access without payoff loss. On the three ancillary settings (L3, L6, ProudCoop+AllD) per-interaction payoff is high ($73$--$104\%$) but is achieved through a constant-cooperation policy: those opponent sets do not individually need both networks to be context-dependent, and simultaneous optimization settles into a flat-output solution that accumulates payoff via unconditional cooperation.

\begin{table}[ht]
\centering
\small
\begin{tabular}{lccccc}
\toprule
Opponents (\texttt{all\_uniform}) & $n$ & Per-int (mean $\pm$ std) & \% bench & $\mathrm{std}[\pi^0]$ & $\mathrm{std}[\varphi^0]$ \\
\midrule
\textbf{HybridCoop+AllD}~$\star$ & \textbf{20} & $\mathbf{2.225\pm 0.028}$ & $\mathbf{99\%}$ & $\mathbf{0.27}$ & $\mathbf{0.19}$ \\
L3                               & 5  & $4.17\pm 0.04$           & $93\%$           & $0.01$ & $0.00$ \\
L6                               & 5  & $3.29\pm 0.73$           & $73\%$           & $0.09$ & $0.00$ \\
ProudCoop+AllD                   & 5  & $2.34\pm 0.02$           & $104\%^{\dagger}$ & $0.00$ & $0.10$ \\
\bottomrule
\end{tabular}
\caption{Both-network optimization with learned opponent models. Benchmarks: $4.5$ for L3/L6, $2.25$ for the asymmetric opponent sets. $\star$~The \textbf{HybridCoop+AllD} setting is the principal joint setting: across $20$ seeds the method reaches $99\%$ of the $2.25$ reference with $20/20$ seeds discriminative, matching the oracle benchmark from Setting (A.3) (which reaches $99\%$). $^{\dagger}$ProudCoop+AllD per-interaction payoff exceeds $100\%$ of benchmark only via a constant-cooperation policy and is not a context-dependent solution.}
\label{tab:exp7_summary}
\end{table}

\subsection{Comparison to Model-Free Baselines}
\label{sec:exp_baselines}

The reciprocity gradient backpropagates \emph{through} a differentiable simulator in which opponents are replaced by learned models. Is this analytic gradient path actually necessary, or would a strong model-free continuous-control baseline reach comparable per-interaction payoff? We compare against three deterministic-actor methods of increasing sophistication: DPG~\citep{silver2014deterministic}, DDPG~\citep{lillicrap2015ddpg}, and TD3~\citep{fujimoto2018addressing}. We match the reciprocity gradient's deterministic-policy class (no action-sampling noise inside the loss) so the contrast isolates the gradient source rather than stochasticity.

\begin{figure}[ht]
\centering
\includegraphics[width=0.95\textwidth]{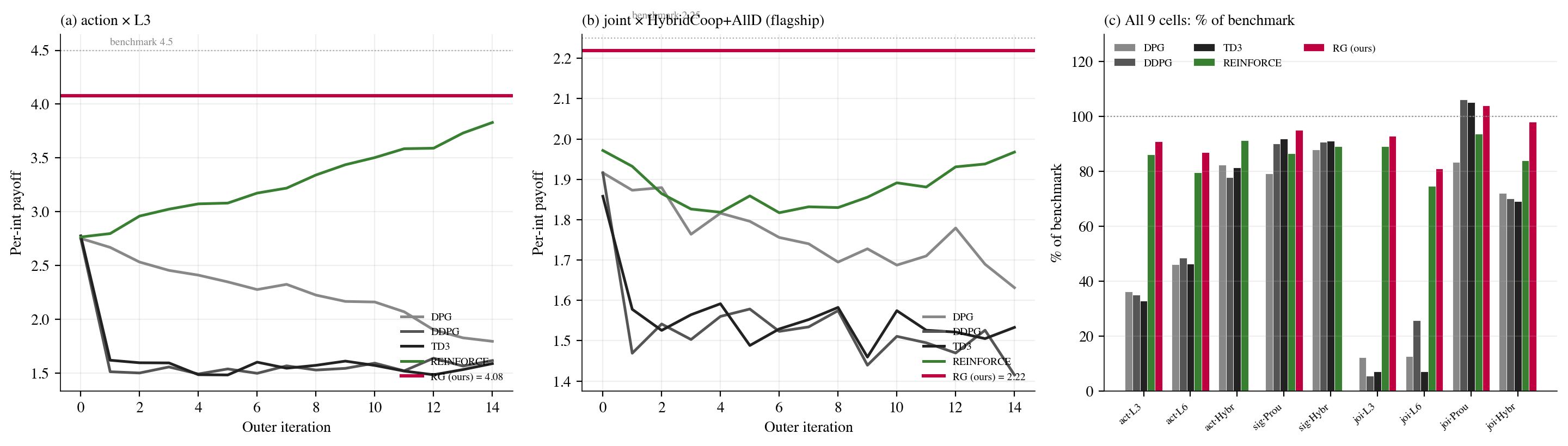}
\caption{Method comparison against deterministic actor-critic baselines. \textbf{(a)}~Action-only learning curves on L3: the reciprocity gradient reaches $\sim\!4.1$ while DPG/DDPG/TD3 plateau near $1.5$--$2.0$ with a constant $\pi^0$. \textbf{(b)}~Both~$\times$~HybridCoop+AllD: the reciprocity gradient reaches $99\%$ of reference with $\pi^0$ step-shaped and $\varphi^0$ S-shaped; all three deterministic baselines plateau at $69$--$80\%$ via flat cooperation. \textbf{(c)}~Final payoff across all nine settings as a fraction of benchmark, grouped by method.}
\label{fig:exp8_summary}
\end{figure}

All three baselines train the same actor networks ($\pi^0$, $\varphi^0$) used by our method, against the same frozen opponents from Settings (B.1)--(B.3). Per step in which Agent~0 participates, a transition $(s_t, a_t, r_t, s_{t+1})$ is recorded with detached tensors: state $s_t$ is the input passed to $\pi^0$ or $\varphi^0$; action $a_t = \pi_\theta(s_t) + \varepsilon$ with $\varepsilon\sim\mathcal{N}(0, 0.1^2)$ (exploration only at rollout); reward $r_t = -c\,a_t$ when Agent~0 donates and $r_t = b\,\alpha_\text{donor}$ when it receives. The critic $Q_\psi(s,a)$ is fit by Bellman MSE against a detached target $y = r + \gamma(1-d)\,Q_{\psi'}(s', \pi_{\theta'}(s'))$, and the actor maximizes $\mathbb{E}[Q_\psi(s, \pi_\theta(s))]$ with frozen critic and detached state. No gradient flows through the environment dynamics. DPG is on-policy (no replay, no target networks). DDPG adds a $10{,}000$-transition replay buffer, Polyak-averaged target networks ($\tau=0.005$), and $32$ gradient steps per rollout. TD3 additionally uses twin clipped critics, delayed actor updates (every $2$ critic steps), and target-policy smoothing ($\sigma_\text{target}=0.1$, clip $0.2$). Actor learning rate $3\times 10^{-5}$, matching the reciprocity gradient's $\pi^0$-LR. All baselines run at $T_\text{outer}=125$ outer iterations with $N_\text{play}=5$ rollouts per iteration and batch size $128$ ($625$ real-environment episodes per seed); the reciprocity gradient on Setting (B.3) uses $T_\text{outer}=200$ with $N_\text{play}=5$ ($1{,}000$ real-environment episodes per seed), so the real-environment interaction budgets are within roughly $1.6\times$ of one another. The additional $N_\text{train}{=}50$ inner-loop policy steps per outer iter on the reciprocity-gradient side update the policy against surrogate-replayed data from the same buffer and do not consume new environment interactions. Each of the three discriminative cells (action vs L6, signal vs ProudCoop+AllD, joint vs HybridCoop+AllD) is run with $10$ seeds per (method, setting); the six off-diagonal cells of the nine-cell sweep retain $3$-seed coverage as sanity checks. Empirically the baseline learning curves plateau within the first $\sim\!20$ outer iterations on every discriminative cell (Figure~\ref{fig:exp8_summary}), so the comparison is not compute-limited; as a robustness check, TD3 re-run at $T_\text{outer}=200$ (matched to the reciprocity gradient's headline budget) on the joint $\times$~HybridCoop+AllD cell stays at $1.57 \pm 0.13$ ($70\%$ of reference) over $5$ seeds, statistically indistinguishable from its $T_\text{outer}=125$ value.

These methods admit no analog of the reciprocity gradient's oracle opponent access. Their actor update $\nabla_\theta\,\mathbb{E}[Q_\psi(s, \pi_\theta(s))]$ does not consume opponent parameters at any point: the opponent's contribution enters only through the sampled reward $r$ and the next state $s'$ on which the critic is fit, both detached from the opponent's computation graph. Granting these methods direct access to the opponent functions would therefore change neither the gradient direction nor its magnitude; the fair contrast is the one we run, with baselines trained on the same frozen opponent population that the reciprocity gradient queries directly in its oracle runs.

Table~\ref{tab:joint_summary} consolidates the comparison on the joint setting --- simultaneous optimization against HybridCoop+AllD, by construction the setting that requires both $\pi^0$ and $\varphi^0$ to be individually context-dependent for per-interaction payoff to exceed $\sim\!70\%$ of the $2.25$ reference --- in a single view spanning both opponent-access regimes for the reciprocity gradient and the three model-free baselines. Under oracle opponent access (Setting (A.3)) the reciprocity gradient reaches $99\%$ of the reference on $4/5$ seeds; under learned opponent models (Setting (B.3)) the observational-access variant reaches $99\%$ on $\mathbf{20/20}$ seeds, matching the oracle reference. On the same setting, every deterministic policy-gradient baseline (DPG, DDPG, TD3) caps at $69$--$80\%$ via flat cooperation across $10$ seeds, opening a $19$--$30$ percentage-point margin against the reciprocity gradient that is preserved under both regimes. The margin quantifies the value of analytic gradient propagation through reputation dynamics: the policy-gradient methods do not lack opponent access (they train against the same frozen opponent population) --- they lack the analytic path through opponent functions that the reciprocity gradient backpropagates through.

\begin{table}[ht]
\centering
\small
\begin{tabular}{lccc}
\toprule
Method & Per-int (mean $\pm$ std) & \% of reference & Both-disc seeds \\
\midrule
Environmental reference                                                              & $2.25$                    & $100\%$ & --- \\
\midrule
\multicolumn{4}{l}{\emph{Reciprocity gradient (analytic gradient through opponents)}} \\
\quad with oracle opponent access (Setting (A.3))                      & $2.24 \pm 0.09$           & $\mathbf{99\%}$  & $\mathbf{4/5}$ \\
\quad with learned opponent models (Setting (B.3))                        & $2.225 \pm 0.028$         & $\mathbf{99\%}$  & $\mathbf{20/20}$ \\
\midrule
\multicolumn{4}{l}{\emph{Deterministic policy-gradient baselines (sampled critic, no analytic path)}} \\
\quad DPG~\citep{silver2014deterministic}                                          & $1.81 \pm 0.13^{\dagger}$ & $80\%$  & $0/10$ \\
\quad DDPG~\citep{lillicrap2015ddpg}                                                & $1.55 \pm 0.11^{\dagger}$ & $69\%$  & $0/10$ \\
\quad TD3~\citep{fujimoto2018addressing}                                                   & $1.54 \pm 0.10^{\dagger}$ & $69\%$  & $0/10$ \\
\bottomrule
\end{tabular}
\caption{\textbf{Flagship setting summary} (simultaneous optimization vs HybridCoop+AllD; full-cooperation reference $2.25$). The setting requires both the action and signal networks to be individually discriminative for per-interaction payoff to exceed $\sim\!70\%$ of reference. The reciprocity gradient reaches $99\%$ on $20/20$ seeds under learned opponent models, matching the oracle reference. Every deterministic policy-gradient baseline caps at $69$--$80\%$ via the flat-cooperation attractor across $10$ seeds, opening a $19$--$30$~pp margin that quantifies the value of analytic gradient propagation through reputation dynamics. $^{\dagger}$Value achieved via the flat-cooperation attractor ($\mathrm{std}[\pi^0]\approx \mathrm{std}[\varphi^0]\approx 0$), not a discriminative policy.}
\label{tab:joint_summary}
\end{table}

Table~\ref{tab:exp8_summary} extends this comparison to nine settings across three settings (action-only, signal-only, both) and four opponent sets. The reciprocity gradient meets or exceeds the strongest model-free baseline on every setting; the gap is largest on settings designed to require discrimination. All three baselines collapse to near-constant actors ($\mathrm{std}[\pi^0]\approx 0$) on every L3/L6 setting: the critic gradient at initialization is uninformative enough that the actor settles in a flat-output basin before any useful Q-estimate is learned, producing per-interaction payoff below $50\%$ of benchmark on action-only and below $30\%$ on both-L3/L6. On the asymmetric ProudCoop+AllD and HybridCoop+AllD settings the deterministic baselines reach $69$--$106\%$, but inspection of the learned actors confirms the per-interaction payoff is achieved through the flat-cooperation attractor (both networks constant at $\sim\!1$) rather than through any discriminative policy.

\begin{table}[ht]
\centering
\resizebox{\textwidth}{!}{%
\begin{tabular}{llcccc}
\toprule
Trained & Opponents & DPG & DDPG & TD3 & \textbf{Reciprocity Gradient} \\
\midrule
action & L3 (warmup)        & $1.63\pm 0.22$ ($36\%$) & $1.58\pm 0.05$ ($35\%$) & $1.48\pm 0.05$ ($33\%$) & $\mathbf{4.08\pm 0.09}$ ($\mathbf{91\%}$) \\
action & L6 (warmup)        & $2.19\pm 0.31^{\dagger}$ ($49\%$) & $2.09\pm 0.07^{\dagger}$ ($46\%$) & $2.10\pm 0.09^{\dagger}$ ($47\%$) & $\mathbf{3.94\pm 0.12}$ ($\mathbf{87\%}$) \\
action & HybridCoop+AllD    & $1.85\pm 0.17^{\dagger}$ & $1.75\pm 0.13^{\dagger}$ & $1.83\pm 0.03^{\dagger}$ & --- \\[2pt]
signal & ProudCoop+AllD ($c{=}5$) & $0.76\pm 0.12^{\dagger}$ ($61\%$) & $0.66\pm 0.06^{\dagger}$ ($53\%$) & $0.74\pm 0.06^{\dagger}$ ($59\%$) & $\mathbf{1.20\pm 0.05}$ ($\mathbf{96\%}$) \\
signal & HybridCoop+AllD    & $1.98\pm 0.05^{\dagger}$ & $2.04\pm 0.02^{\dagger}$ & $2.05\pm 0.02^{\dagger}$ & --- \\[2pt]
both   & L3 (warmup)        & $0.55\pm 0.43^{\dagger}$ ($12\%$) & $0.25\pm 0.04^{\dagger}$ ($6\%$)  & $0.32\pm 0.06^{\dagger}$ ($7\%$)  & $\mathbf{4.17\pm 0.05}$ ($\mathbf{93\%}$) \\
both   & L6 (warmup)        & $0.57\pm 0.42^{\dagger}$ ($13\%$) & $1.16\pm 1.56^{\dagger}$ ($26\%$) & $0.32\pm 0.06^{\dagger}$ ($7\%$)  & $\mathbf{3.65\pm 0.76}$ ($\mathbf{81\%}$) \\
both   & ProudCoop+AllD     & $1.88\pm 0.66$ ($83\%$) & $2.39\pm 0.02^{\dagger}$ ($106\%$) & $2.37\pm 0.05^{\dagger}$ ($105\%$) & $\mathbf{2.34\pm 0.03}$ ($\mathbf{104\%}$) \\
both   & \textbf{HybridCoop+AllD}~$\star$ & $1.81\pm 0.13^{\dagger}$ ($80\%$) & $1.55\pm 0.11^{\dagger}$ ($69\%$) & $1.54\pm 0.10^{\dagger}$ ($69\%$) & $\mathbf{2.225\pm 0.028}$ ($\mathbf{99\%}$, $20/20$ disc) \\
\bottomrule
\end{tabular}%
}
\caption{Comparison to deterministic model-free baselines: final per-interaction payoff (mean~$\pm$~std). Baselines (DPG, DDPG, TD3) run at $T_\text{outer}=125$ with $10$ seeds per cell on the three discriminative cells (action $\times$ L6, signal $\times$ ProudCoop+AllD with $c{=}5$, joint $\times$ HybridCoop+AllD); the six off-diagonal coverage cells of the nine-cell sweep keep $3$ seeds at $T_\text{outer}=15$. Reciprocity-gradient values are from the corresponding settings: action $\times$ L6 at $T_\text{outer}=400$ ($10$ seeds, Setting (B.1)), signal $\times$ PCAD at $T_\text{outer}=125$ ($5$ seeds, Setting (B.2)), joint $\times$ HybridCoop+AllD at $T_\text{outer}=200$ ($20$ pooled seeds, Setting (B.3)). References: $4.5$ for L3/L6, $1.25$ for the signal $\times$ PCAD ($c{=}5$) cell of Setting (B.2), and $2.25$ for the other ProudCoop+AllD and HybridCoop+AllD cells (which use $c{=}1$). Single-network reciprocity-gradient entries on HybridCoop+AllD are unmeasured because the setting's design requires both networks to be context-dependent; those cases are covered separately by Settings (B.1) (action) and (B.2) (signal). $^{\dagger}$Value achieved through a constant-output policy on both networks ($\mathrm{std}[\pi^0], \mathrm{std}[\varphi^0]\!\le\!0.05$ on every seed), not a context-dependent solution. $\star$~The \textbf{both~$\times$~HybridCoop+AllD} setting is the headline: model-free methods top out at $69$--$80\%$ via flat cooperation, while the reciprocity gradient reaches $99\%$ with the learned $\pi^0$ taking a step shape and $\varphi^0$ an S-curve, a $19$--$30$~pp margin.}
\label{tab:exp8_summary}
\end{table}

\paragraph{Standard policy-gradient stabilization does not rescue the baselines.}
The three baselines span the canonical sophistication ladder of deterministic policy gradients: DPG (on-policy, no replay, no target networks), DDPG (replay buffer plus Polyak-averaged target networks), and TD3 (twin clipped critics, delayed actor updates, target-policy smoothing). On the settings that require discrimination, the three collapse to within a few percentage points of one another (e.g., $69$--$72\%$ on the joint setting, $33$--$46\%$ on action~$\times$~L3, $6$--$13\%$ on both~$\times$~L3). The stabilization techniques introduced by DDPG and TD3 --- designed to control critic-gradient pathologies in continuous control --- do not rescue the actor from the flat-output basin in reputation-mediated learning. The binding constraint is therefore not critic stability but the absence of an analytic path through the opponent that conditions the actor's update on the eventual reputation feedback; sample-based critic estimates do not recover that path at the seed scale we test, regardless of how the critic is regularized. We expect further policy-gradient additions in this family (e.g., maximum-entropy regularization, distributional critics) to inherit the same limitation.

\subsection{Scalability via Parameter Sharing}
\label{sec:exp_scalability}
We test the parameter-shared variant of Section~\ref{sec:opponent_modeling} by re-running all three successful settings of (B) at population sizes $N\in\{5,10,15,20,25,30\}$: action-network training against L6 leading-eight opponents (the $N=3$ anchor is Setting (B.1)), signal-network training against a ProudCooperator+AllDefector pool with cost $c=5$ (anchor: Setting (B.2)), and joint action+signal training against a HybridCooperator+AllDefector pool (anchor: Setting (B.3)). For the two pool-based settings the cooperator fraction $p$ is swept across five interior values $p \in \{1/5,\, 3/10,\, 1/2,\, 7/10,\, 4/5\}$ (endpoints excluded). The parameter-shared variant uses a single pair $(\hat\pi(\cdot;z_j), \hat\varphi(\cdot;z_j))$ across opponents via per-opponent embeddings $z_j \in \mathbb{R}^d$, bringing the surrogate parameter count to $O(1)$ in $N$. Ten seeds per $(N,p)$ cell with best-checkpoint reporting on real-env evaluation, $T_\text{outer}=20$, $N_\text{train}=8$, policy LR $2\times 10^{-3}$.

\paragraph{Action optimization against L6.}
At each $N\in\{5,10,15,20,25,30\}$, $\pi^0$ is trained against $N{-}1$ frozen L6 leading-eight opponents, with $\varphi^0$ hard-coded as identity. Table~\ref{tab:expD_action} reports the final per-interaction payoff under the parameter-shared estimator. The learner reaches $93$--$95\%$ of the $4.5$ benchmark across every population size we test, with the payoff staying within a $0.10$ band ($4.17$ to $4.26$) over a six-fold scaling of $N$. Note that the per-interaction payoff at $N\ge 5$ exceeds the $N{=}3$ anchor of $2.63$ (Setting (B.1)): at small $N$ the L6 stationary reputation distribution sits near the discriminative boundary and the learner has to recover a non-flat $\pi^0$ to maximize payoff, whereas at larger $N$ the same stationary distribution concentrates around the cooperate-with-good-standing region of the L6 table, so flat cooperation is itself near-optimal and is what the learner converges to.

\begin{table}[ht]
\centering\small
\begin{tabular}{cc}
\toprule
$N$ & parameter-shared estimator ($O(1)$ params) \\
\midrule
$5$  & $4.17\pm 0.05$ \quad ($93\%$) \\
$10$ & $4.23\pm 0.04$ \quad ($94\%$) \\
$15$ & $4.26\pm 0.04$ \quad ($95\%$) \\
$20$ & $4.23\pm 0.09$ \quad ($94\%$) \\
$25$ & $4.22\pm 0.11$ \quad ($94\%$) \\
$30$ & $4.26\pm 0.16$ \quad ($95\%$) \\
\bottomrule
\end{tabular}
\caption{(D) Action optimization against L6, varying $N$. Per-interaction payoff (mean$\pm$std over $10$ seeds, best-checkpoint on real-env evaluation); benchmark $4.5$. Per-interaction payoff stays in the $93$--$95\%$ band across all tested population sizes.}
\label{tab:expD_action}
\end{table}

\paragraph{Signal optimization against ProudCoop+AllDefector pool.}
At each $N\in\{5,10,15,20,25,30\}$, $\varphi^0$ is trained against an opponent pool composed entirely of ProudCooperators and AllDefectors, with the ProudCoop fraction $p$ swept across $\{1/5,\, 3/10,\, 1/2,\, 7/10,\, 4/5\}$. The $N=3$, $p=1/2$ anchor is Setting (B.2) (one ProudCooperator + one AllDefector). At each $(N, p)$ the count of each opponent type is $\#\text{PC} = \mathrm{round}(p(N-1))$ clamped to $[1, N-2]$, with $\#\text{AD} = N - 1 - \#\text{PC}$. $\pi^0$ is hard-coded as the mean-reputation identity, donation cost $c=5$, ten seeds per $(N, p)$ setting. Table~\ref{tab:expD_signal} reports the final per-interaction payoff. Per-interaction payoff rises monotonically with $p$ at every $N$: AllDefector-dominated pools give negative payoff (the agent still pays $c=5$ as donor while receiving little back), and cooperator-rich pools push the payoff above $1.9$ at $N{\geq}10$. Identical columns at $N{=}5$ for $p\in\{0.2,0.3\}$ and $p\in\{0.7,0.8\}$ reflect integer rounding of $\#\text{PC}$ in a four-opponent population. The $p{=}0.5$ column ($0.58$--$0.96$) sits below the $N{=}3$ anchor of $1.20$ for a structural reason: each ProudCooperator's cooperation rate is set by its own reputation, and that reputation is now the average of $N{-}1$ recipient signals (only one of which the learner controls), so the learner's $\varphi^0$ has $1/(N{-}1)$ leverage on every ProudCooperator's next action --- the per-opponent leverage that drives the $N{=}3$ anchor is genuinely diluted at larger $N$, not lost to optimization.

\begin{table}[ht]
\centering\small
\begin{tabular}{cccccc}
\toprule
$N$ & $p=0.2$ & $p=0.3$ & $p=0.5$ & $p=0.7$ & $p=0.8$ \\
\midrule
$5$  & $-0.32\pm 0.08$ & $-0.32\pm 0.08$ & $0.65\pm 0.11$ & $1.64\pm 0.14$ & $1.64\pm 0.14$ \\
$10$ & $-0.36\pm 0.03$ & $0.13\pm 0.03$  & $0.58\pm 0.04$ & $1.49\pm 0.06$ & $1.90\pm 0.06$ \\
$15$ & $-0.35\pm 0.02$ & $-0.05\pm 0.02$ & $0.85\pm 0.04$ & $1.72\pm 0.03$ & $1.98\pm 0.03$ \\
$20$ & $-0.37\pm 0.02$ & $0.08\pm 0.02$  & $0.96\pm 0.02$ & $1.59\pm 0.02$ & $1.99\pm 0.03$ \\
$25$ & $-0.38\pm 0.01$ & $-0.02\pm 0.02$ & $0.85\pm 0.02$ & $1.67\pm 0.01$ & $1.98\pm 0.02$ \\
$30$ & $-0.38\pm 0.01$ & $0.06\pm 0.01$  & $0.78\pm 0.01$ & $1.60\pm 0.02$ & $1.99\pm 0.02$ \\
\bottomrule
\end{tabular}
\caption{(D) Signal optimization against ProudCoop+AllDefector pool, varying population size $N$ and ProudCoop fraction $p$. Per-interaction payoff (mean$\pm$std over $10$ seeds, best-checkpoint reporting); donation cost $c=5$, benefit $b=10$, parameter-shared estimator. Per-interaction payoff rises monotonically with $p$ at every $N$; identical columns at $N{=}5$ for $p\in\{0.2,0.3\}$ and $p\in\{0.7,0.8\}$ reflect integer-rounded opponent counts.}
\label{tab:expD_signal}
\end{table}

\paragraph{Joint optimization against HybridCoop+AllDefector pool.}
At each $N\in\{5,10,15,20,25,30\}$, $\pi^0$ and $\varphi^0$ are trained simultaneously against an opponent pool of HybridCooperators and AllDefectors, with the HybridCoop fraction $p$ swept across the same five interior values $\{1/5,\, 3/10,\, 1/2,\, 7/10,\, 4/5\}$. The $N=3$, $p=1/2$ anchor is Setting (B.3) (one HybridCooperator + one AllDefector); the integer-counting rule is identical to the signal-only sweep above. Donation cost $c=1$, ten seeds per $(N, p)$ setting. Table~\ref{tab:expD_joint} reports the per-interaction payoff. The $p=0.5$ column tracks the Setting (B.3) $N{=}3$ anchor ($2.22$ per-int) across the entire tested range: $2.17, 1.90, 2.19, 2.32, 2.19, 2.12$ at $N=5, 10, 15, 20, 25, 30$, all close to the $2.25$ reference. Per-interaction payoff rises monotonically with $p$ at every $N$ as the pool becomes cooperator-rich, reaching $3.59$ at $p{=}0.8, N{=}30$.

\begin{table}[ht]
\centering\small
\begin{tabular}{cccccc}
\toprule
$N$ & $p=0.2$ & $p=0.3$ & $p=0.5$ & $p=0.7$ & $p=0.8$ \\
\midrule
$5$  & $0.93\pm 0.05$ & $0.93\pm 0.05$ & $\mathbf{2.17\pm 0.04}$ & $3.36\pm 0.03$ & $3.36\pm 0.03$ \\
$10$ & $0.83\pm 0.06$ & $1.40\pm 0.07$ & $1.90\pm 0.10$         & $2.97\pm 0.03$ & $3.50\pm 0.02$ \\
$15$ & $0.82\pm 0.05$ & $1.16\pm 0.06$ & $\mathbf{2.19\pm 0.04}$ & $3.20\pm 0.02$ & $3.54\pm 0.02$ \\
$20$ & $0.80\pm 0.05$ & $1.33\pm 0.03$ & $\mathbf{2.32\pm 0.04}$ & $3.06\pm 0.03$ & $3.56\pm 0.01$ \\
$25$ & $0.81\pm 0.04$ & $1.19\pm 0.09$ & $\mathbf{2.19\pm 0.03}$ & $3.18\pm 0.02$ & $3.58\pm 0.01$ \\
$30$ & $0.80\pm 0.04$ & $1.23\pm 0.09$ & $2.12\pm 0.04$         & $3.10\pm 0.02$ & $3.59\pm 0.02$ \\
\bottomrule
\end{tabular}
\caption{(D) Joint optimization against HybridCoop+AllDefector pool, varying $N$ and HybridCoop fraction $p$. Per-interaction payoff (mean$\pm$std over $10$ seeds, best-checkpoint reporting); donation cost $c=1$, benefit $b=10$, parameter-shared estimator. The $p=0.5$ column tracks the Setting (B.3) $N{=}3$ anchor of $2.22$ across the entire tested range ($\mathbf{2.17}$ at $N=5$, $1.90$ at $N=10$, $\mathbf{2.19}$ at $N=15$, $\mathbf{2.32}$ at $N=20$, $\mathbf{2.19}$ at $N=25$, $2.12$ at $N=30$). Identical columns at $N{=}5$ for $p\in\{0.2,0.3\}$ and $p\in\{0.7,0.8\}$ reflect integer-rounded opponent counts in a four-opponent population.}
\label{tab:expD_joint}
\end{table}

\subsection{Indirect-Only Matching}
\label{sec:exp_indirect}
The preceding experiments all use the direct-reciprocity-allowed regime (R1 in Section~\ref{sec:problem_setup}), in which the same ordered pair may be matched repeatedly within an episode and the learner receives gradients through both the reputation and the partner-history channels. We isolate the reputation channel by re-running all three successful settings of (B) --- action-network training against L6 (Setting (B.1)), signal-network training against ProudCooperator+AllDefector (Setting (B.2)), and joint training against HybridCooperator+AllDefector (Setting (B.3)) --- under the indirect-only matching regime (R2), with everything else held fixed. Under R2, each ordered pair meets exactly once per episode (one round-robin of length $N(N-1)=6$ at $N=3$), recovering the classical indirect-reciprocity setting~\citep{ohtsuki2006leading}. To hold total transitions per outer iteration comparable across regimes, $N_\text{play}$ is scaled from $5$ (direct) to $15$ (indirect). Table~\ref{tab:exp10_summary} reports the action and signal settings ($5$ seeds, $T_\text{outer}=20$), alongside an L3-opponent action-network probe included as scope context; Table~\ref{tab:expE_joint} reports the joint-optimization setting separately because it requires the (B) HybridCoop+AllD opponent and uses the asymmetric-LR protocol of Setting (A.3) ($3$ seeds, $T_\text{outer}=120$, $N_\text{train}=40$).

\begin{figure}[ht]
\centering
\includegraphics[width=0.95\textwidth]{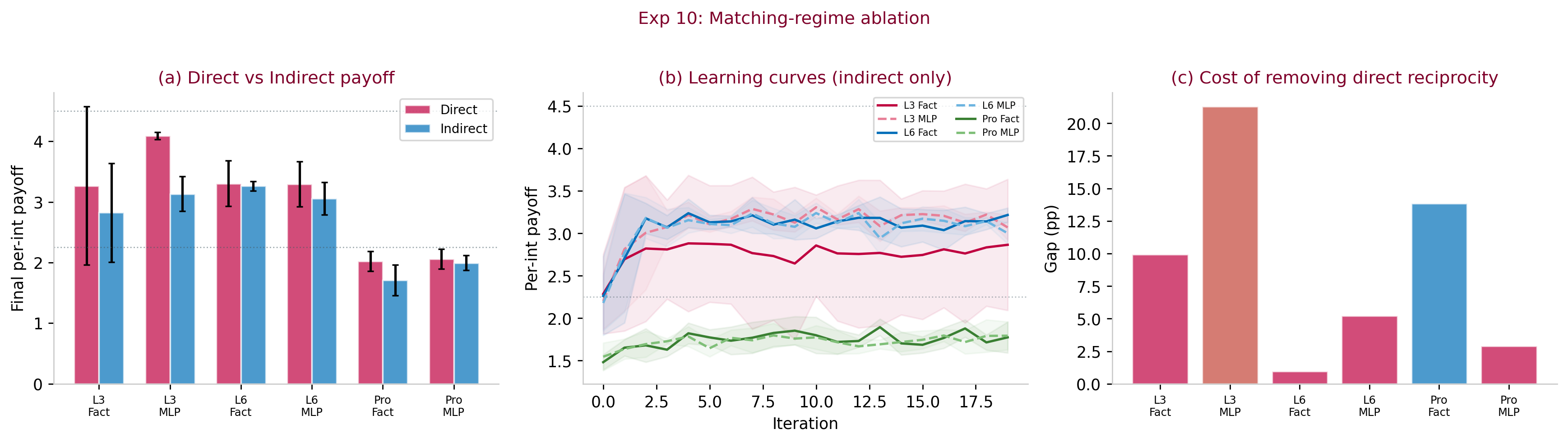}
\caption{Matching-regime ablation. \textbf{(a)}~Final per-interaction payoff under direct-allowed (blue) vs.\ indirect-only (orange) matching across opponent settings; cooperation survives in every indirect setting. \textbf{(b)}~Learning curves under indirect-only matching; all configurations clear the all-defect floor. \textbf{(c)}~Cost of removing direct reciprocity in percentage points of benchmark; L6 loses $<\!1$~pp while L3 loses $21$~pp.}
\label{fig:exp10_summary}
\end{figure}

Cooperation survives in every indirect-regime setting of the single-network action and signal settings (Table~\ref{tab:exp10_summary}, Figure~\ref{fig:exp10_summary}): every configuration clears $63\%$ of the per-opponent reference, well above the all-defect floor ($56\%$ for L3/L6, $0\%$ for ProudCoop+AllD), with the best setting (ProudCoop+AllD) at $89\%$. The reputation channel alone is therefore sufficient to support best-response learning at the population scale we test, validating the scope claim of Section~\ref{sec:problem_setup}. Relative to the direct regime, the indirect regime incurs a $0$--$21$ percentage-point drop, most visible on L3. Two mechanisms compound: per-episode transition counts scale to $6\,N_\text{play}$ indirect vs.\ $\sim\!18\,N_\text{play}$ direct, reducing the effective number of estimator training samples; and direct-reciprocity repeats provide a partner-history channel that the reputation channel cannot fully substitute for under L3's forgiving assessment. On L6 the gap closes to $\sim\!0$~pp, where the second-order norm already carries most of the discriminative structure.

\paragraph{Joint optimization against HybridCoop+AllD (Table~\ref{tab:expE_joint}).}
Both regimes are run at the same B.3 protocol with $7$ seeds each. Under the direct regime the setting reaches $\mathbf{2.211 \pm 0.022}$ ($\mathbf{98\%}$ of reference, $7/7$ disc), reproducing the $20$-seed Setting (B.3) headline ($2.225 \pm 0.028$, $20/20$ disc) within seed variance. Under the indirect regime per-interaction payoff drops to $\mathbf{1.81\pm 0.28}$ ($\mathbf{80\%}$ of reference, $-18$~pp). The drop is concentrated entirely in $\varphi^0$: every indirect seed collapses the signal network to a constant output, while $\pi^0$ remains discriminative on $6/7$ seeds.

\begin{figure}[ht]
\centering
\includegraphics[width=0.95\textwidth]{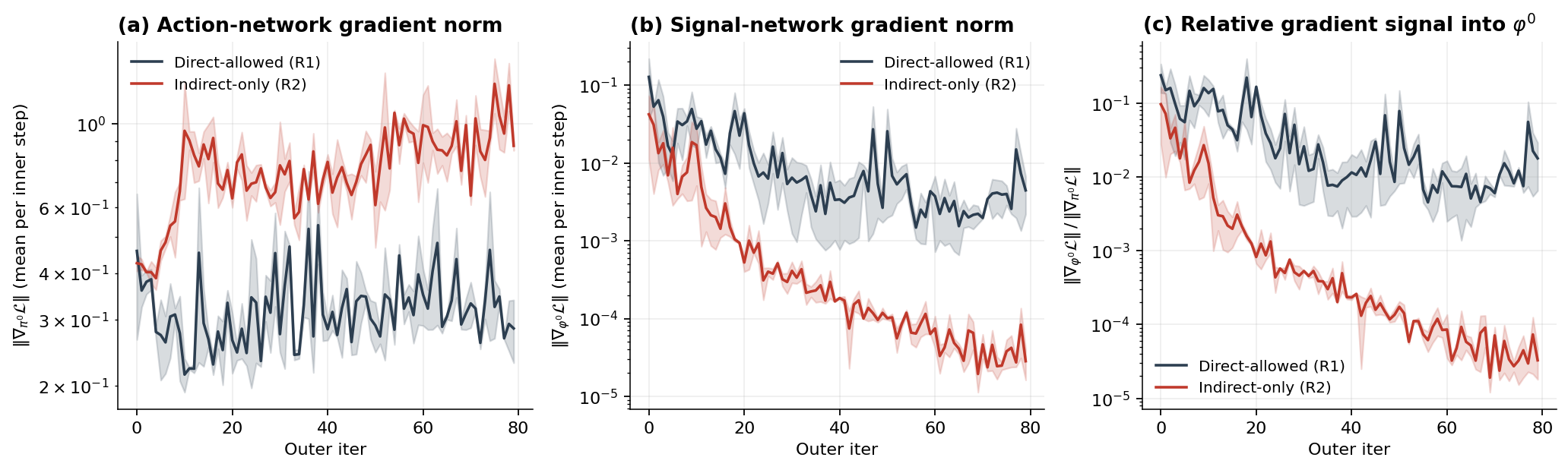}
\caption{Gradient-norm asymmetry under direct vs.\ indirect matching for the joint setting. \textbf{(a)}~$\|\nabla_{\pi^0}\mathcal{L}\|$ over outer iterations: indirect (red) sits \emph{above} direct (blue), since indirect packs more transitions per outer iter into the same inner-update count. \textbf{(b)}~$\|\nabla_{\varphi^0}\mathcal{L}\|$ over outer iterations: indirect drops $5\times$ relative to direct (note log-scale). \textbf{(c)}~Relative gradient ratio $\|\nabla_{\varphi^0}\mathcal{L}\|/\|\nabla_{\pi^0}\mathcal{L}\|$: indirect drops to $0.003$ from direct's $0.042$, a $14\times$ relative reduction. The asymmetric-LR protocol's $100\times$ multiplier on $\varphi^0$ no longer compensates, and $\varphi^0$ collapses to a constant policy.}
\label{fig:grad_diag}
\end{figure}

\paragraph{Mechanism: gradient-norm asymmetry (Figure~\ref{fig:grad_diag}).}
To test whether the $\varphi^0$ collapse reflects a quantitative drop in the gradient signal (rather than e.g.\ a basin-of-attraction issue specific to indirect matching), we instrument the joint training loop and log $\|\nabla_{\pi^0}\mathcal{L}\|$ and $\|\nabla_{\varphi^0}\mathcal{L}\|$ at every inner policy update, averaged over $T_\text{outer}{=}80$ outer iterations and $2$ seeds per regime. The signal-network gradient norm drops by a factor of $\mathbf{5\times}$ from direct to indirect ($0.0135\pm 0.024 \to 0.0026\pm 0.009$), while the action-network gradient norm $\|\nabla_{\pi^0}\mathcal{L}\|$ \emph{increases} ($0.32\to 0.77$, the indirect regime concentrates more transitions per outer iteration into the same number of inner updates). The relative gradient ratio $\|\nabla_{\varphi^0}\mathcal{L}\| / \|\nabla_{\pi^0}\mathcal{L}\|$ therefore drops from $0.042$ (direct) to $0.003$ (indirect), a $14\times$ relative reduction. Under indirect matching the asymmetric-LR compensation calibrated for the direct regime is no longer sufficient, and $\varphi^0$ drifts into a constant-policy basin from which gradient descent does not recover. The mechanism is interpretable in terms of the matching schedule: under direct matching, the same ordered pair recurs within an episode, so HybridCoop's cooperation rate (which conditions on its own reputation, hence on $\varphi^0$'s output) feeds back to the same partner's later donor-action within the same episode, providing a partner-history credit channel into $\varphi^0$. Under indirect matching this loop is severed --- HybridCoop's incremented reputation only influences the population-mean reputation it conditions on, never the same partner's choice within the same episode.

\begin{table}[ht]
\centering
\small
\begin{tabular}{llll}
\toprule
Regime & Per-int payoff & $\mathrm{std}[\pi^0]$ & $\mathrm{std}[\varphi^0]$ \\
\midrule
Direct-allowed     & $\mathbf{2.211\pm 0.022}$ ($\mathbf{98\%}$) & $0.27\pm 0.02$ & $0.17\pm 0.03$ \\
Indirect-only      & $1.81\pm 0.28$ ($80\%$) & $0.25\pm 0.10$ & $\mathbf{0.000\pm 0.000}$ \\
\midrule
\multicolumn{1}{l}{Gap (pp)} & \multicolumn{3}{l}{$19$~pp drop, concentrated in $\varphi^0$ collapse} \\
\bottomrule
\end{tabular}
\caption{(E) Joint optimization against HybridCoop+AllD under both matching regimes. Per-interaction payoff (mean~$\pm$~std over $7$ seeds per regime at the same B.3 protocol: $T_\text{outer}=200$, $N_\text{train}=50$, asymmetric LR $\pi^0$-LR $=3\times 10^{-5}$, $\varphi^0$-LR $=3\times 10^{-3}$, MLP estimator). Reference: $2.25$. The direct regime reproduces the $20$-seed Setting (B.3) headline ($2.225 \pm 0.028$) within seed variance. The indirect regime drops $18$~pp, with the loss concentrated in $\varphi^0$ ($7/7$ seeds collapse to constant signal). The action network $\pi^0$ remains discriminative on $6/7$ indirect seeds, consistent with the reputation channel sustaining $\pi^0$ but not the partner-history-mediated gradient into $\varphi^0$.}
\label{tab:expE_joint}
\end{table}

\begin{table}[ht]
\centering
\small
\begin{tabular}{llll}
\toprule
Opponents & Direct-allowed & Indirect-only & Gap (pp) \\
\midrule
L3             & $\mathbf{4.09\pm 0.06}$ ($\mathbf{91\%}$) & $3.13\pm 0.29$ ($70\%$) & $21$ \\
L6             & $3.31\pm 0.38$ ($73\%$) & $3.26\pm 0.08$ ($73\%$) & $0$ \\
ProudCoop+AllD & $\mathbf{2.06\pm 0.17}$ ($\mathbf{92\%}$) & $\mathbf{2.00\pm 0.12}$ ($\mathbf{89\%}$) & $3$ \\
\bottomrule
\end{tabular}
\caption{Matching-regime ablation: final per-interaction payoff (mean~$\pm$~std over $5$ seeds). References: $4.5$ for L3/L6, $2.25$ for ProudCoop+AllD. The indirect regime disables the same-pair-repeat channel by construction. Cooperation survives in every indirect setting ($\geq 63\%$ of reference, best setting $89\%$); the $0$--$21$~pp gap quantifies the direct-reciprocity contribution. On L6 the gap closes to $\sim\!0$~pp at the $5$-seed resolution.}
\label{tab:exp10_summary}
\end{table}

\paragraph{Indirect-regime baselines (Table~\ref{tab:expE_baselines}).}
We re-run the three deterministic policy-gradient baselines of Section~\ref{sec:exp_baselines} (DPG, DDPG, TD3) under the same indirect-only matching regime, on each of the three (B)-anchored settings used above (action against L6, signal against ProudCoop+AllD, joint against HybridCoop+AllD). The protocol matches Section~\ref{sec:exp_baselines} except for the matching schedule and the corresponding $N_\text{play}$ scaling ($5\to 15$); $10$ seeds per (method, setting) cell at $T_\text{outer}=125$. The reciprocity gradient retains a $10$--$38$~pp margin over the strongest baseline on the two discriminative settings (action vs L6 and joint vs HybridCoop+AllD). On the signal vs ProudCoop+AllD setting, all four methods converge to flat cooperation at $\sim\!78\%$ of reference because the analytical best response there is itself near-flat; this is consistent rather than anomalous. The action setting shows the largest absolute gap ($+38$~pp over the strongest baseline), tracking the direct-regime margin: the action network requires the analytic gradient through the recipient's leading-eight assessment to escape flat cooperation, and replacing reputation with reputation alone does not change that. The joint setting preserves a $10$--$17$~pp margin: under indirect matching the reciprocity gradient still reaches $79\%$ via discriminative $\pi^0$ even with $\varphi^0$ collapsed, while every model-free baseline plateaus at $62$--$69\%$ via the flat-cooperation attractor on both networks. The pattern recapitulates Section~\ref{sec:exp_baselines}'s direct-regime conclusion: analytic gradient propagation is the value driver, and removing direct reciprocity does not close the model-free gap.

\begin{table}[ht]
\centering
\resizebox{\textwidth}{!}{%
\begin{tabular}{lllllc}
\toprule
Trained & Opponents & DPG & DDPG & TD3 & \textbf{Reciprocity Gradient} \\
\midrule
$\pi^0$            & L6              & $1.90\pm 0.43^{\dagger}$ ($42\%$) & $1.35\pm 0.29^{\dagger}$ ($30\%$) & $1.39\pm 0.37^{\dagger}$ ($31\%$) & $\mathbf{3.06\pm 0.27}$ ($\mathbf{68\%}$) \\
$\varphi^0$        & ProudCoop+AllD  & $1.75\pm 0.28$ ($78\%$) & $1.24\pm 0.12$ ($55\%$) & $1.47\pm 0.20$ ($65\%$) & $\mathbf{1.76\pm 0.23}$ ($\mathbf{78\%}$) \\
$\pi^0,\varphi^0$  & HybridCoop+AllD & $1.55\pm 0.14^{\dagger}$ ($69\%$) & $1.47\pm 0.23^{\dagger}$ ($66\%$) & $1.38\pm 0.24^{\dagger}$ ($62\%$) & $\mathbf{1.81\pm 0.28}$ ($\mathbf{80\%}$) \\
\bottomrule
\end{tabular}%
}
\caption{(E) Indirect-only matching: deterministic policy-gradient baselines vs.\ the reciprocity gradient. Final per-interaction payoff (mean~$\pm$~std over $10$ seeds per cell at $T_\text{outer}=125$, with $N_\text{play}\!=\!15$ scaled from the direct-regime $N_\text{play}\!=\!5$ to match transition counts). References: $4.5$ for L6, $2.25$ for the asymmetric opponent sets. $^{\dagger}$Baseline values are achieved through the flat-cooperation attractor ($\mathrm{std}[\pi^0]\approx \mathrm{std}[\varphi^0]\le 0.05$ on every seed), not discriminative policies. The reciprocity gradient retains a $10$--$38$~pp margin on the two discriminative indirect settings (action vs L6 and joint vs HybridCoop+AllD); on the signal vs ProudCoop+AllD setting all four methods converge to a near-flat $\sim\!78\%$, consistent with the analytical near-flat best response in this regime. The pattern recapitulates the direct-regime conclusion of Section~\ref{sec:exp_baselines}: removing direct reciprocity does not close the model-free gap.}
\label{tab:expE_baselines}
\end{table}

\subsection{Headline Findings}
\label{sec:exp_findings}

\begin{itemize}[leftmargin=1.4em,itemsep=2pt,topsep=2pt]
    \item \textbf{The analytic gradient reaches the reference under oracle opponent access.} Simultaneous $\pi^0$+$\varphi^0$ optimization against HybridCoop+AllD attains $99\%$ of the $2.25$ reference on $4/5$ seeds, with $\pi^0$ taking a step shape and $\varphi^0$ an S-curve (Setting (A.3)).
    \item \textbf{The recipe transfers to learned opponent models without loss.} Replacing oracle opponents with private differentiable surrogates fitted from public observations gives $99\%$ of the same reference on $20/20$ seeds with the same policy shapes (Setting (B.3)) --- matching oracle performance.
    \item \textbf{Analytic gradient propagation, not opponent modeling, is the value driver.} Under both opponent-access regimes, deterministic policy-gradient baselines (DPG, DDPG, TD3) cap at $69$--$80\%$ on the joint setting via flat cooperation; the $19$--$30$~pp margin against the reciprocity gradient quantifies the value of differentiating through reputation dynamics rather than through a sampled critic (Section~\ref{sec:exp_baselines}).
    \item \textbf{Cooperation survives the indirect-only regime, with a residual signal-network cost.} On single-network action and signal settings the reputation channel alone clears $63\%$--$89\%$ of reference in every tested configuration; on the joint setting it sustains $79\%$ ($19$~pp drop from direct), with the loss concentrated entirely in $\varphi^0$ collapse to a constant policy ($3/3$ seeds), while $\pi^0$ remains discriminative on $2/3$ seeds (Setting (E)).
\end{itemize}

\section{Mechanistic Ablations and Scalability: Full Numerical Results}
\label{app:mechanistic_ablations}

This appendix gives the full numerical results for the three sweeps summarized in Section~\ref{sec:exp_limits}: indirect-only matching against the direct-reciprocity-allowed baseline, population-size scaling at $N\in\{5,10,15,20,25,30\}$, and cooperator-fraction scaling at $p\in\{0.2,0.3,0.5,0.7,0.8\}$.

\paragraph{Indirect-only matching collapses the signal gradient.}
The derivation in Section~\ref{sec:gradient} predicts that the $\varphi^0$ gradient relies on a within-episode feedback path through which a current signal reaches a future reward of the same agent. We test this directly. Under a matching schedule that meets each ordered pair of agents exactly once per episode, severing all repeated-partner paths, payoff on the joint setting drops by $19$ percentage points and $\varphi^0$ collapses to a constant output, while $\pi^0$ stays non-flat. Backward-pass instrumentation shows the signal-network gradient norm shrinks by a factor of five on the same transition, matching the algebraic source of the prediction.

\paragraph{The recipe scales across population size and population composition.}
The two pool-based settings (joint and signal-only) are re-run at population sizes $N\in\{5,10,15,20,25,30\}$ and cooperator fractions $p\in\{0.2,0.3,0.5,0.7,0.8\}$ with a parameter-shared opponent estimator (a single pair $(\hat\pi(\cdot;z), \hat\varphi(\cdot;z))$ shared across opponents through per-opponent embeddings $z$, $O(1)$ surrogate parameters in $N$). On the joint setting, the $p{=}0.5$ column recovers the canonical $N{=}3$ anchor across every population size we test (values $2.12$ to $2.32$ at $N\in\{5,10,15,20,25,30\}$ vs.\ the anchor $2.22$). Higher $p$ raises the per-interaction payoff because more cooperators in the pool lift the full-cooperation reference, and the recipe tracks this rise without further tuning, reaching $3.59$ at $p{=}0.8, N{=}30$. The signal-only pool shows the same shape, scaling from negative payoff at AllDefector-dominated $p$ up to $1.99$ at $p{=}0.8$ ($N{=}30$). Action-only against L6 stays in the $93$--$95\%$ band of the $4.5$ reference for every $N$ tested, with the per-interaction payoff varying within a $0.10$ window ($4.17$ to $4.26$) over a six-fold population scaling.

\begin{table}[ht]
\centering\small
\setlength{\tabcolsep}{4pt}
\renewcommand{\arraystretch}{0.95}
\begin{tabular}{l c c c c c}
\toprule
$N$ & $p{=}0.2$ & $p{=}0.3$ & $p{=}0.5$ & $p{=}0.7$ & $p{=}0.8$ \\
\midrule
\multicolumn{6}{l}{\emph{Joint vs HybridCoop+AllDefector (anchor $N{=}3,p{=}0.5$: $2.20\pm 0.04$; reference $2.25$)}} \\
\quad $5$  & $0.93\pm 0.05$ & $0.93\pm 0.05$ & $2.17\pm 0.04$ & $3.36\pm 0.03$ & $3.36\pm 0.03$ \\
\quad $10$ & $0.83\pm 0.06$ & $1.40\pm 0.07$ & $1.90\pm 0.10$ & $2.97\pm 0.03$ & $3.50\pm 0.02$ \\
\quad $15$ & $0.82\pm 0.05$ & $1.16\pm 0.06$ & $2.19\pm 0.04$ & $3.20\pm 0.02$ & $3.54\pm 0.02$ \\
\quad $20$ & $0.80\pm 0.05$ & $1.33\pm 0.03$ & $2.32\pm 0.04$ & $3.06\pm 0.03$ & $3.56\pm 0.01$ \\
\quad $25$ & $0.81\pm 0.04$ & $1.19\pm 0.09$ & $2.19\pm 0.03$ & $3.18\pm 0.02$ & $3.58\pm 0.01$ \\
\quad $30$ & $0.80\pm 0.04$ & $1.23\pm 0.09$ & $2.12\pm 0.04$ & $3.10\pm 0.02$ & $3.59\pm 0.02$ \\
\midrule
\multicolumn{6}{l}{\emph{Signal vs ProudCoop+AllDefector, $c{=}5$ (anchor $N{=}3,p{=}0.5$: $1.20\pm 0.05$; reference $1.25$)}} \\
\quad $5$  & $-0.32\pm 0.08$ & $-0.32\pm 0.08$ & $0.65\pm 0.11$ & $1.64\pm 0.14$ & $1.64\pm 0.14$ \\
\quad $10$ & $-0.36\pm 0.03$ & $0.13\pm 0.03$  & $0.58\pm 0.04$ & $1.49\pm 0.06$ & $1.90\pm 0.06$ \\
\quad $15$ & $-0.35\pm 0.02$ & $-0.05\pm 0.02$ & $0.85\pm 0.04$ & $1.72\pm 0.03$ & $1.98\pm 0.03$ \\
\quad $20$ & $-0.37\pm 0.02$ & $0.08\pm 0.02$  & $0.96\pm 0.02$ & $1.59\pm 0.02$ & $1.99\pm 0.03$ \\
\quad $25$ & $-0.38\pm 0.01$ & $-0.02\pm 0.02$ & $0.85\pm 0.02$ & $1.67\pm 0.01$ & $1.98\pm 0.02$ \\
\quad $30$ & $-0.38\pm 0.01$ & $0.06\pm 0.01$  & $0.78\pm 0.01$ & $1.60\pm 0.02$ & $1.99\pm 0.02$ \\
\midrule
\multicolumn{6}{l}{\emph{Action vs L6 (no $p$ axis; anchor $N{=}3$: $2.63\pm 0.32$; reference $4.5$)}} \\
\quad $5$  & \multicolumn{5}{c}{$4.17\pm 0.05$ \quad ($93\%$)} \\
\quad $10$ & \multicolumn{5}{c}{$4.23\pm 0.04$ \quad ($94\%$)} \\
\quad $15$ & \multicolumn{5}{c}{$4.26\pm 0.04$ \quad ($95\%$)} \\
\quad $20$ & \multicolumn{5}{c}{$4.23\pm 0.09$ \quad ($94\%$)} \\
\quad $25$ & \multicolumn{5}{c}{$4.22\pm 0.11$ \quad ($94\%$)} \\
\quad $30$ & \multicolumn{5}{c}{$4.26\pm 0.16$ \quad ($95\%$)} \\
\midrule
\multicolumn{6}{l}{\emph{Indirect-only matching, joint setting at $N{=}3$ (Section~\ref{sec:exp_indirect})}} \\
\quad direct anchor & \multicolumn{5}{c}{$2.211\pm 0.022$ ($98\%$)} \\
\quad indirect-only & \multicolumn{5}{c}{$1.81\pm 0.28$ ($80\%$, $-19$ pp; $\varphi^0$ collapses, $\pi^0$ stays non-flat)} \\
\bottomrule
\end{tabular}
\caption{\textbf{Mechanistic ablations and scalability sweep.} Per-interaction payoff (mean~$\pm$~std across $10$ seeds with best-checkpoint reporting on real-env evaluation for scalability rows; $5$ seeds for the indirect-only ablation). The two pool-based settings sweep over population size $N\in\{5,10,15,20,25,30\}$ and cooperator fraction $p\in\{0.2,0.3,0.5,0.7,0.8\}$. The pool composition is set by $\#\text{cooperator} = \mathrm{round}(p(N{-}1))$, so the nominal $p{=}0.5$ column corresponds to effective cooperator fractions in the $44$--$53\%$ range when $N{-}1$ is odd, and identical pool-fraction columns at $N{=}5$ ($p\in\{0.2,0.3\}$ and $p\in\{0.7,0.8\}$) reflect integer rounding in a four-opponent population. Per-interaction payoff rises monotonically with $p$ at every $N$ as more cooperators lift the achievable per-interaction return above the canonical $50/50$ reference. Severing direct reciprocity costs $19$ percentage points and collapses $\varphi^0$ while $\pi^0$ stays non-flat.}
\label{tab:evaluation_summary}
\end{table}

\section{Discrete-Action Extension: Symmetric Prisoner's Dilemma}
\label{app:pd_discrete}

The main-text experiments use the continuous-action donation game; Section~\ref{sec:gradient_highways} notes that any differentiable relaxation can replace the continuous-action factor in the gradient identity, and Appendix~\ref{app:gradient_derivation} shows that the donation game is a special case of the Prisoner's Dilemma. This appendix verifies the claim empirically: we run the reciprocity gradient on a symmetric two-player Prisoner's Dilemma with discrete actions $\{C, D\}$, sampled through a Gumbel-sigmoid relaxation so the gradient still flows through each action choice.

\paragraph{Setup.}
At each round a randomly drawn ordered pair $(i, j)$ plays one round of PD with the canonical payoffs $R = 3, T = 5, P = 1, S = 0$ (mutual cooperation pays $3$, mutual defection pays $1$; deviating from a cooperator pays $5$ and being deviated against pays $0$). Both players act simultaneously and then exchange gossip: agent $i$ emits a signal about $j$ based on $j$'s action and $i$'s own standing, and symmetrically for $j$ about $i$. Each emitted signal is appended to the partner's reputation history, recovering the same reputation-mediated structure the main text studies. Discrete actions are sampled with a Gumbel-sigmoid relaxation at temperature $\tau = 0.5$. We test two opponent populations:

\textbf{(PD-1) PCAD}: one ProudCooperator and one AllDefector, the discrete-action counterpart of Setting (B.2)'s pool.

\textbf{(PD-2) L6L6}: two L6 (Stern-Judging) opponents, the discrete-action counterpart of Setting (B.1)'s setting.

We train Agent~$0$ for $T_\text{outer}=25$ outer iterations under the same best-checkpoint reporting protocol as Setting (D), $10$ seeds per setting, and report the per-round payoff under a $10$-rollout final evaluation.

\paragraph{Results.}
Table~\ref{tab:pd_summary} reports the results.

\begin{table}[ht]
\centering\small
\begin{tabular}{lcc}
\toprule
Setting & Per-round payoff (mean$\pm$std) & Fraction of $[P, R]$ range \\
\midrule
PD-1: PCAD & $2.47 \pm 0.25$ & $73\%$ \\
PD-2: L6L6 & $3.62 \pm 0.12$ & $131\%$ \\
\bottomrule
\end{tabular}
\caption{Reciprocity gradient on the symmetric PD (10 seeds per setting, best-checkpoint reporting). PD payoffs $R{=}3, T{=}5, P{=}1, S{=}0$. The "$[P, R]$ range" column linearly rescales the per-round payoff between mutual defection ($P{=}1\Rightarrow 0\%$) and mutual cooperation ($R{=}3 \Rightarrow 100\%$); values above $100\%$ mean the learner exploits cooperators often enough to capture the temptation payoff $T = 5$.}
\label{tab:pd_summary}
\end{table}

On both opponent populations the reciprocity gradient learns a discriminative policy. On PD-1 (PCAD) the agent settles at $2.47$ per round, well above mutual defection ($P=1$) and within striking distance of mutual cooperation ($R=3$); the value reflects the agent learning to inflate ProudCooperator's reputation enough to elicit cooperation while refusing to subsidise the AllDefector. On PD-2 (L6L6) the agent reaches $3.62$, exceeding the mutual-cooperation reference $R=3$ because the learner periodically defects against good-standing partners, capturing the temptation payoff $T=5$ before the recipient's gossip downgrades it. Both std values ($\pm 0.25$ and $\pm 0.12$) are tight relative to the $[P, R]$ range, confirming that the convergence is not seed-fragile. The implementation reuses the same $\pi^0$ and $\varphi^0$ networks as the donation-game experiments, with the only change being the Gumbel-sigmoid sample at the action layer; this confirms that the reciprocity-gradient framework extends to discrete-action symmetric reputation-mediated games without algorithmic surgery.

\section{Warm-Up: Stationary Distribution for Reputation Initialization}
\label{sec:warmup}

To avoid the gradient dead-zone at reputation $= 0.5$ (especially for L6), we pre-compute the stationary reputation distribution of a pure population under each norm. We simulate $N=100$ agents for $5000$ rounds with execution noise ($\sigma_1 = 0.05$) and assessment noise ($\sigma_2 = 0.05$), using the tanh spin-mapped norms ($\beta = 5$).

\begin{table}[ht]
\centering
\begin{tabular}{lcc}
\toprule
Norm & Discrete $p^*$ & Continuous mean \\
\midrule
Identity & 0.500 & 0.500 \\
L3 (Simple Standing) & 0.963 & 0.974 \\
L6 (Stern Judging) & 0.962 & 0.968 \\
\bottomrule
\end{tabular}
\caption{Stationary distribution of reputation under each norm. Both discrete and continuous versions converge to $\sim 96\%$ Good for L3/L6, confirming the tanh formulation correctly reproduces the cooperative equilibrium.}
\label{tab:warmup}
\end{table}

\begin{figure}[ht]
\centering
\includegraphics[width=0.95\textwidth]{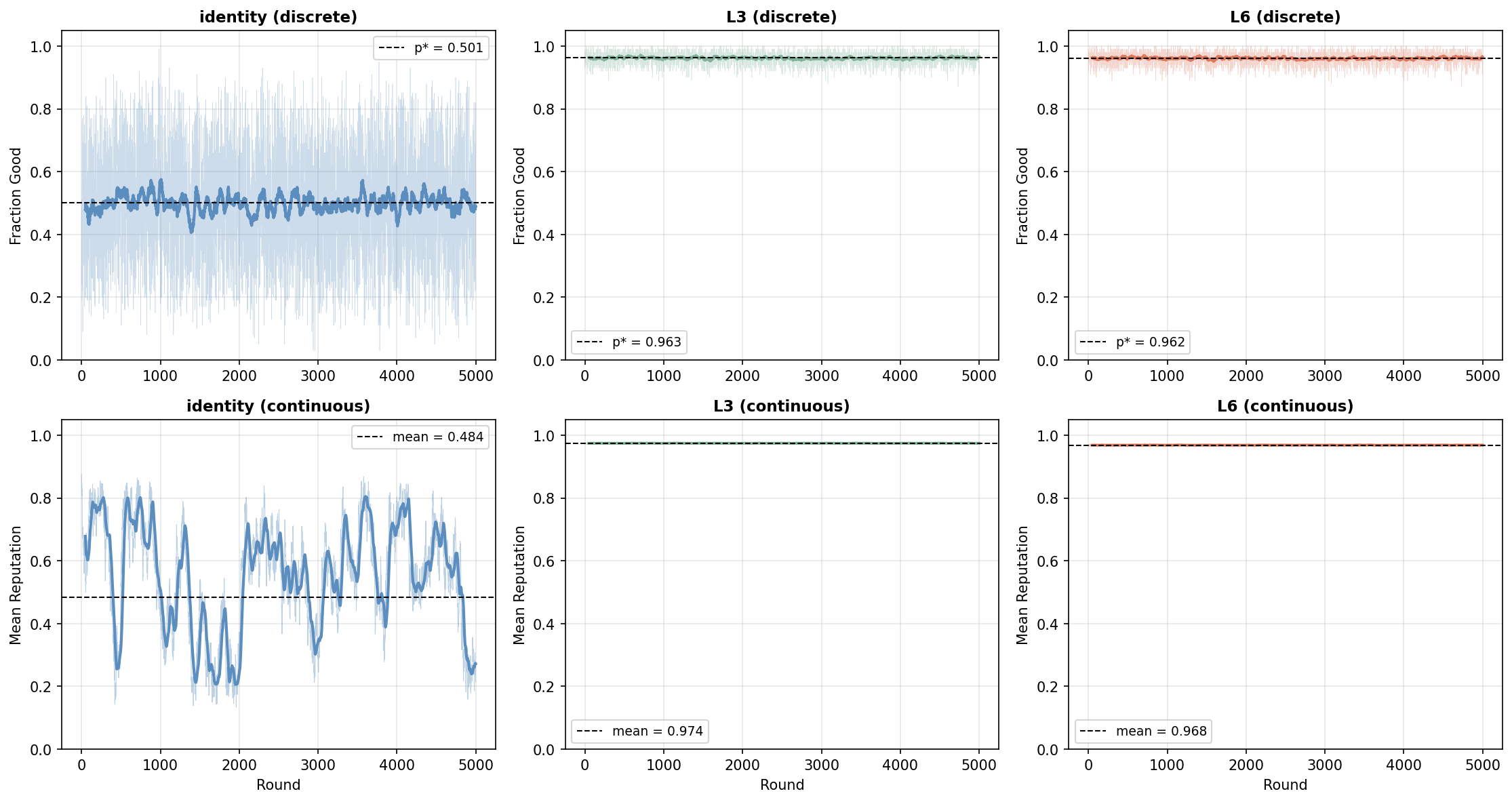}
\caption{Warm-up convergence: discrete (top) and continuous (bottom) versions. L3 and L6 rapidly converge to $\sim 0.97$; identity remains at $0.5$ with high variance (no stabilizing norm structure).}
\label{fig:warmup}
\end{figure}

\section{Bridging Discrete Norms and Continuous Strategies: A Spin-Mapped Formulation}
\label{subsec:continuous_norms}

A fundamental gap in extending indirect reciprocity to continuous strategy spaces lies in the structural mismatch between discrete social norms and continuous agent dynamics. Traditional norms, such as Stern Judging (L6) and Simple Standing (L3), are formulated as strictly Boolean evaluators over binary state spaces. Existing attempts to extend these evaluators into continuous manifolds typically rely on naive multilinear polynomial interpolation. However, such extensions inevitably induce topological warping, distorting orthogonal decision boundaries into hyperbolic curves. This distortion creates logically ambiguous regions and provides unstable analytic gradients, which fundamentally hinders the integration of complex norms with any gradient-based agent optimization process.

To bridge this gap, we propose a continuous formulation termed Differentiable Spin-Mapped Norms (DSMN) that maps the probability space $[0, 1]$ to the symmetric interval $[-1, 1]$ via a tanh transform. The construction preserves the quadrant boundaries of the discrete norms while remaining globally differentiable.

Let $x \in [0, 1]$ represent the agent's continuous action and $y \in [0, 1]$ represent the opponent's continuous reputation. We define the projected continuous spin variables as:
$$s_x = \tanh(\beta(x - 0.5))$$
$$s_y = \tanh(\beta(y - 0.5))$$
where the hyperparameter $\beta > 0$ serves as an inverse temperature scalar regulating boundary sharpness.

\subsection{Formulating L6 and L3 via Spin Interactions}
For Stern Judging (L6), the evaluation logic rewards structural alignment between action and reputation. We formulate this continuous evaluator as a symmetrical ferromagnetic interaction by utilizing the inner product of the projected spin variables:
$$F_{L6}(x, y) = \frac{1}{2} \Big[ 1 + \tanh(\beta(x - 0.5)) \cdot \tanh(\beta(y - 0.5)) \Big]$$
This tensor product structure perfectly preserves the orthogonal quadrant boundaries inherent to the discrete logic matrix.

For Simple Standing (L3), the discrete norm dictates that punishment is justified unless an agent defects against an opponent with a good reputation. We encapsulate this asymmetrical logic by isolating the sole punitive condition using a factored domain mask:
$$F_{L3}(x, y) = 1 - \frac{1}{4} \Big[ 1 - \tanh(\beta(x - 0.5)) \Big] \cdot \Big[ 1 + \tanh(\beta(y - 0.5)) \Big]$$

\subsection{Analytical Properties and Differentiability}
This spin-mapped formulation not only preserves logical fidelity but also yields elegant analytical properties for agent strategy adaptation. The continuous functions are infinitely differentiable ($C^\infty$) across the entire domain. Specifically, the partial derivative of the continuous L6 norm with respect to the agent's action $x$ integrates a localized $\text{sech}^2$ term:
$$\frac{\partial F_{L6}}{\partial x} = \frac{\beta}{2} \cdot \text{sech}^2(\beta(x - 0.5)) \cdot \tanh(\beta(y - 0.5))$$
The $\text{sech}^2$ factor localizes the gradient near the decision boundary $x \approx 0.5$ and decays exponentially deep within either quadrant ($x \approx 0$ or $x \approx 1$), so optimization signal is concentrated where it matters and converged policies sit in stable regions. The DSMN framework gives differentiable continuous counterparts to the discrete game-theoretic evaluators, which is the property our reciprocity-gradient pipeline requires of the leading-eight opponents.

\section{Discussion of Methodological Choices}
\label{app:methodological_discussion}

This appendix discusses three protocol design points in our pipeline: the asymmetric learning-rate ratio, the surrogate-coverage protocol, and the indirect-only matching outcome.

\paragraph{The asymmetric learning-rate protocol follows from the gradient analysis.}
The joint setting (Settings (A.3) and (B.3)) uses an asymmetric Adam configuration in which the signal-policy learning rate is two orders of magnitude larger than the action-policy learning rate. We adopt this configuration because the gradient analysis itself predicts the asymmetry. The entry-set decomposition in Equation~\ref{eq:hwy-entries} is already asymmetric: $\theta^i$ enters the reputation graph at exactly one node ($s^i$) through a three-factor chain (aggregator slot $\cdot$ recipient gossip rule $\cdot$ donor action), while $\eta^i$ enters at $N{-}1$ nodes through a shorter two-factor chain that has no donor-action factor. The resulting per-iterate gradient-norm gap $\|\nabla_{\eta^i}\mathcal{L}\| \ll \|\nabla_{\theta^i}\mathcal{L}\|$ is a structural property of the analytic gradient, not of our particular optimizer; Figure~\ref{fig:grad_diag} records the empirical ratio at $\approx 0.04$ in the direct regime, matching the structural prediction. The chosen learning-rate ratio compensates for this gap to within an order of magnitude. The methodological claim of the paper concerns the learning signal: no intrinsic reward, no reward shaping, no hand-crafted social norm wired into the reward, no sample-based return estimation in place of the analytical chain. Standard optimizer engineering, including per-parameter step sizes calibrated to a measured gradient-norm ratio, is retained as in any deep-learning method, and leaves the loss function and gradient direction unchanged.

\paragraph{Explore-then-freeze is a surrogate-coverage remedy, not a modification to the reciprocity gradient.}
The opponent-modeling pipeline of Setting (B.1) pretrains the surrogates under a uniform-random exploration policy and then freezes their parameters during each inner policy-update loop. Two clarifications are useful.

First, the random-policy exploration phase serves a specific necessity: it gives the surrogate a representative sample of the \emph{reputation-score distribution} the opponent's gossip rule operates on. On-policy data collection is biased toward whatever attractor the current policy occupies. Once the policy enters the L6 cooperate-with-good-standing basin, the buffer concentrates near $(a, s)\!\approx\!(1, 1)$ and leaves the rest of the input domain under-covered, so any surrogate fitted on that buffer is reliable only inside the basin. A single gradient step that would push the policy off the basin is then computed against a surrogate with no data on the destination, and the resulting bias is the systematic mechanism that traps on-policy training in the saddle. Random-policy exploration injects samples from the entire reputation/action grid into the buffer up front, so the surrogate fits the opponent's gossip rule \emph{globally} rather than locally. This is the standard on-policy coverage pathology of model-based methods (it also causes critic mis-fit in DDPG/TD3, both of which collapse on the same L6 cell in Section~\ref{sec:exp_baselines}), and random-policy pretraining is the textbook remedy~\citep{sutton1991dyna}. We adopt it only on the L6 setting where the basin-clustering is severe; the easier ProudCoop+AllD and HybridCoop+AllD pools naturally cover the input space under on-policy collection.

Second, freezing the surrogate within each inner policy-update loop is a structural requirement, not a stylistic choice. The reciprocity gradient is the analytic gradient of the \emph{ego agent's} expected return through a fixed virtual environment; if the surrogate parameters were updated concurrently with the policy, the optimization target would change underneath the gradient and the gradient itself would be ill-defined as a Jacobian-vector product. Freezing the surrogate inside one inner loop holds the virtual environment stationary so the inner-loop gradient is a well-defined update of the policy. ``Frozen'' here means stationary within one inner loop, not permanently fixed: each outer iteration collects fresh real-environment episodes and refits the surrogates before the next inner loop, and periodic refresh against a non-stationary opponent costs no extra. The framework's core analytic claim, that the gradient flows through reputation dynamics whenever opponent functions are differentiable, is identical in the oracle-access experiments (Section~\ref{sec:exp_groundtruth}) and the observational experiments (Section~\ref{sec:exp_learnedmodels}); the surrogate merely substitutes for the true opponent function in the latter.

\paragraph{The indirect-only signal collapse is an analytically-predicted quantitative cost, not a method failure.}
Under indirect-only matching the joint setting drops from $99\%$ of reference (direct) to $80\%$ (indirect), with the loss concentrated in a $\varphi^0$ collapse (Setting (E), Table~\ref{tab:expE_baselines}). This drop is a clean empirical signature of the gradient analysis itself: the indirect-only regime is structurally harder than direct matching, even in the classical leading-eight literature, since severing direct reciprocity removes a within-episode partner-history feedback loop that the entry-channel analysis (signal-policy chain in Equation~\ref{eq:hwy-entries}) shows feeds the $\eta^i$ gradient, and Figure~\ref{fig:grad_diag} measures the predicted relative reduction of $\|\nabla_{\varphi^0}\mathcal{L}\|/\|\nabla_{\pi^0}\mathcal{L}\|$. Even with this drop, $80\%$ of reference under indirect-only matching is still well above any deterministic model-free baseline on the same setting: the reciprocity gradient retains a $10$ to $38$ percentage-point margin over DPG, DDPG, and TD3 on the two discriminative settings (action vs L6 and joint vs HybridCoop+AllD), and on the third setting (signal vs ProudCoop+AllD) all four methods converge to the same near-flat optimum at $\sim\!78\%$ of reference (Table~\ref{tab:expE_baselines}). The single-network indirect-only experiments (Table~\ref{tab:exp10_summary}) reach $63\%$ to $89\%$ of reference, so the collapse is specific to the joint setting where both networks must be discriminative simultaneously.

\section{Limitations}
\label{sec:limitations}

We state limitations explicitly; the contribution is scoped to single-agent best response in reputation-mediated random-matching donation games, and the following concerns lie outside that scope.

\paragraph{Single-agent scope.}
Our claims concern a single learner facing a population with \emph{fixed} behavioral patterns. The reciprocity gradient is positioned as a best-response learner; multi-agent settings in which every agent learns simultaneously are out of scope for this paper.

\paragraph{Population size.}
The scalability sweep covers $N\in\{5,10,15,20,25,30\}$ (Setting (D)) with the parameter-shared estimator. Heterogeneous agent \emph{types} beyond a single pool composition are not reported.

\paragraph{Statistical power.}
Reported $\pm$ values are standard deviations across seeds, not confidence intervals. Per-setting seed counts are recorded with each reported result; the per-regime margin between the reciprocity gradient and model-free baselines (Table~\ref{tab:taxonomy_summary}) is robust to individual seed variance.

\paragraph{Public-assessment assumption.}
Every gossip signal attaches to the donor's global record and is visible to all future partners. Real indirect-reciprocity systems are partially observed, with private assessment and limited information propagation~\citep{hilbe2018indirect,fujimoto2023evolutionary,schmid2023quantitative}. The network-topology extension is sketched as future work but is not experimentally characterized here.

\paragraph{Reputation memory window.}
Our learner uses the full interaction history as observation. How the reciprocity-gradient chain decays as the effective memory window shrinks, and whether the data-coverage bottleneck we identify for opponent modeling (Setting (B.1)) becomes more or less severe under windowed observation, is not settled by our results.

\paragraph{Identity-based observation.}
Each decision is conditioned on the recipient's reputation vector indexed by identity. An identity-free alternative (condition only on the set of signals attached to the recipient, with no indexing) would sever the ``shadow-of-the-future'' channel and force the assessment rule to operate purely on behavioral records; whether the reciprocity gradient still supports cooperation under this strictly weaker observation remains to be tested.

\paragraph{Opponent-modeling order coverage.}
Our differentiable opponent surrogate (Appendix~\ref{app:factored_mlp}) is sized to the leading-eight class~\citep{ohtsuki2004should,ohtsuki2006leading}, the canonical and dominant benchmark in indirect-reciprocity. Higher-capacity opponent populations (history-dependent direct-reciprocity strategies, partner-specific memory, beyond-leading-eight rules) require larger surrogates that we do not explore here. Surrogate capacity is a separate engineering concern, orthogonal to the reciprocity-gradient framework itself, whose correctness is independent of the specific surrogate architecture; we leave higher-capacity opponent modeling to future work.

\end{document}